\documentclass[letterpaper, 10 pt, conference]{ieeeconf}% Comment this line out if you need a4paper

\IEEEoverridecommandlockouts                              % This command is only needed if 
\usepackage[dvipsnames, rgb, table]{xcolor}
\usepackage{lipsum}
\usepackage{soul}
\usepackage{color}
\usepackage[hidelinks]{hyperref}

\usepackage{tikz}
\usepackage{bm}
\usepackage{pgfplots}
\pgfplotsset{compat=1.18}
\usepackage{pgfplotstable}
\usepackage{booktabs}
\usepackage{float}
\usepackage{multirow}
\usepackage{etoolbox}
\usepackage{adjustbox}
\usepackage{diagbox}
\usepackage{graphics} 
\usepackage{epsfig} 
\usepackage{mathptmx} % assumes new font selection scheme installed % -- this will change math symbol style
\usepackage{times} 
\usepackage{amsmath}
\usepackage{amsfonts}
\usepackage{amssymb} 
\usepackage{multicol}

\usepgfplotslibrary{fillbetween}
\pgfdeclarelayer{background}
\pgfdeclarelayer{back background}
\pgfsetlayers{back background, background, main}
\usetikzlibrary{calc}
\usetikzlibrary{fit}
\usetikzlibrary{bayesnet}
\usetikzlibrary{arrows}
\usetikzlibrary{arrows.meta}
\usetikzlibrary{positioning}
\usetikzlibrary{shapes}
\usetikzlibrary{shapes.geometric}
\usetikzlibrary{shadows}
\usetikzlibrary{patterns}
\usetikzlibrary{decorations.pathreplacing}
\usetikzlibrary{fadings}

\definecolor{ffblue}{RGB}{097, 108, 140}
\definecolor{ffdarkgreen}{RGB}{086, 140, 135}
\definecolor{fflightgreen}{RGB}{178, 213, 155}
\definecolor{ffyellow}{RGB}{242, 222, 121}
\definecolor{ffred}{RGB}{217, 095, 024}

\definecolor{ffred_pv}{RGB}{202, 074, 046}
\definecolor{fforange_pv}{RGB}{232, 141, 047}
\definecolor{ffgreen_pv}{RGB}{059, 165, 149}
\definecolor{ffgreendark_pv}{RGB}{032, 117, 106}
\definecolor{nature_tab_gray1}{HTML}{D8D6C2}
\definecolor{nature_tab_gray2}{HTML}{ECEADF}
\title{\LARGE \bf
Robotic Long-Horizon Manipulation with Bayesian Non-parametric Skill Priors
}

\author{Yuan Meng$^{1}$, 
Xiangtong Yao$^{1}$,
Yansong Wu$^{1}$, 
Liding Zhang$^{1}$,
Yixiao Nie$^{1}$,\\
Zhenshan Bing$^{2,\dagger}$, 
and Alois Knoll$^{1}$ \textit{IEEE fellow}% <-this % stops a space
\thanks{$^{1}$The School of Computation, Information and Technology, Technical University of Munich, Germany.}%
\thanks{$^{2}$The State Key Laboratory for Novel Software Technology, School of Intelligence Science and Technology, Nanjing University, China.}%
\thanks{$^{\dagger}$ Corresponding author: \tt\small bing@nju.edu.cn}
\thanks{Videos and data are available at \url{https://ghiara.github.io/HELIOS/}}
}

\begin{document}

\maketitle
\thispagestyle{empty}
\pagestyle{empty}

%%%%%%%%%%%%%%%%%%%%%%%%%%%%%%%%%%%%%%%%%%%%%%%%%%%%%%%%%%%%%%%%%%%%%%%%%%%%%%%%
\begin{abstract}
Long-horizon manipulation under sparse rewards remains challenging for reinforcement learning due to delayed feedback and inefficient exploration. Existing skill-based approaches often assume a fixed parametric prior (e.g., a single Gaussian), limiting their ability to capture diverse and multi-modal skill structures required for complex tasks. We propose a Bayesian non-parametric skill prior that models temporally extended skills in a structured latent space using a Dirichlet Process Mixture, enabling adaptive skill discovery without predefining the number of components. Integrated into a hierarchical RL framework, the learned prior guides high-level skill selection while a pretrained decoder generates temporally abstracted actions, improving exploration efficiency in sparse-reward settings. Experiments on Franka Kitchen, LIBERO-Long, Meta-World, and a real robot demonstrate consistent gains in long-horizon manipulation, achieving over 0.8 success rate within 1.5M steps, whereas SAC fails to converge even after 5M steps ($<$0.1). Compared to a single-Gaussian prior baseline, our model yields an average improvement of 21.8\%.
\end{abstract}

%%%%%%%%%%%%%%%%%%%%%%%%%%%%%%%%%%%%%%%%%%%%%%%%%%%%
% -- introduction --
\section{INTRODUCTION}

Advances in artificial intelligence have shown that leveraging prior experience can substantially improve learning efficiency on new tasks \cite{pertsch2021accelerating,zhang2024review,11123767,meng2025growing}.
In reinforcement learning (RL), however, agents often start each task from scratch, requiring extensive data and exploration.
While dense reward design can partially alleviate this issue, it introduces significant manual engineering effort and lacks scalability across diverse tasks.
Under sparse-reward settings, the difficulty becomes more pronounced, especially in long-horizon manipulation, where delayed credit assignment and combinatorial skill compositions severely hinder exploration efficiency.
These challenges motivate the use of temporally extended skills as structured priors to guide exploration.
Recent work has explored extracting ``skills'' or reusable behaviors from unstructured experience \cite{pertsch2021accelerating}.
These temporally extended actions can be transferred across tasks by training a high-level policy to select from a learned skill set.
However, most existing approaches assume a simple and fixed structure, such as a single Gaussian distribution, to represent the skill prior \cite{pertsch2021accelerating}.
Although computationally convenient, such a unimodal and pre-defined assumption inherently limits the modeling of various skill distributions required for complex long-horizon manipulation.
Furthermore, in long-horizon tasks, the number of required underlying skills is typically unknown a priori; fixing a predefined prior structure prevents the agent from adaptively adjusting skill complexity as learning progresses, and exploration in such restricted latent spaces remains inefficient.

To overcome these limitations, we propose a Bayesian non-parametric skill prior that adaptively models temporally extended skills with flexible structure.
We argue that primitive manipulation skills naturally exhibit multi-peak and non-parametric characteristics \cite{yu2020meta}, where the effective number of skill modes should be inferred from data rather than pre-specified.
We leverage a Dirichlet Process Mixture (DPM) model \cite{hughes2013memoized} combined with a GRU-based skill variational auto-encoder (VAE) to dynamically discover and organize diverse behaviors without predefining the number of components.
We further incorporate a birth-and-merge heuristic to refine the latent structure, making discovered skills explicitly trackable and interpretable.
Integrated into a hierarchical RL framework, the non-parametric prior guides high-level skill selection while a pretrained decoder generates temporally extended actions, improving exploration efficiency under sparse rewards.
This structured prior reshapes exploration in sparse-reward long-horizon tasks, improving sample efficiency and stabilizing training.
We evaluate our method on Franka Kitchen \cite{fu2020d4rl}, LIBERO-Long \cite{liu2023libero}, Meta-World \cite{yu2020meta}, and a real robotic platform.
Across these benchmarks, the non-parametric skill prior captures structured long-horizon skill representations and consistently improves learning efficiency compared to parametric baselines.
In sparse-reward settings, our approach achieves over 21\% average relative improvement across benchmarks while requiring substantially fewer training steps ($\leq$1.5M), whereas baseline methods fail to converge even after 5M steps. 
These results demonstrate that adaptive Bayesian skill modeling effectively addresses sparse-reward exploration and long-horizon manipulation.

Our main contributions are threefold:
(1) We introduce a Bayesian non-parametric formulation of skill priors that adaptively captures multi-peak and long-horizon manipulation skills in an interpretable latent space.
(2) We integrate the adaptive skill prior into a hierarchical maximum-entropy RL framework, reshaping exploration under sparse rewards and improving sample efficiency.
(3) We validate the framework across multiple simulated benchmarks and real-world manipulation tasks, demonstrating consistent gains over parametric skill priors and conventional RL baselines.
We name our proposed framework \textbf{HELIOS}: \textbf{H}ierarchical \textbf{E}ncoding of \textbf{L}ong-horizon \textbf{I}nference \textbf{O}ff-policy Reinforcement Learning Framework with Bayesian Non-parametric \textbf{S}kill Prior.

\section{RELATED WORK}
\textbf{Meta-reinforcement learning} \cite{rakelly2019efficient,bing2023meta} aims to learn transferable priors from previous tasks to improve adaptation to new ones. 
However, meta-RL typically requires a predefined task distribution and online interaction during meta-training, limiting its ability to leverage large offline datasets. 
Offline RL methods \cite{meng2025preserving,10.1109/TNNLS.2023.3270298} instead learn from logged experience without further environment interaction, but often rely on dense task-specific rewards, which are expensive to design for diverse long-horizon tasks. 
In contrast, our framework learns temporally extended skills directly from unstructured offline action data without reward annotations, enabling scalable transfer under sparse rewards.

\textbf{Behavior priors} have been widely used in offline RL to regularize policy learning and reduce distributional shift. 
Recent works model priors over individual actions \cite{siegel2020keep} or over temporally extended skills \cite{pertsch2021accelerating}. 
Most of these approaches assume a fixed parametric form for the prior, such as a single Gaussian. 
Unlike them, we model a flexible skill prior with an adaptive structure, allowing the number of underlying skill components to grow with data complexity, which is crucial for long-horizon manipulation.

\textbf{Bayesian (non-)parametric models} \cite{hughes2013memoized,reynolds2009gaussian,nalisnick2016stick,goyal2017nonparametric} provide tools to model complex and uncertain distributions. 
Parametric approaches, such as Gaussian Mixture Models \cite{reynolds2009gaussian}, assume a predefined number of components and have been used in robotic representation learning (e.g., TIGR \cite{10.1109/TNNLS.2023.3270298}). 
However, fixing the number of mixture components limits adaptability when the true skill complexity is unknown. 
Bayesian non-parametric models, specifically Dirichlet Process Mixtures \cite{hughes2013memoized}, instead adjust their complexity dynamically based on observations. 
While widely studied in machine learning, their use as adaptive skill priors for robotic long-horizon RL remains underexplored. 
Our work leverages Bayesian non-parametric modeling to construct an interpretable and dynamically structured skill prior that explicitly improves long-horizon manipulation tasks' performance under a sparse-reward setting.

\begin{figure*}[ht!]
% \vskip -0.1in
    \centering
    \resizebox{\textwidth}{!}{
    \begin{tikzpicture}[font=\sffamily]
        % backgrounds
        \fill[ffyellow!50, rounded corners=4pt, shading=axis, left color=ffyellow!70, right color=ffyellow!40]
            (5.5,0) -- (5.5,8) -- (18,8) -- (18,0) -- cycle;
        \fill[fflightgreen!50, rounded corners=4pt, shading=axis, left color=fflightgreen!40, right color=fflightgreen!70]
            (-2,0) -- (-2,8) -- (10,8) -- (10,5) -- (6,3) -- (6,0)-- cycle;
        % modules
        \node[draw, fill=ffdarkgreen!50, trapezium, trapezium stretches=true, trapezium angle=100, rounded corners=4pt, 
          minimum width=1cm, minimum height=1cm, text width=3cm, align=center] (motion_enc) at(3.5,6) {$q(z|\bm{a}_{i})$ \\ \footnotesize \textsf{Primitive Skill Encoder}};
        \node[draw, shading=axis, left color=ffdarkgreen!50, right color=fforange_pv!50, rectangle, rounded corners=4pt, 
          minimum width=1cm, minimum height=1cm, text width=3cm, align=center] (state_enc) at(8,6) {$p_\vartheta(z|s_t)$ \\ \footnotesize \textsf{Skill Prior}};
        \node[draw, fill=ffdarkgreen!50, trapezium, trapezium stretches=true, trapezium angle=80, rounded corners=4pt, 
          minimum width=1cm, minimum height=1cm, text width=3cm, align=center] (motion_dec) at(3.5,1.5) {$p(\bm{a}_i|z, s_t)$ \\ \footnotesize \textsf{Primitive Skill Decoder}};
        \node[inner sep=0pt, rounded corners=4pt, draw=none, fill=white, opacity=0.35, text opacity=1, rectangle] (dpmm) at([yshift=-1.75cm]motion_enc.south){\includegraphics[width=3cm]{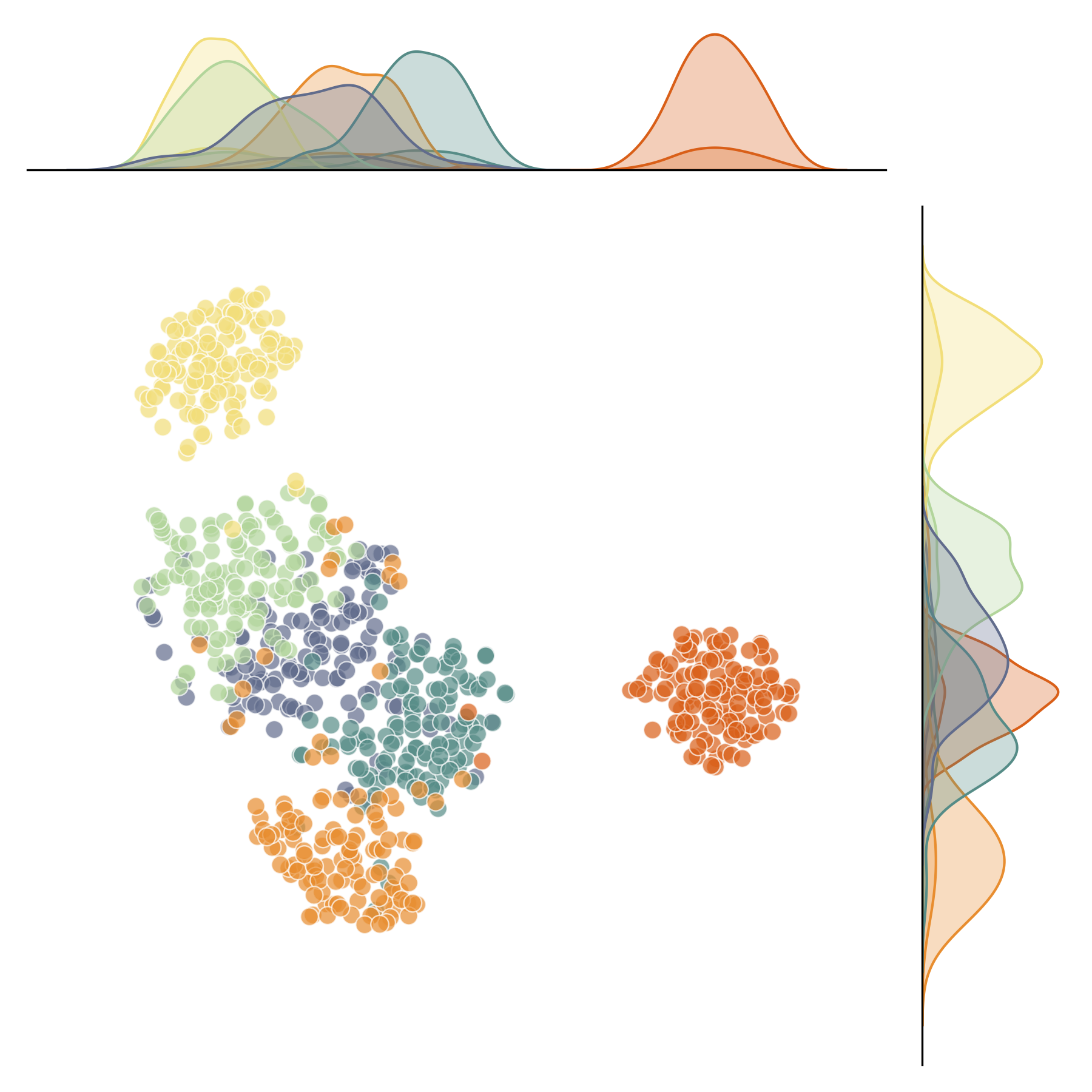}};
        
        \node[draw, fill=fforange_pv!50, rectangle, rounded corners=4pt, 
          minimum width=1cm, minimum height=1cm, text width=3cm, align=center] (rl_inference) at(12.5,6) {$\pi_\varphi(z|s_t)$ \\ \footnotesize \textsf{RL Upstream Inference}};
        \node[draw, fill=fforange_pv!50, trapezium, trapezium stretches=true, trapezium angle=80, rounded corners=4pt, 
          minimum width=1cm, minimum height=1cm, text width=3cm, align=center] (pre_motion_dec) at(12.5,1.5) {$p(\bm{a}_i|z, s_t)$ \\ \footnotesize \textsf{Primitive Skill Decoder}};

        % -- Phase 1: Pretrain --
        \shade[fflightgreen!40, rounded corners=4pt, shading=axis, top color=ffdarkgreen!50, bottom color=fflightgreen!40]
            (-1.5,0) -- (-1.5,4) -- (0.5,4) -- (0.5,0) -- cycle;
        \shade[fflightgreen!40, rounded corners=4pt, shading=axis, top color=fflightgreen!40, bottom color=ffdarkgreen!50]
            (-1.5,3.8) -- (-1.5,8) -- (0.5,8) -- (0.5,3.8) -- cycle;
        \node[inner sep=-1pt, rectangle, rounded corners=4pt, clip, draw=none, fill=none,] (motion1) at(-0.5,1.8) {\includegraphics[width=1.5cm]{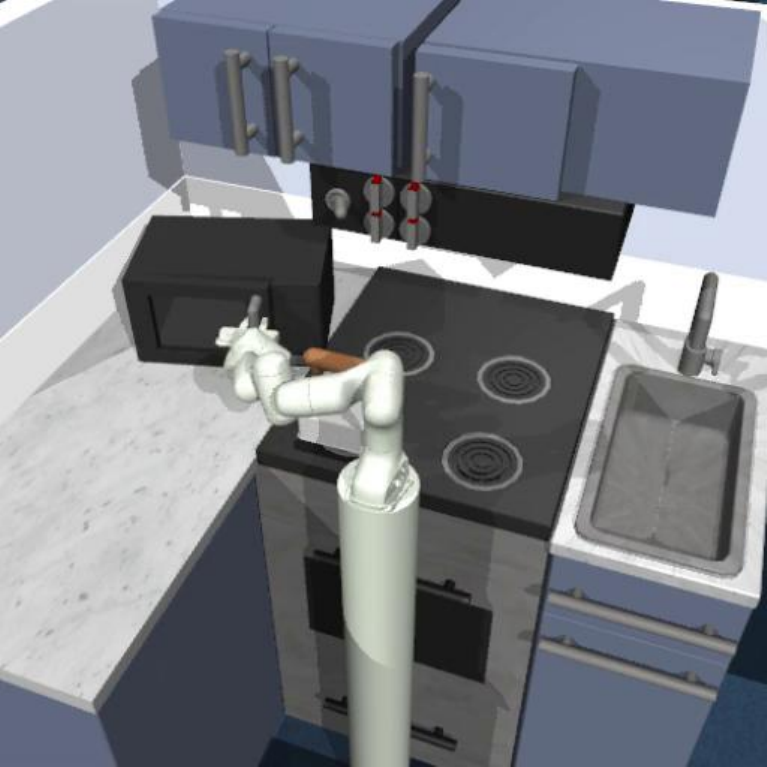}};
        \node[inner sep=-1pt, rectangle, rounded corners=4pt, clip, draw=none, fill=none,] (motion2) at([yshift=0.8cm]motion1.north) {\includegraphics[width=1.5cm]{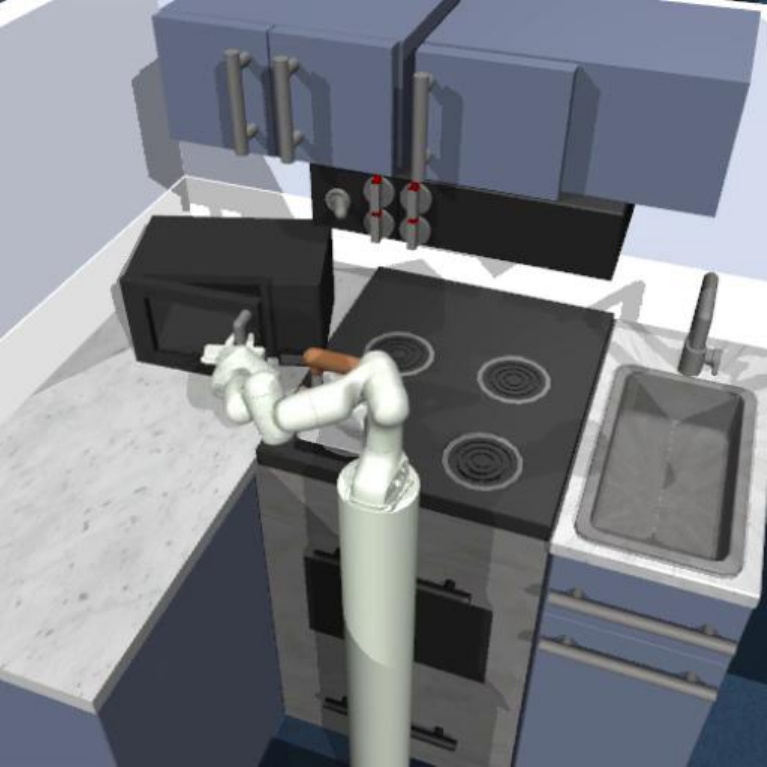}};
        \node[inner sep=-1pt, rectangle, rounded corners=4pt, clip, draw=none, fill=none,] (motion3) at([yshift=0.8cm]motion2.north) {\includegraphics[width=1.5cm]{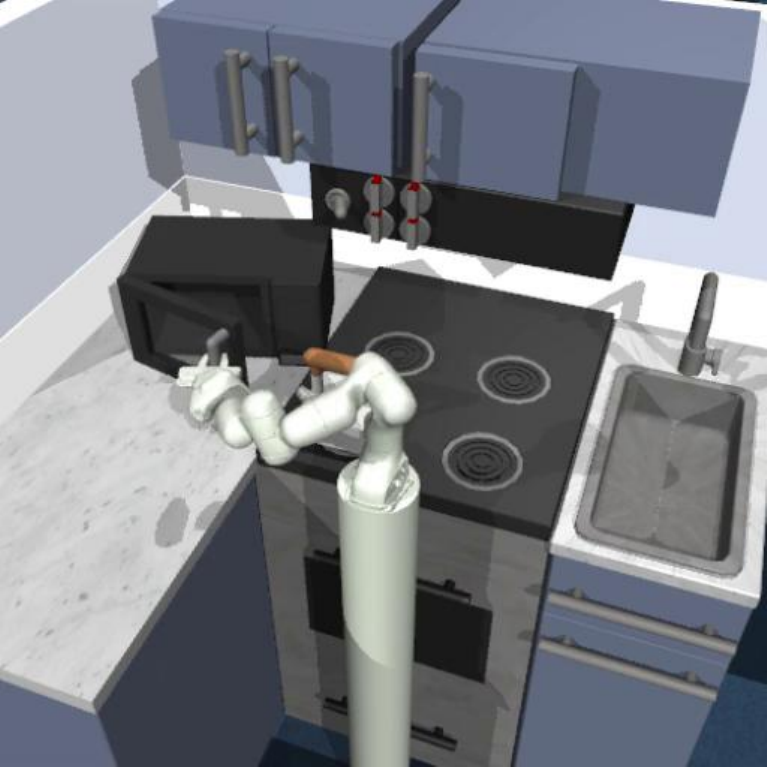}};
        \node[inner sep=-1pt, rectangle, rounded corners=4pt, clip, draw=none, fill=none,] (motion4) at([yshift=0.8cm]motion3.north) {\includegraphics[width=1.5cm]{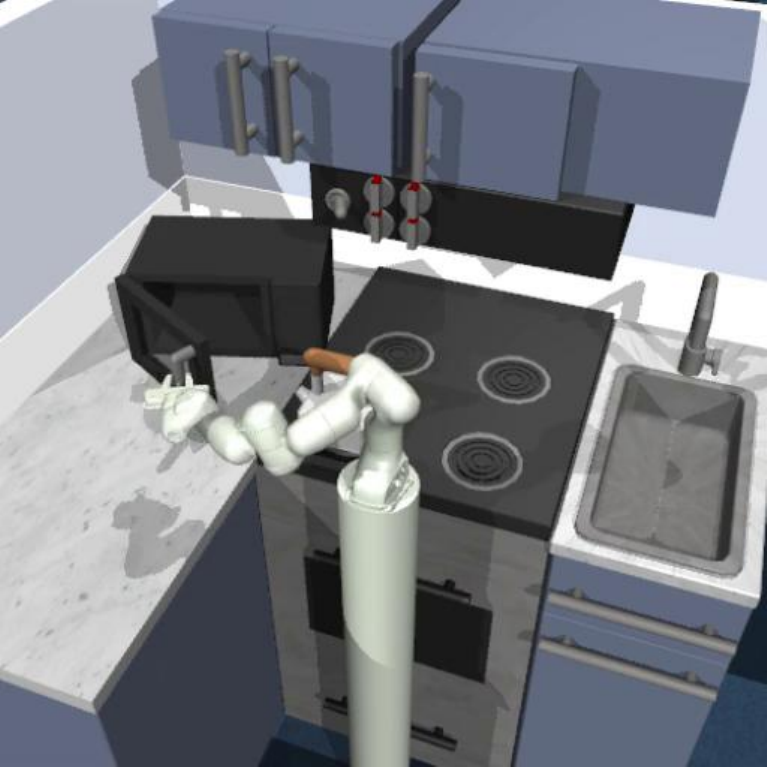}};
        \node[inner sep=-1pt, rectangle, rounded corners=4pt, clip, draw=none, fill=none,] (motionhead1) at([yshift=-0.4cm]motion1.south) {\includegraphics[width=1.5cm]{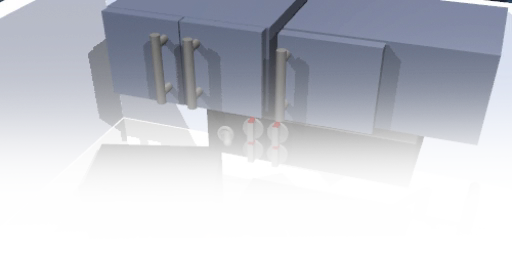}};
        \node[inner sep=-1pt, rectangle, rounded corners=4pt, clip, draw=none, fill=none,] (motiontail1) at([yshift=0.4cm]motion4.north) {\includegraphics[width=1.5cm]{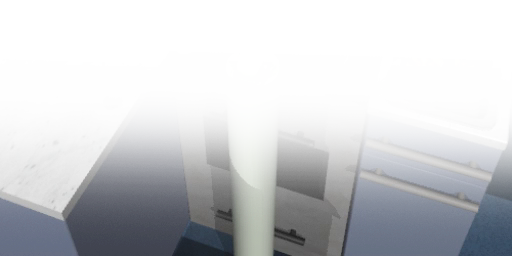}};
        \node[draw, rectangle, rounded corners=4pt, minimum width=1.5cm, minimum height=0.2cm](l_recon) at([yshift=-0.5cm]motion_dec.south) {$\mathcal{L}_{rec}$};
        \draw[-{Triangle Cap []. Fast Triangle[] Fast Triangle[]}, draw=ffgreen_pv!70, line width=.8mm] (motion1.east) to[out=45, in=-45] (motion2.east);
        \draw[-{Triangle Cap []. Fast Triangle[] Fast Triangle[]}, draw=ffgreen_pv!70, line width=.8mm] (motion2.east) to[out=45, in=-45] (motion3.east);
        \draw[-{Triangle Cap []. Fast Triangle[] Fast Triangle[]}, draw=ffgreen_pv!70, line width=.8mm] (motion3.east) to[out=45, in=-45] (motion4.east);
        \draw[-{Triangle Cap []. Fast Triangle[] Fast Triangle[]}, draw=ffgreen_pv!70, line width=.8mm] (motion4.east) to[out=45, in=-45] (motiontail1.north east);
        \node[rounded corners=2pt, rectangle, fill=ffdarkgreen!80, opacity=0.8, above right]at(motion1.south west){\textcolor{white}{$s_t$}};
        \node[rounded corners=2pt, rectangle, minimum width=0.9cm, fill=ffgreen_pv!70, opacity=0.8, below right]at([xshift=0.4cm, yshift=0.25cm]motion1.north east){$a_t$};
        \node[rounded corners=2pt, rectangle, minimum width=0.9cm, fill=ffgreen_pv!70, opacity=0.8, below right]at([xshift=0.4cm, yshift=0.25cm]motion2.north east){$a_{t+1}$};
        \node[rounded corners=2pt, rectangle, minimum width=0.9cm, fill=ffgreen_pv!70, opacity=0.8, below right]at([xshift=0.4cm, yshift=0.25cm]motion3.north east){$a_{t+i}$};
        \node[rounded corners=2pt, rectangle, minimum width=0.9cm, fill=ffgreen_pv!70, opacity=0.8, below right]at([xshift=0.4cm, yshift=0.25cm]motion4.north east){$a_{t+L}$};
        \node[](help1)at([xshift=-6.5cm, yshift=0.5cm]state_enc.north){};
        \draw[-{Triangle Cap []. Fast Triangle[] Fast Triangle[]}, draw=ffgreendark_pv!70, line width=1mm, rounded corners=1pt] (help1) -- ([yshift=0.5cm]state_enc.north) -- (state_enc.north);
        \draw[-{Triangle Cap []. Fast Triangle[] Fast Triangle[]}, draw=ffgreen_pv!70, line width=1mm, rounded corners=1pt] (help1) -- ([yshift=0.5cm]motion_enc.north) -- (motion_enc.north);
        \draw[-{Triangle Cap []. Fast Triangle[]}, draw=ffgreen_pv!70, line width=1mm, rounded corners=1pt] (motion_enc.south) --++ (0, -0.3cm);
        \draw[-{Triangle Cap []. Fast Triangle[]}, draw=ffgreen_pv!70, line width=1mm, rounded corners=1pt] ([yshift=0.3cm]motion_dec.north) -- (motion_dec.north);
        \draw[-{Triangle Cap []. Fast Triangle[]}, draw=ffgreen_pv!60, line width=1mm] (motion_dec.south) -- (l_recon);
        \draw[-{Triangle Cap []. Fast Triangle[] Fast Triangle[]}, draw=ffgreen_pv!70, line width=1mm, rounded corners=1pt] ([xshift=-2cm]l_recon.west) -- (l_recon);

        % -- Phase 2: RL --
        % \fill[ffyellow!50, rounded corners=4pt, shading=axis, bottom color=fforange_pv!30, top color=fforange_pv!50]
        %     (15,0.5) -- (15,7) -- (16,7.5) -- (17,7) -- (17,0.5) -- cycle;
        \fill[ffyellow!40, rounded corners=4pt, shading=axis, bottom color=ffyellow!40, top color=fforange_pv!50]
            (15.5,0) -- (15.5,4) -- (17.5,4) -- (17.5,0) -- cycle;
        \fill[ffyellow!40, rounded corners=4pt, shading=axis, bottom color=fforange_pv!50, top color=ffyellow!40]
            (15.5,3.8) -- (15.5,8) -- (17.5,8) -- (17.5,3.8) -- cycle;
        \node[inner sep=-1pt, rectangle, rounded corners=4pt, clip, draw=none, fill=none,label={[label distance=-0.4cm, fill=white, fill opacity=0.7, text opacity=1, text=black, rounded corners=2pt]-90:\scriptsize \textsf{\textbf{Microwave}}}] (task1) at(16.5,1.8) {\includegraphics[width=1.5cm]{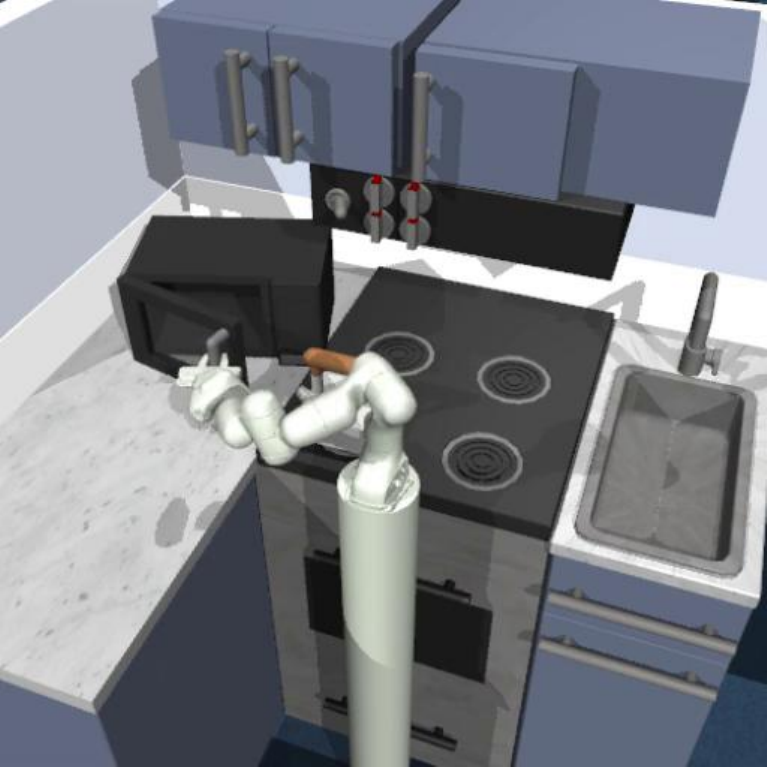}};
        \node[inner sep=-1pt, rectangle, rounded corners=4pt, clip, draw=none, fill=none,label={[label distance=-0.4cm, fill=white, fill opacity=0.7, text opacity=1, text=black, rounded corners=2pt]-90:\scriptsize \textsf{\textbf{kettel}}}] (task2) at([yshift=0.8cm]task1.north) {\includegraphics[width=1.5cm]{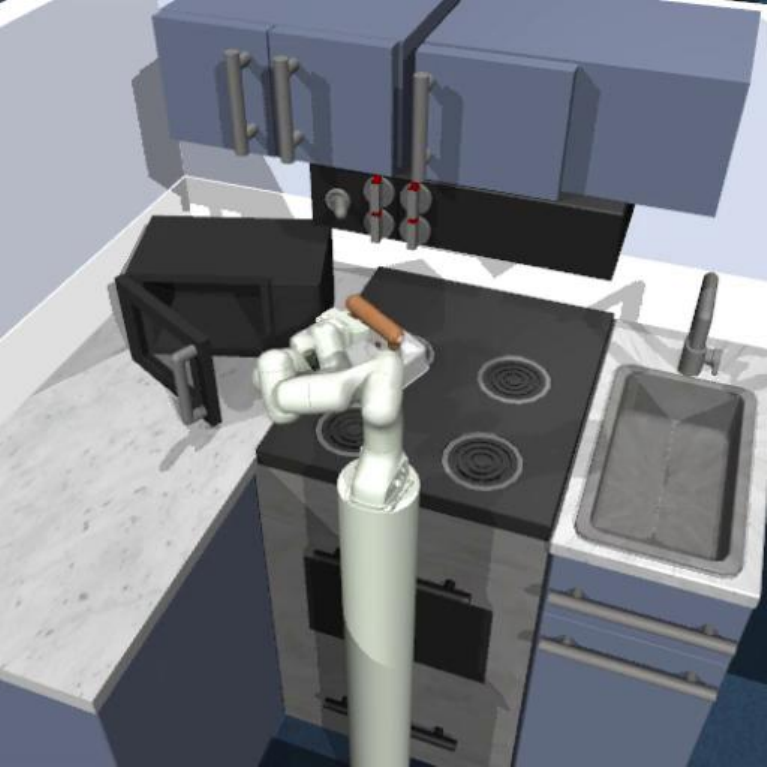}};
        \node[inner sep=-1pt, rectangle, rounded corners=4pt, clip, draw=none, fill=none,label={[label distance=-0.4cm, fill=white, fill opacity=0.7, text opacity=1, text=black, rounded corners=2pt]-90:\scriptsize \textsf{\textbf{Burner}}}] (task3) at([yshift=0.8cm]task2.north) {\includegraphics[width=1.5cm]{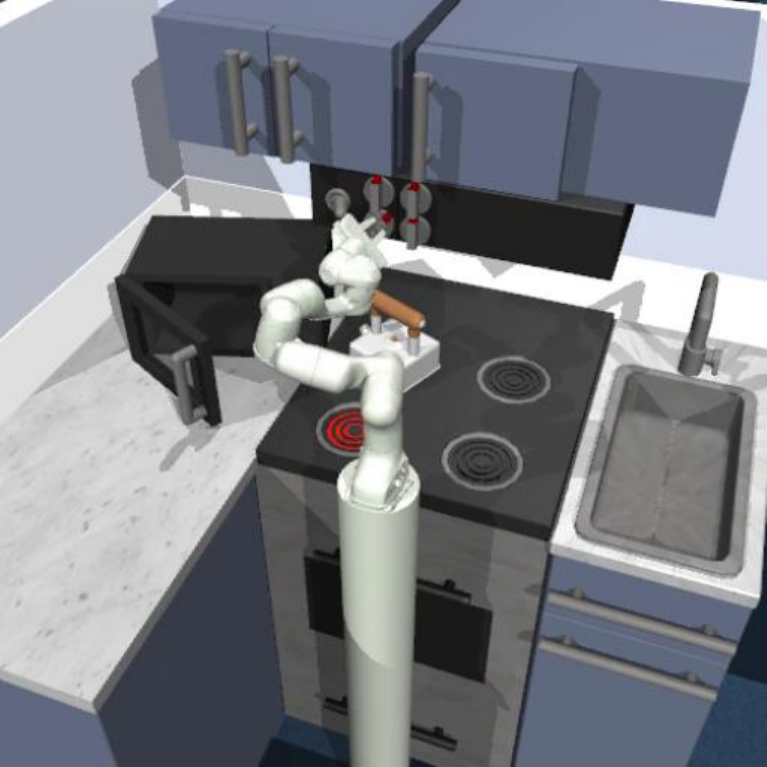}};
        \node[inner sep=-1pt, rectangle, rounded corners=4pt, clip, draw=none, fill=none,label={[label distance=-0.4cm, fill=white, fill opacity=0.7, text opacity=1, text=black, rounded corners=2pt]-90:\scriptsize \textsf{\textbf{Light}}}] (task4) at([yshift=0.8cm]task3.north) {\includegraphics[width=1.5cm]{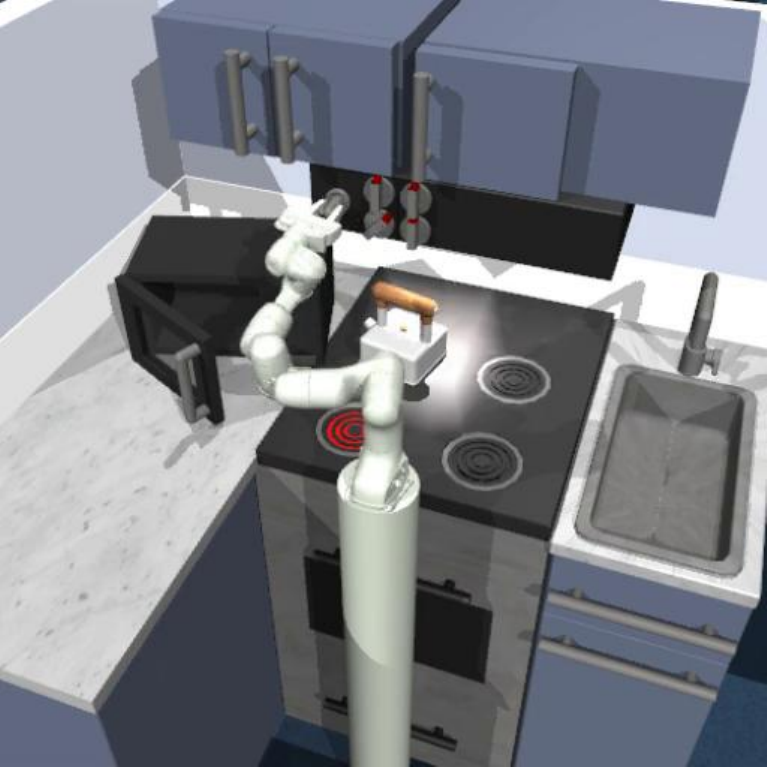}};
        \node[inner sep=-1pt, rectangle, rounded corners=4pt, clip, draw=none, fill=none,] (motionhead2) at([yshift=-0.4cm]task1.south) {\includegraphics[width=1.5cm]{imgs/motion_head.png}};
        \node[inner sep=-1pt, rectangle, rounded corners=4pt, clip, draw=none, fill=none,] (motiontail2) at([yshift=0.4cm]task4.north) {\includegraphics[width=1.5cm]{imgs/motion_tail.png}};
        \draw[-{Triangle Cap []. Fast Triangle[] Fast Triangle[]}, draw=fforange_pv!70, line width=1mm, rounded corners=1pt] ([xshift=3cm, yshift=0.5cm]rl_inference.north) -- ([yshift=0.5cm]rl_inference.north) -- (rl_inference.north);
        \draw[-{Triangle Cap []. Fast Triangle[] Fast Triangle[]}, draw=fforange_pv!70, line width=1mm, rounded corners=1pt] (pre_motion_dec.south) -- ++ (0, -0.2cm) -- ++ (3cm, 0);

        \node[draw, rectangle, rounded corners=4pt, minimum width=1.2cm, minimum height=0.2cm, text width=4.5cm, align=center](l_kld)at(8,4){\footnotesize \textsf{Kullback–Leibler Divergence} \\ \normalsize $\mathcal{L}_{KLD}$};
        \node[](help2)at([xshift=1cm]l_kld.east){};
        \draw[-{Triangle Cap []. Fast Triangle[] Fast Triangle[]}, draw=fforange_pv!70, line width=1mm, rounded corners=1pt] ([xshift=-1cm]rl_inference.south) |- (l_kld.east);
        \draw[-{Triangle Cap []. Fast Triangle[] Fast Triangle[]}, draw=ffgreendark_pv!70, line width=1mm, rounded corners=1pt] (state_enc.south) -- (l_kld.north);
        \draw[-{Triangle Cap []. Fast Triangle[] Fast Triangle[]}, draw=ffgreendark_pv!70, line width=1mm, rounded corners=1pt] (l_kld.west) --++ (-0.6cm,0);
        \draw[-{Triangle Cap []. Fast Triangle[] Fast Triangle[]}, draw=ffgreendark_pv!70, line width=1mm, rounded corners=1pt] (motion_dec.east) -- (pre_motion_dec.west) node[pos=0.5, below]{\footnotesize\textsf{Deploy}};
        \node[draw, rectangle, rounded corners=4pt, minimum width=0.5cm, minimum height=0.2cm, text width=2.5cm, align=center](entropy)at([xshift=3.65cm]l_kld.east){\footnotesize \textsf{Maximum Entropy} \\ \normalsize $J(\varphi)$};
        % \draw[-{Triangle Cap []. Fast Triangle[] Fast Triangle[]}, draw=fforange_pv!70, line width=1mm, rounded corners=1pt] ([xshift=1cm]rl_inference.south) --++ (0, -1cm);
        \draw[-{Triangle Cap []. Fast Triangle[] Fast Triangle[]}, draw=fforange_pv!70, line width=1mm, rounded corners=1pt] ([xshift=-0.5cm]entropy.north) --++ (0, 1cm);
        \draw[-{Triangle Cap []. Fast Triangle[] Fast Triangle[]}, draw=fforange_pv!70, line width=1mm, rounded corners=1pt] ([xshift=1.5cm]l_kld.south) |- (10cm, 3cm) -| (entropy.south);
        \draw[-, draw=ffyellow!50, line width=2mm, rounded corners=1pt] (12.35cm, 3cm) --++ (0.3cm, 0);
        \draw[-{Triangle Cap []. Fast Triangle[] Fast Triangle[]}, draw=fforange_pv!70, line width=1mm, rounded corners=1pt] (rl_inference.south) -- (pre_motion_dec.north);  
        
        % annotations
        \node[]at(state_enc.north east){\includegraphics[width=0.5cm]{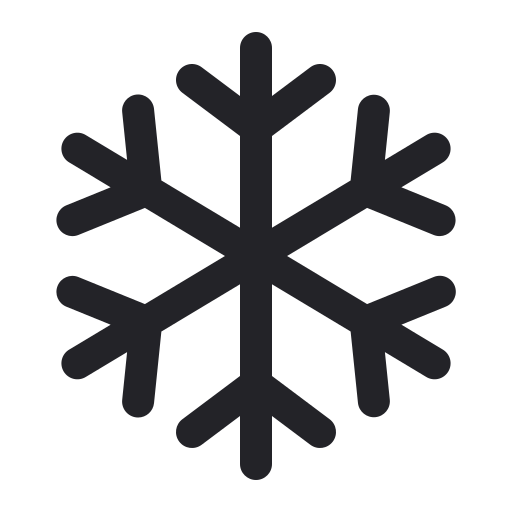}};
        \node[]at([xshift=-0.9cm]pre_motion_dec.north west){\includegraphics[width=0.5cm]{imgs/snowflake.png}};
        \node[draw=none]at(7.5, 7.7){\textcolor{ffdarkgreen}{\textsf{\textbf{Phase I: Skill Prior Pretrain}}}};
        \node[draw=none]at(9.8, 0.3){\textcolor{fforange_pv}{\textsf{\textbf{Phase II: RL for Long Horizon Manipulation}}}};
        \node[draw=none]at(motionhead1.south){\small\textcolor{ffdarkgreen}{\textsf{\textbf{Skill dataset}}}};
        \node[draw=none, above right, text width=3.3cm, align=center]at(dpmm.south east){\footnotesize \textsf{Bayesian Non-parametric \\ Skill Prior}};
        \node[draw=none, above right]at(motion_enc.north){$\bm{a}_{t:t+L}$};
        \node[draw=none, above right]at(state_enc.north){$s_t$};
        \node[draw=none, above right]at([xshift=1.4cm, yshift=-0.1cm]pre_motion_dec.south east){$\bm{a}_{t:t+L}$};
        \node[draw=none, above right]at(rl_inference.north){$s_t$};
        \node[draw=none]at(motionhead2.south){\small\textcolor{fforange_pv}{\textsf{\textbf{Environments}}}};
        
    \end{tikzpicture}
    }
    \vskip -0.1in
    \caption{\textbf{HELIOS Framework overview.} The training process is divided into two phases. In Phase I, a VAE with GRU modules is used to pre-train a skill representation model from a dataset of action trajectories. The model leverages a DPM to capture the non-parametric nature of skill priors, aiding in learning precise action patterns and subsequent effective task representations. In Phase II, this pre-trained skill decoder and prior are deployed within a RL framework to address long-horizon manipulation tasks. Here, the upstream inference model uses soft actor-critic structure to learn specific task reasoning, ensuring the successful execution of complex, extended long-horizon tasks.}
    \vskip -0.2in
    \label{fig:framework}
\end{figure*}
% --------------------------------

%%%%%%%%%%%%%%%%%%%%%%%%%%%%%%%%%%%%%%%%%%%%%%%%%%%%
% -- Preliminaries --
\section{PRELIMINARIES}
In this section, we first introduce the memorized online variational inference method for Dirichlet Process Mixtures, followed by the integration of maximum entropy principles in downstream off-policy RL learning.
\subsection{Variational Inference of DPM}
DPM is a widely used Bayesian non-parametric model designed to capture an infinite mixture of clusters when modeling a set of observations $\bm{x} = x_{1:N}$. 
The generative process of the DPM defines how clusters and data points are formed under a Bayesian non-parametric framework. In this model, each component $\theta^{*}_k$ is a cluster parameter sampled from a base distribution $\mathcal{H}(\lambda)$, parameterized by $\lambda$.
To determine the mixture proportions of these clusters, a mixing proportion distribution $\pi$ is drawn from the Generalized Ewens Distribution (GEM) with concentration parameter $\alpha$, which specifies the probability of assigning new data points to existing clusters versus creating new ones.
Given the $\pi$, a cluster assignment variable $c_i$ is sampled for each data point $x_i$ from a categorical distribution $\text{Cat}(\pi)$, where $c_i=k$ indicates that $x_i$ is assigned to cluster $k$. Finally, the data point $x_i$ itself is generated from the distribution $\mathcal{F}(\theta_i)$, parametrized by parameterized by the corresponding cluster parameter $\theta^{*}_{c_i}$. This process can be expressed as:
\begin{equation}
\begin{aligned}
    \theta^{*}_k | \lambda  & \sim \mathcal{H}(\lambda) \text{,}  \\  
    \pi | \alpha & \sim \text{GEM}(\alpha) \text{,} \\
    c_i | \pi & \sim \text{Cat}(\pi) \text{,} \\
    x_i | c_i  & \sim \mathcal{F}(\theta^{*}_{c_i}) \text{.}
\end{aligned}
\label{eq:dpmm_gen}
\end{equation}
In this work, we use an online variational inference-based method to estimate the true posterior density of observations, as it provides faster and more scalable solutions than sampling-based approaches. Given Eq.\eqref{eq:dpmm_gen}, the joint probability of the DPM parameters is expressed as:
\begin{equation}
\begin{aligned}
    p(\bm{x}, \bm{c}, \bm{\theta}, \bm{\beta}) =
        & \prod_{n=1}^N \mathcal{F}(x_n|\theta_{c_n})\text{Cat}(c_n|\bm{\pi}(\bm{\beta}))\\
        & \prod_{k=1}^{\infty}\mathcal{B}(\beta_k|1, \alpha)\mathcal{H}(\theta_k | \lambda) \text{.}
\end{aligned}
\label{eq:dpmm_p}
\end{equation}
Here, the parameter $\beta$ of mixing proportion $\pi$ is sampled from a beta distribution parameterized by $\alpha$. Since we cannot obtain the exact posterior $p(\bm{c}, \bm{\theta}, \bm{\beta}|\bm{x})$, the goal in variational inference is to find an optimal variational distribution $q^*(\bm{c}, \bm{\theta}, \bm{\beta})$ that minimizes the Kullback–Leibler (KL) divergence from the true posterior. This turns out to maximize the evidence lower bound (ELBO), which can be derived as follows:
\begin{equation}
\begin{aligned}
    \text{ELBO}(q) =& \mathbb{E} [\log p(\bm{x}|\bm{c}, \bm{\theta}, \bm{\beta})] \\
    & - \mathbb{KL}(q(\bm{c}, \bm{\theta}, \bm{\beta}) || p(\bm{c}, \bm{\theta}, \bm{\beta})) \text{.}
\end{aligned}
\label{eq:elbo1}
\end{equation}
For the DPM, we define the variational distribution $q$ using the mean field assumption \cite{blei2006variational}, where each latent variable has its own independent variational factor (indicated by $\hat{*}$). Specifically, we obtain:
\begin{equation}
\resizebox{.5\textwidth}{!}{
$\begin{aligned}
    q(\bm{c}, \bm{\theta}, \bm{\beta})  &= \prod_{n=1}^N q(c_n|\hat{r}_n)\prod_{k=1}^K q(\beta_k|\hat{\alpha}_{k_1}, \hat{\alpha}_{k_0}) q(\theta_k|\hat{\lambda}_k) \text{,}\\
    &=\prod_{n=1}^N \underbrace{\text{Cat}(c_n|\hat{r}_{n_1:n_K})}_{q_{c_n}}\prod_{k=1}^K \underbrace{\mathcal{B}(\beta_k|\hat{\alpha}_{k_1}, \hat{\alpha}_{k_0})}_{q_{\beta_k}} \underbrace{\mathcal{H}(\theta_k|\hat{\lambda}_k)}_{q_{\theta_k}} \text{,}\\
\end{aligned}$
}
\end{equation}
where $q_{c_n}$, $q_{\beta_k}$, and $q_{\theta_k}$ are categorical factors, beta distribution factors, and base distribution factors, respectively. Since the true posterior is infinite, we make the $q$ tractable by truncating the assignment factor to $K$ components, setting $q(c_n=k)=0$ for $k>K$. This allows inference to focus on a finite set of $K$ components, approximating the true infinite posterior effectively when $K$ is large enough. 
In the special case where both $\mathcal{H}$ and $\mathcal{F}$ belong to the exponential family, the optimization objective for the ELBO in Eq.\eqref{eq:elbo1}  can be reformulated in terms of the expected mass $\hat{N}_k$ and the expected sufficient statistics $s_k(x)$ for each component $k$ \cite{hughes2013memoized}:
%
% \begin{equation}
% \resizebox{.5\textwidth}{!}{
% $\begin{aligned}
%     \text{ELBO}(q) =& \sum_{k=1}^K\left[\mathbb{E}_q[\theta_k]^\top s_k(x) - \hat{N}_k[a(\theta_k)] + \hat{N}_k [\log \pi_k(\beta)] - \sum_{n=1}^N\hat{r}_{nk}\log \hat{r}_{nk} \right.\\
%     &\left. + \mathbb{E}_q[\log \frac{\mathcal{B}(\beta_k|1, \alpha)}{q(\beta_k|\hat{\alpha}_{k_1}, \hat{\alpha}_{k_0})}] + \mathbb{E}_q[\log \frac{\mathcal{H}(\theta_k|\lambda)}{q(\theta_k|\hat{\lambda}_{k})}]\right]\text{.}
% \end{aligned}$
% }
% \end{equation}
\begin{equation}
\resizebox{.485\textwidth}{!}{$
\begin{aligned}
    \text{ELBO}(q) =& \sum_{k=1}^K \Bigg[ \mathbb{E}_q[\theta_k]^\top s_k(x) - \hat{N}_k[a(\theta_k)] + \hat{N}_k [\log \pi_k(\beta)] \\
    & + \mathbb{E}_q[\log \frac{\mathcal{B}(\beta_k|1, \alpha)}{q(\beta_k|\hat{\alpha}_{k_1}, \hat{\alpha}_{k_0})}] + \mathbb{E}_q[\log \frac{\mathcal{H}(\theta_k|\lambda)}{q(\theta_k|\hat{\lambda}_{k})}]\\ 
    & - \sum_{n=1}^N\hat{r}_{nk}\log \hat{r}_{nk} \Bigg] \text{.}
\end{aligned}
$}
\label{eq:dpmm_opt_objective}
\end{equation}
We use coordinate ascent to iteratively update each variational factor. First, we update the local parameters $\hat{r}_{nk}$ for each cluster assignment factor $q_{c_n}$. Then, we adjust the global parameters in $q_{\beta_k}$ and $q_{\theta_k}$ to maximize the ELBO \cite{hughes2013memoized}.

In this work, we apply an online heuristic method, MemoVB, which introduces birth and merge moves for dynamic cluster adjustments \cite{hughes2013memoized}. To create new clusters, we collect subsamples $x'$ that are poorly described by a single cluster in each batch and fit a separate DPM with $K'$ initial clusters. Assuming the number of active clusters before the birth move is $K$, we decide to accept or reject new cluster proposals by comparing the likelihood of assigning $x'$ to $K + K'$ clusters versus $K$ clusters alone. 
To complement the birth move, the merge move consolidates two clusters into one if merging improves the ELBO objective, reducing the number of clusters to $K - 1$.

\subsection{Maximum Entropy in RL}
In offline RL, the maximum entropy principle promotes exploration by encouraging agents to maximize both the expected return $r$ and the entropy $H$ of their policy \cite{haarnoja2018soft}. The objective is formulated as:
\begin{equation}
% \begin{aligned}
    J(\varphi) = \sum_{t=1}^{T} \mathbb{E}_{(s_0, a_0) \sim \rho_{\pi}} \left[ r(s_t, a_t) + \omega H(\pi_\varphi(\cdot | s_t)) \right] \text{,}
    % & H(\pi_\varphi(\cdot|s)) = - \mathbb{E}_{a \sim \pi_\varphi(\cdot|s)} \left[ \log \pi_\varphi(a|s) \right] \text{,}
% \end{aligned}
\label{eq:sac_entropy}
\end{equation}
where $r(s_t, a_t)$ is the reward for the state-action pair, $\rho_\pi$ is the state-action distribution induced by policy $\pi_\varphi$ (parameterized by $\varphi$), $\omega$ is the temperature parameter that balances the trade-off between reward and entropy. 
This approach is particularly advantageous in offline settings, where the agent's exploration is constrained to a large offline dataset $\mathbb{D}$. Balancing exploration and exploitation is challenging in such settings, especially as the number of embedded skills increases rapidly.

% -- Method --
\section{METHOD}
In this section, we present the methodology of our framework, as illustrated in Fig. \ref{fig:framework}. The framework is organized into two main phases: skill prior pretraining and RL for long-horizon manipulation. 
% We start with the problem formulation in the RL setting, followed by a detailed explanation of the Bayesian non-parametric skill prior to pretraining (Phase I), and conclude with the downstream training process for long-horizon manipulation tasks (Phase II) using the pre-trained skill prior.

\subsection{Problem Statement}
In this work, we address the challenging problem of long-horizon manipulation in reinforcement learning by developing a framework that utilizes Bayesian non-parametric skill priors to guide downstream RL policy learning.  
We formulate this problem within the standard RL framework, represented as a Markov Decision Process (MDP) defined by the tuple $\{S, A, T, R, \gamma\}$, where $S$ is the state space, $A$ the action space, $T$ the transition function, $R$ the reward function, and $\gamma$ the discount factor.
Our approach is motivated by the need for efficient long-horizon task learning within a data-driven deep RL setting, where conventional RL methods often struggle due to the complexity of learning from scratch. 
Inspired by prior work \cite{pertsch2021accelerating}, we recognize that existing unstructured datasets contain informative data that can be captured by a skill prior module to accelerate exploration in downstream tasks.
In this context, we assume that action-based skills within the dataset exhibit an unknown number of features, embodying non-parametric properties \cite{yu2020meta} that cannot be adequately represented by a single Gaussian distribution.
To address this, we employ a DPM for skill prior learning, allowing our model to capture flexible, non-parametric skill representations. 
The pre-trained skill prior is then integrated into the RL framework, enhancing both exploration efficiency and adaptability in complex, long-horizon tasks.

\subsection{Bayesian Non-parametric Skill Prior}\label{sec:skill_prior_method}
The training process is divided into two phases. 
As shown in Fig.\ref{fig:framework}, the first phase (highlighted in green) focuses on training a Bayesian non-parametric skill prior using a DPM model in the latent representation space.
This phase leverages a VAE with a GRU-based backbone to model the non-parametric distributions of temporally extended skills based on an unstructured dataset. 

During this phase, training proceeds iteratively. 
First, we freeze the network parameters to update the DPM.
In the initial epoch, we start with a single component in the DPM to facilitate computation.
In the subsequent epochs, we collect embeddings of the inferred skill posterior $q(z|\bm{a}_i)$ and fit them to the DPM from the previous epoch.
According to the DPM optimization objective from Eq.\eqref{eq:dpmm_opt_objective}, new components can be created or existing ones merged to improve the ELBO. 
The latest DPM configuration replaces the previous one, preparing for the next iteration. 
Note that we do not require the DPM to converge very early, since the skill posterior is also not well-trained.
Additionally, we compute the soft KL divergence as a weighted sum between the skill posterior distribution and the DPM components: $\sum_{k=1}^K p_{ik} \mathbb{KL}(q(z|\bm{a}_i) || \mathcal{N}(\mu_k, \Sigma_k))$, where $K$ is is the current number of components, $p_{ik}$ denotes the probability of assigning the $i$-th data point to the $k$-th component, and $\mu_k$, $\Sigma_k$ represent the parameters of the $k$-th Gaussian component. 
The latent embeddings contain the respective non-parametric properties from the origin, which can be described by the DPM. 
This weighted sum helps to reduce assignment errors by aligning the embeddings with DPM components.

After updating the DPM parameters, we fix the mixture prior and optimize the neural networks.
% A batch of temporally extended skills with horizon length $L$ is fed into the skill posterior encoder $q(z|\bm{a}_i)$, generating embeddings in the latent space.
A batch of temporally extended skill sequences with horizon length $L$ is encoded by the skill posterior encoder $q(z|\bm{a}_i)$ producing latent embeddings in the skill space. 
The skill decoder $p(\bm{a}_i|z,s_t)$ then reconstructs the action sequence conditioned on both the latent embedding and the initial state $s_t$. 
The reconstruction loss between the generated actions and ground-truth actions is used to update the encoder–decoder parameters.
To ensure consistency between the inferred embeddings and the adaptive skill prior, we simultaneously learn a state-conditioned prior $p(z|s_t)$.
A KL divergence term $\mathbb{KL}(q(z|\bm{a}_i) || p_\vartheta(z|s_t))$ regularizes the posterior towards the predicted prior, encouraging alignment between the learned skill representations and the non-parametric mixture structure. 
This regularization stabilizes training and promotes structured latent skill organization for downstream RL.

The overall objective in this phase is to capture the non-parametric nature of skills, allowing the model to represent an unknown number of skill features that can adapt based on observed data. The total skill prior learning objective includes three components:
\begin{align}
    \mathcal{L}_{total} &= \zeta_1 \mathcal{L}_{rec}(\bm{a}_i, \hat{\bm{a}}_i) \nonumber\\
                        &\quad + \zeta_2 \sum_{k=1}^K p_{ik} \mathcal{L}_{KLD}(q(z|\bm{a}_i), \mathcal{N}(\mu_k, \Sigma_k)) \nonumber\\
                        &\quad + \zeta_3 \mathcal{L}_{KLD} (q(z|\bm{a}_i), p_\vartheta(z|s_t)) \text{,}
\end{align}
where $\{\zeta_i\}_{i=1,2,3}$ are the respective weighting factors, $\mathcal{L}_{rec}(\bm{a}_i, \hat{\bm{a}}_i)$ represents the mean squared reconstruction error between generated actions $\hat{a}_i$ and ground truth $a_i$, $\mathcal{L}_{KLD}$ is the KL divergence between two distributions.
This pretraining phase aims to build a flexible skill prior that can represent diverse actions, providing a foundation for the downstream RL policy to focus on meaningful action patterns without having to learn from scratch.

\subsection{Downstream Long-Horizon RL with Skill Priors}\label{sec:rl_method}
In the downstream task learning phase, as illustrated on the right side of Fig.\ref{fig:framework} (highlighted in yellow), we use a hierarchical policy structure. The upstream inference module, implemented with the state-of-the-art (SOTA) Soft Actor-Critic (SAC) framework, learns to produce latent skill embeddings $z$ based on observations and the pre-trained non-parametric skill prior. For the downstream component, we integrate the pre-trained skill decoder $p(\bm{a}_i|z, s_t)$ from Phase I, with its parameters kept fixed. The decoder takes in the latent embeddings and outputs a sequence of actions $\{a_t, a_{t+1}, \cdots, a_{t+L}\} \sim p(\bm{a}_i | z, s_t)$ of length $L$, continuing until the next state update.
Thus, downstream task learning can be formulated as a standard MDP problem with temporal abstraction. Following Sutton's temporal abstraction framework \cite{sutton1999between}, we replace single-step rewards $r$ with $L$-step cumulative rewards $\widetilde{r}=\sum_{t=1}^L r_t$, and state transition with $L$-step transitions $s_{t+L} \sim p_\varphi(s_{t+L}| s_t, z_t)$.

\begin{figure}[t!]
% \vskip -0.2in
    \centering
    \resizebox{.45\textwidth}{!}{
    \begin{tikzpicture}[font=\sffamily]
        \node[]at(0,0){\includegraphics[width=.5\textwidth]{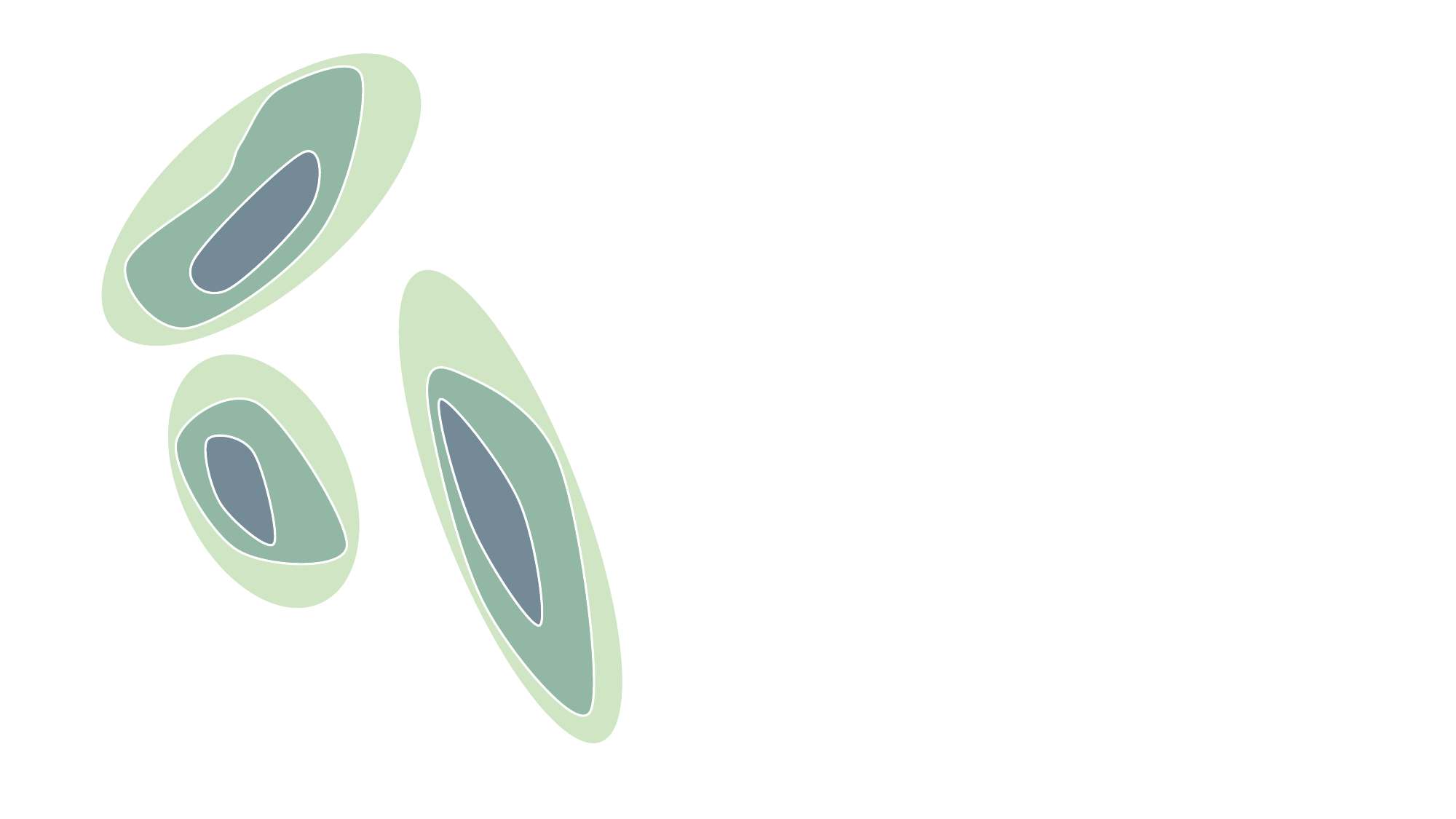}};
        \node[text width=9em](pzs)at(2.0,1){\small \textsf{Skill Prior $p(z|s_t)$}};
        \node[text width=9em](pza)at(2.0,0){\small \textsf{Skill Posterior $p(z|\bm{a}_i)$}};
        \node[text width=9em](pz)at(2.0,-1){\small \textsf{DPM-based Prior $p(z)$}};
        \draw[line width=1pt] (pz.west) -- ++ (-3em, -1em);
        \draw[line width=1pt] (pza.west) -- ++ (-4.8em, -1.5em);
        \draw[line width=1pt] (pzs.west) -- ++ (-4.5em, -3.2em);
        \node[text width=3em]at(-1,1.8){\small \textsf{Skill $1$}};
        \node[text width=3em]at(-2.5,-1.5){\small \textsf{Skill $2$}};
        \node[text width=8em]at(1,-1.8){\small \textsf{Skill $i$, with $i<\infty$}};
    \end{tikzpicture}
    }
    \vskip -0.25in
    \caption{\textbf{Skill distribution assumption of our proposed framework.}}
    \vskip -0.2in
    \label{fig:distribution}
\end{figure}

When learning from a large dataset containing a diverse set of primitive skills, the policy may face exploration challenges due to the rich skill space. To address this, we use the pre-trained Bayesian non-parametric skill prior to guide the upstream inference module, improving exploration efficiency. This skill prior can be naturally incorporated into a maximum-entropy RL framework.
Given Eq.\eqref{eq:sac_entropy}, the policy entropy term is effectively equivalent to the negative KL divergence between the policy distribution and our pre-trained skill prior, up to a constant. To guide the upstream inference module's exploration in skill space, we replace the traditional entropy term with this KL divergence, which encourages alignment with the skill prior distribution. 
The objective for the upstream module is defined as:
\begin{equation}
\resizebox{.48\textwidth}{!}{$
\begin{aligned}
    J(\varphi) = \mathbb{E}_\pi\Bigg[ &\sum_{t=1}^T  \widetilde{r}(s_t, z_t) - \omega \mathbb{KL}(\pi_\varphi(z_t|s_t), p(z_t|s_t)) \Bigg] ~\text{with} \\
    H(\pi_\varphi(\cdot|s_t)) &= - \mathbb{E}_\pi[log\pi(\cdot|s_t)] \propto - \mathbb{KL}(\pi_\varphi(\cdot|s_t), p(\cdot|s_t)) \text{,}
\end{aligned}
$}
\end{equation}
where $\varphi$ represents the parameters of the upstream module.
Additionally, Fig. \ref{fig:distribution} illustrates our proposed skill distributions to enable intuitive understanding.

%%%%%%%%%%%%%%%%%%%%%%%%%%%%%%%%%%%%%%%%%%%%%%%%%%
% -- experiments --
\section{EXPERIMENTS}
In this section, we present the experimental performance of our framework. We begin with a description of our experiment setup, followed by an evaluation of our Bayesian non-parametric skill prior in Phase I. Finally, we evaluate our agent's performance in downstream long-horizon RL tasks. Compared to SOTA baselines, our framework shows substantial improvements in accelerating downstream task exploration.
\subsection{Experiment Setup}
We evaluate our framework on Franka Kitchen \cite{fu2020d4rl}, LIBERO-Long \cite{liu2023libero}, Meta-World-MT10 \cite{yu2020meta}, and a real robotic platform. 
\textbf{Franka Kitchen} is used as a case study to analyze long-horizon skill composition. The environment contains seven manipulation subtasks (e.g., microwave, burners, cabinets). 
We use the ``mixed'' unstructured dataset solely for skill prior learning, which provides 400 teleoperated trajectories containing four primitive subtasks. 
In downstream RL, the agent must compose skills to solve long-horizon tasks with four sub-goals in unseen orders. 
Three additional subtasks are introduced to evaluate zero-shot generalization.
\textbf{LIBERO-Long} and \textbf{Meta-World} further evaluate scalability across tasks and embodiments. 
LIBERO-Long involves 10 multi-stage household manipulation tasks with varying object layouts. 
Meta-World provides a large set of robotic manipulation tasks, where we collect offline data for MT10 using scripted policies to pretrain the skill prior. 
All simulated benchmarks are conducted under \textbf{sparse reward settings}, where subtask success signals serve as the only reward signal (1 for completion, 0 otherwise). 
% This design emphasizes the difficulty of long-horizon credit assignment and exploration.
We report results over at least five random seeds and present the mean and standard deviation ($\mu \pm \sigma$).
For real-world validation, we deploy our method on a KUKA iiwa R800 manipulator equipped with a Robotiq 2F-85 gripper. 
Two Intel RealSense D456 cameras are mounted near the tabletop for perception. 
Object poses are estimated using SAM3 \cite{carion2025sam} combined with color filtering and marker-based calibration, enabling spatial reasoning for long-horizon manipulation tasks.

We compare HELIOS with standard RL and skill-based approaches: 
(1) \textbf{SAC} \cite{haarnoja2018soft}, trained from scratch; 
(2) \textbf{Behavior Cloning (BC)} with downstream finetuning; 
(3) \textbf{Skill-Space Policy (SSP)}, which learns directly in the skill embedding space without an explicit prior; 
(4) \textbf{Flat Prior}, a single-step action prior without temporal modeling; and 
(5) \textbf{SPIRL} \cite{pertsch2021accelerating}, which employs a fixed single-Gaussian skill prior. 
These comparisons isolate the impact of our adaptive Bayesian non-parametric skill prior.

\begin{figure}[t!]
    \centering
    \resizebox{.48\textwidth}{!}{
    \begin{minipage}{0.2\textwidth}
        \centering
        \begin{tikzpicture}[font=\sffamily]
            \begin{axis}[
              width=5cm, height=4.5cm,
              grid=major,
              xlabel={Epochs},
              ylabel={\# Clusters},
              xmin=-5, xmax=105,
              ymin=0, ymax=13,
            ]
            \pgfplotstableread[col sep=comma]{data/num_clusters_mean_std.csv}\datatable
            % Plotting data
            \addplot[fforange_pv!80, line width=1pt] table [x=epoch, y=num_cluster_mean] {\datatable};
            \addplot [name path=upper,draw=none, forget plot] table[x=epoch, y=num_cluster_upper] {\datatable};
            \addplot [name path=lower,draw=none, forget plot] table[x=epoch, y=num_cluster_lower] {\datatable};
            \addplot [fill=fforange_pv!50, forget plot, opacity=0.3] fill between[of=upper and lower];
            \end{axis}
            \node[]at(-0.75cm, 3cm){\small \textbf{\textsf{a}}};
        \end{tikzpicture}
    \end{minipage}
    ~~~~~~
    \begin{minipage}{0.2\textwidth}
        \centering
        \begin{tikzpicture}[font=\sffamily]
            \node[inner sep=0pt, anchor=center](img)at(0,0){\includegraphics[width=4cm]{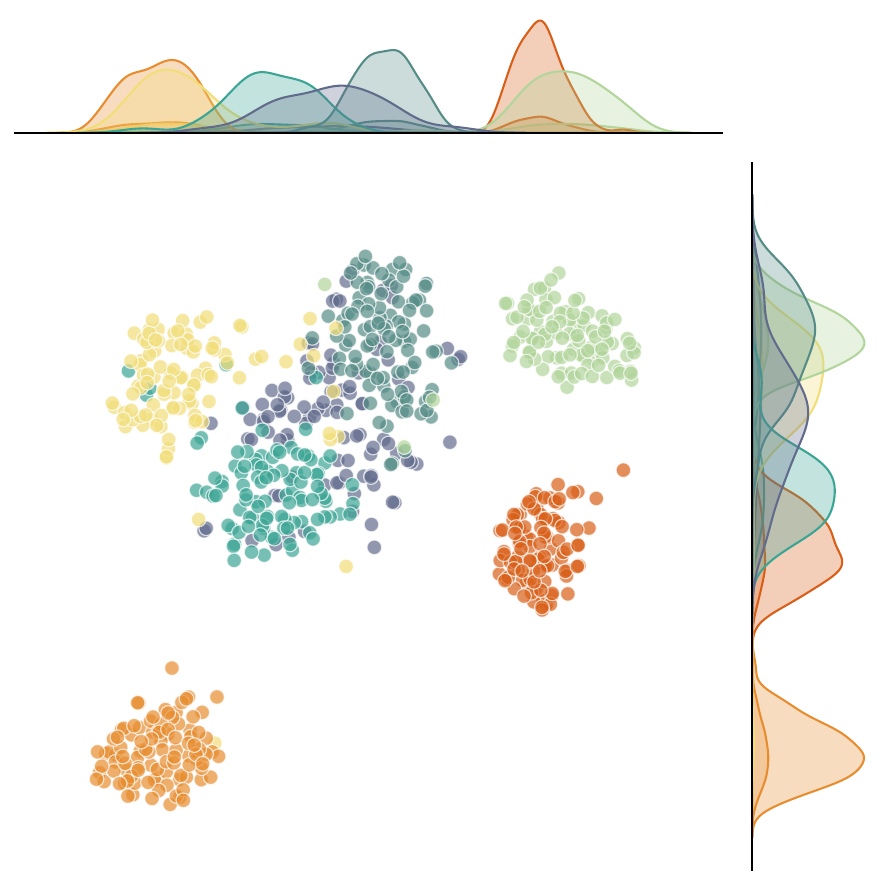}};
            \node[]at([yshift=.1cm]img.north west){\small \textbf{\textsf{b}}};
        \end{tikzpicture}
    \end{minipage}
    ~
    \begin{minipage}{0.2\textwidth}
        \centering
        \begin{tikzpicture}[font=\sffamily]
            \node[inner sep=0pt, anchor=center](img)at(0,0){\includegraphics[width=4cm]{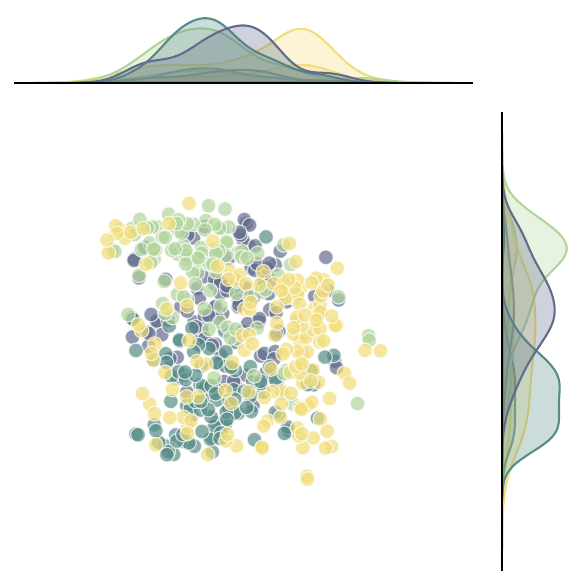}};
            \node[]at([yshift=.1cm]img.north west){\small \textbf{\textsf{c}}};
        \end{tikzpicture}
    \end{minipage}
    }
    \vskip -0.05in
    \caption{\textbf{Evaluation of Bayesian non-parametric skill prior in Franka Kitchen.} 
    % \textbf{\textsf{a}}, The total training loss ($\mathcal{L}_{total}$) observed during the skill prior pretraining phase. 
    \textbf{\textsf{a}}, The evolution of the number of generated clusters in the Bayesian non-parametric skill prior space throughout training. We conduct at least five trials and report the mean and standard deviation ($\mu\pm\sigma$). 
    \textbf{\textsf{b}}, t-SNE projection of the DPM-based skill prior.
    \textbf{\textsf{c}}, t-SNE projection of Gaussian-based skill prior (grounded by subtasks).
    }
    \vskip -0.2in
    \label{fig:skill_prior_results}
\end{figure}
% --------------------------------

\subsection{Representation Learning of Skill Prior}
We first train the proposed Bayesian non-parametric skill prior model in Franka Kitchen dataset as presented in Sec.~\ref{sec:skill_prior_method} Phase I. Fig.~\ref{fig:skill_prior_results}a illustrates the evolution of the number of components in the skill prior. 
Initially, the DPM starts with a single Gaussian component. 
After each epoch, online buffered data are used to adapt the number, shape, and density of components in DPM. 
In early stages, when both network parameters and mixture structure are not yet stabilized, extra components may appear, occasionally reaching up to 12. 
As training progresses and noise decreases, redundant components are merged, and the prior gradually stabilizes around 6–8 components. 
Fig.~\ref{fig:skill_prior_results}b shows the t-SNE projection of our learned non-parametric prior, where distinct clusters emerge, indicating that the DPM effectively identifies underlying skill modes and forms a structured latent space. 
In contrast, Fig.~\ref{fig:skill_prior_results}c presents a single-Gaussian prior implemented with a $\beta$-VAE. Since this simple parametric prior cannot explicitly model multiple skill modes, we analyze the embeddings using subtask labels as references. Although coarse separation appears due to $\beta$-VAE’s disentanglement effect, the latent space remains entangled at a finer level, limiting reusable skill representation and reducing efficiency in downstream long-horizon RL.
%

% -- average success --
\begin{figure}[b!]
\vskip -0.15in
    \centering
    \begin{tikzpicture}[font=\sffamily]
        \begin{axis}[
          width=8.2cm, height=5cm,
          grid=major,
          xlabel={Steps},
          ylabel={Average reward},
          xmin=0, xmax=1520000,
          ymin=-0.2, ymax=4.2,
          xtick={0, 5e5, 1e6, 1.5e6},
          ytick={0,1,2,3,4},
          legend pos=south east,
          legend style = { 
                yshift=-2cm,
                draw=gray, 
                fill=white,
                fill opacity=0.5,
                text opacity=1,
                rounded corners=0.5em,
                align=left, 
                text width=38pt,
                legend columns=3, 
                font=\tiny\sffamily,
            }
        ]
        \pgfplotstableread[col sep=comma]{data/helios_rew_v2.csv}\datatable
        % bc
        \addplot [ffblue!80, line width=1pt] table [x=step, y=bc_mean] {\datatable};\addlegendentry{BC}
        \addplot [name path=upper,draw=none, forget plot] table[x=step, y=bc_upper] {\datatable};
        \addplot [name path=lower,draw=none, forget plot] table[x=step, y=bc_lower] {\datatable};
        \addplot [fill=ffblue!50, forget plot, opacity=0.3] fill between[of=upper and lower];
        % ssp
        \addplot [ffdarkgreen!80, line width=1pt] table [x=step, y=ssp_mean] {\datatable};\addlegendentry{SSP}
        \addplot [name path=upper,draw=none, forget plot] table[x=step, y=ssp_upper] {\datatable};
        \addplot [name path=lower,draw=none, forget plot] table[x=step, y=ssp_lower] {\datatable};
        \addplot [fill=ffdarkgreen!50, forget plot, opacity=0.3] fill between[of=upper and lower];
        % sac
        \addplot [ffyellow, line width=1pt] table [x=step, y=sac_mean] {\datatable};\addlegendentry{SAC}
        \addplot [name path=upper,draw=none, forget plot] table[x=step, y=sac_upper] {\datatable};
        \addplot [name path=lower,draw=none, forget plot] table[x=step, y=sac_lower] {\datatable};
        \addplot [fill=ffyellow!50, forget plot, opacity=0.3] fill between[of=upper and lower];
        % flat
        \addplot [ffred!80, line width=1pt] table [x=step, y=flat_mean] {\datatable};\addlegendentry{Flat Prior}
        \addplot [name path=upper,draw=none, forget plot] table[x=step, y=flat_upper] {\datatable};
        \addplot [name path=lower,draw=none, forget plot] table[x=step, y=flat_lower] {\datatable};
        \addplot [fill=ffred!50, forget plot, opacity=0.3] fill between[of=upper and lower];
        % spirl
        \addplot [fforange_pv!60, line width=1pt] table [x=step, y=spirl_mean] {\datatable};\addlegendentry{SPIRL}
        \addplot [name path=upper,draw=none, forget plot] table[x=step, y=spirl_upper] {\datatable};
        \addplot [name path=lower,draw=none, forget plot] table[x=step, y=spirl_lower] {\datatable};
        \addplot [fill=fforange_pv!40, forget plot, opacity=0.3] fill between[of=upper and lower];
        % helios
        \addplot [ffgreen_pv!80, line width=1pt] table [x=step, y=mean] {\datatable};\addlegendentry{\textbf{HELIOS (Ours)}}
        \addplot [name path=upper,draw=none, forget plot] table[x=step, y=upper] {\datatable};
        \addplot [name path=lower,draw=none, forget plot] table[x=step, y=lower] {\datatable};
        \addplot [fill=ffgreen_pv!50, forget plot, opacity=0.3] fill between[of=upper and lower];
        
        \end{axis}
        % \node[]at(-1, 3){\textbf{\textsf{a}}};
    \end{tikzpicture}
    \vskip -0.1in
    \caption{\textbf{Average reward of long-horizon manipulation task in Franka Kitchen.} In the Benchmark, we use sparse rewards to train the agent, awarding a score of 1 only for successfully completed subtasks and 0 otherwise. For each model, we run at least five trials, reporting the average reward and standard deviation ($\mu\pm\sigma$) for comparison.}
    % \vskip -0.2in
    \label{fig:success}
\end{figure}
%
%
% ---------------------
% -- subtasks snapshots --
\begin{figure*}[ht!]
% \vskip -0.1in
    \centering
    \resizebox{.95\textwidth}{!}{
    \begin{tikzpicture}[font=\sffamily]
        % background
        \pgfdeclarehorizontalshading{bgshading}{100bp}
            {color(0bp)=(ffblue); 
            color(10bp)=(ffblue!60);
            color(30bp)=(ffdarkgreen!60);
            color(50bp)=(fflightgreen!60); 
            color(70bp)=(ffyellow!60);
            color(90bp)=(ffred!60);
            color(100bp)=(ffred_pv)
            }
        % \draw[path picture={\fill [shading=bgshading] (path picture bounding box.south west) rectangle (path picture bounding box.north east);}, rounded corners=4pt, draw=none] 
        % (-1.6cm, -2cm) -- (-1.6cm, 1.6cm) -- (22.9cm, 1.6cm) -- (22.9cm,-2cm) -- cycle;
        \draw[path picture={\fill [shading=bgshading] (path picture bounding box.south west) rectangle (path picture bounding box.north east);}, rounded corners=4pt, draw=none] 
        (-1.6cm, 1.8cm) -- (-1.6cm, 5.4cm) -- (22.9cm, 5.4cm) -- (22.9cm,1.8cm) -- cycle;
        % snapshots
        \node[inner sep=-1pt, rectangle, rounded corners=4pt, clip, draw=none, fill=none](1)at(0,3.8cm){\includegraphics[width=3cm]{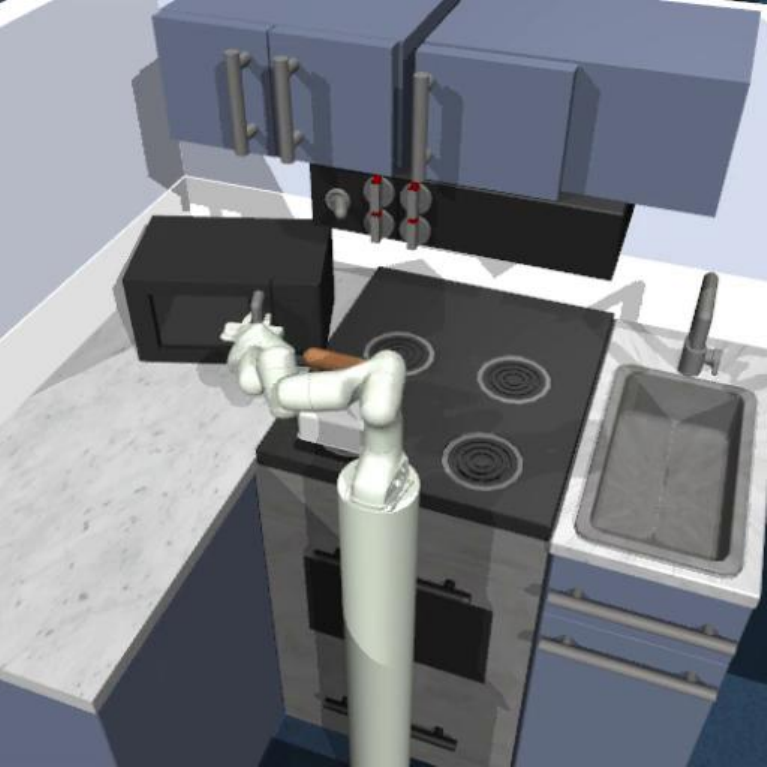}};
        \node[inner sep=-1pt, rectangle, rounded corners=4pt, clip, draw=none, fill=none, right](2)at([xshift=0.1cm]1.east){\includegraphics[width=3cm]{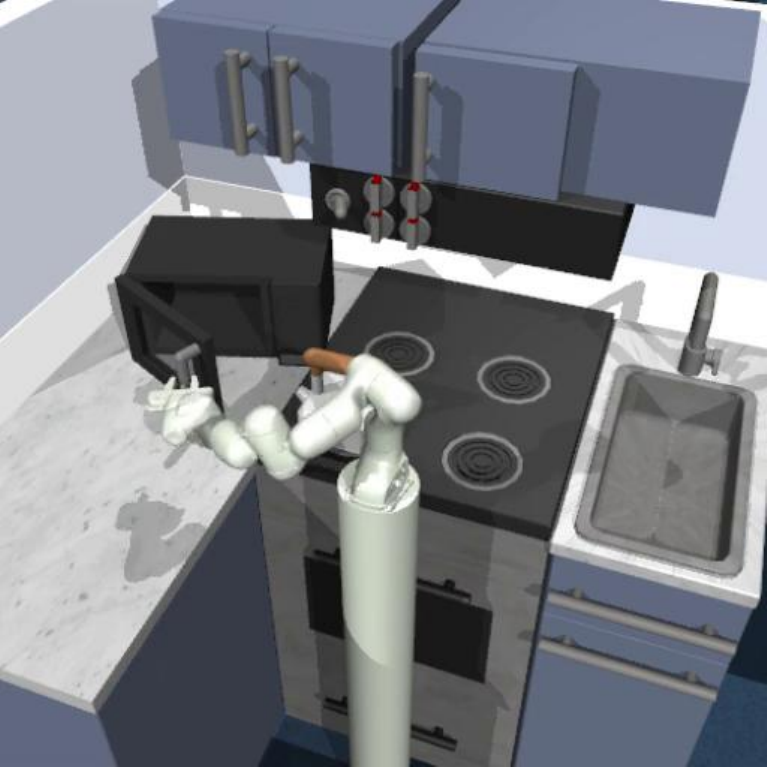}};
        \node[inner sep=-1pt, rectangle, rounded corners=4pt, clip, draw=none, fill=none, right](3)at([xshift=0.1cm]2.east){\includegraphics[width=3cm]{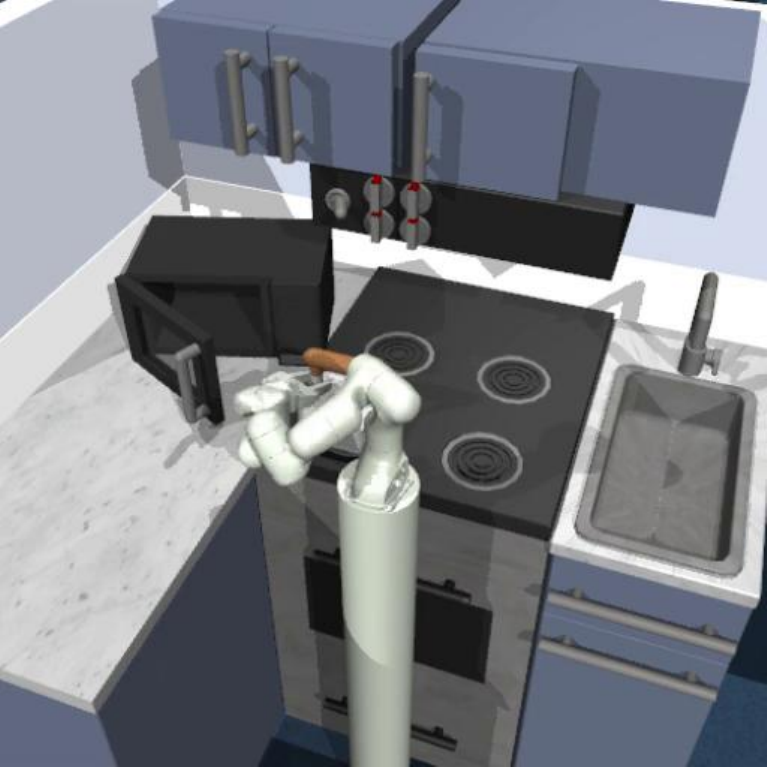}};
        \node[inner sep=-1pt, rectangle, rounded corners=4pt, clip, draw=none, fill=none, right](4)at([xshift=0.1cm]3.east){\includegraphics[width=3cm]{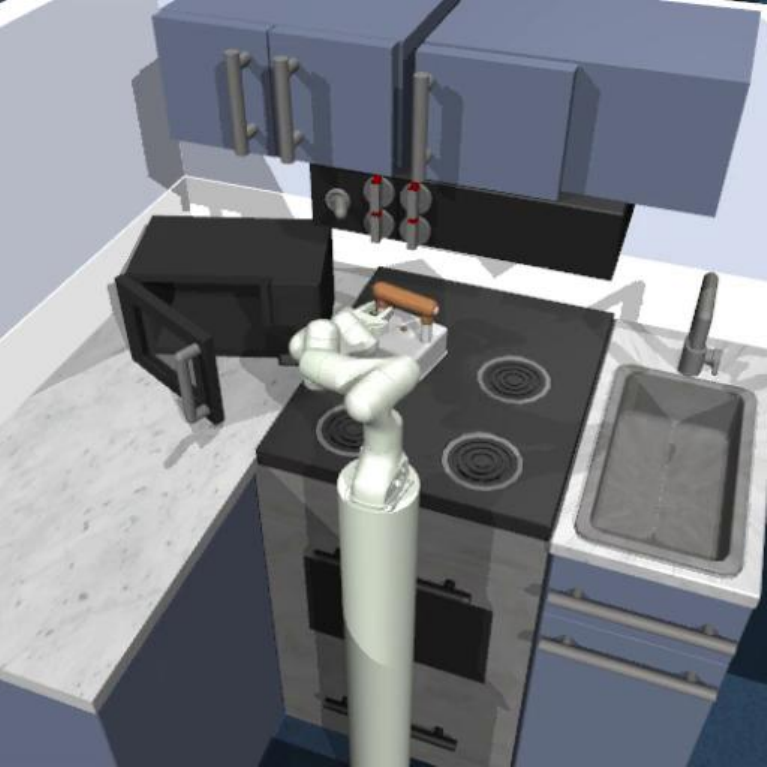}};
        \node[inner sep=-1pt, rectangle, rounded corners=4pt, clip, draw=none, fill=none, right](5)at([xshift=0.1cm]4.east){\includegraphics[width=3cm]{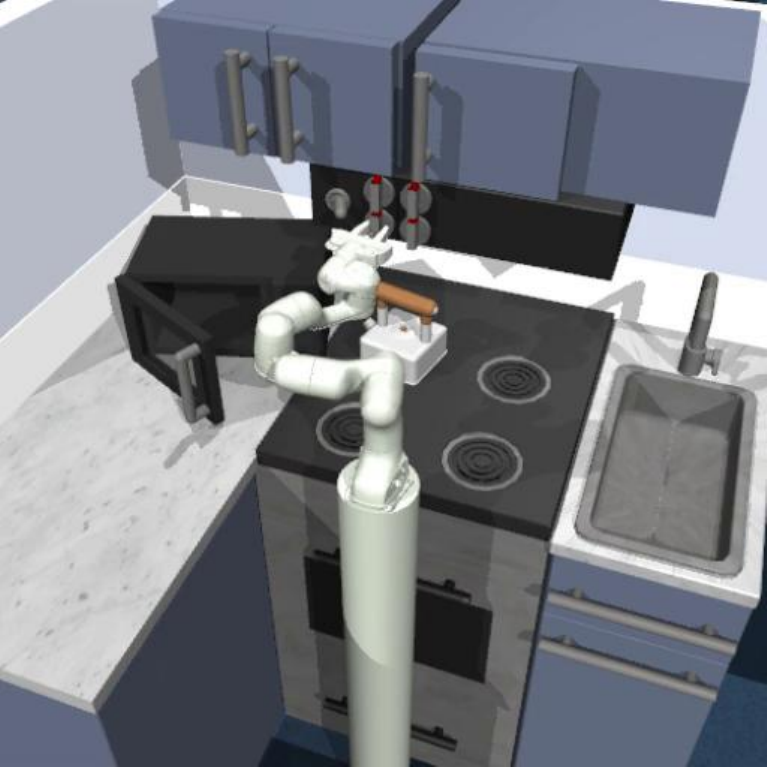}};
        \node[inner sep=-1pt, rectangle, rounded corners=4pt, clip, draw=none, fill=none, right](6)at([xshift=0.1cm]5.east){\includegraphics[width=3cm]{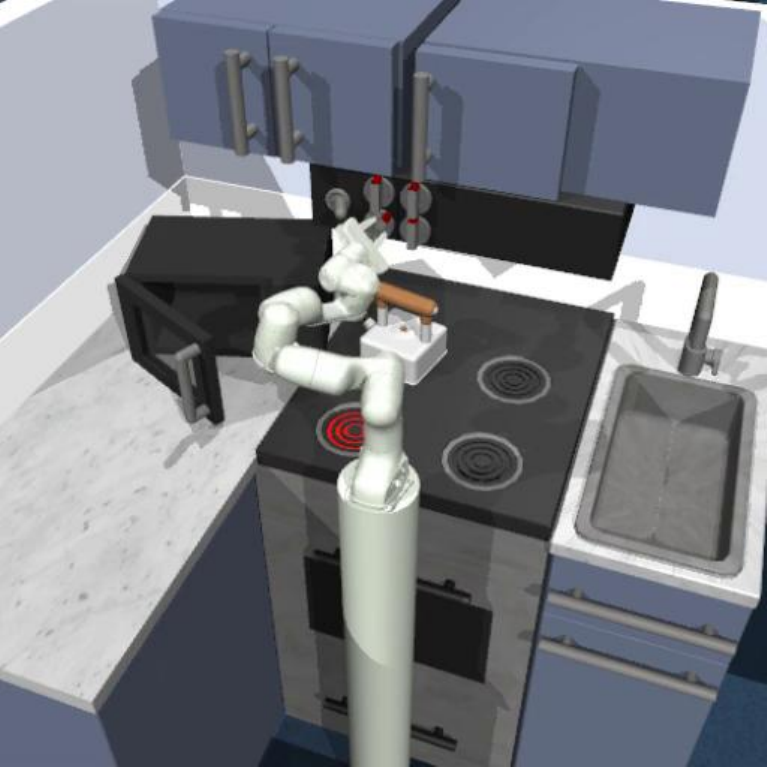}};
        \node[inner sep=-1pt, rectangle, rounded corners=4pt, clip, draw=none, fill=none, right](7)at([xshift=0.1cm]6.east){\includegraphics[width=3cm]{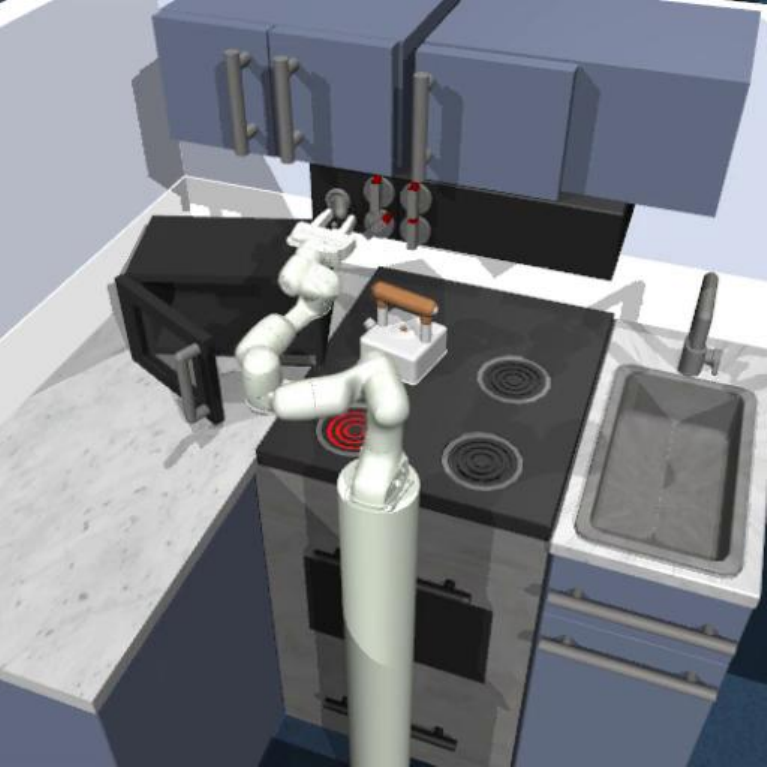}};
        \node[inner sep=-1pt, rectangle, rounded corners=4pt, clip, draw=none, fill=none, right](8)at([xshift=0.1cm]7.east){\includegraphics[width=3cm]{imgs/train8.pdf}};
        \node[below]at(1.south east)(a1){\textbf{\textsf{Microwave}}};
        \node[below]at(3.south east)(a2){\textbf{\textsf{Kettle}}};
        \node[below]at(5.south east)(a3){\textbf{\textsf{Burner}}};
        \node[below]at(7.south east)(a4){\textbf{\textsf{Light}}};
        \draw[-, draw=black, line width=1mm] ([xshift=-1.7cm]a1.west) -- (a1.west);
        \draw[-{Triangle Cap []. Fast Triangle[] Fast Triangle[]}, draw=black, line width=1mm] (a1.east) -- ([xshift=1.7cm]a1.east);
        \draw[-, draw=black, line width=1mm] ([xshift=-2.1cm]a2.west) -- (a2.west);
        \draw[-{Triangle Cap []. Fast Triangle[] Fast Triangle[]}, draw=black, line width=1mm] (a2.east) -- ([xshift=2.1cm]a2.east);
        \draw[-, draw=black, line width=1mm] ([xshift=-2cm]a3.west) -- (a3.west);
        \draw[-{Triangle Cap []. Fast Triangle[] Fast Triangle[]}, draw=black, line width=1mm] (a3.east) -- ([xshift=2cm]a3.east);
        \draw[-, draw=black, line width=1mm] ([xshift=-2.2cm, yshift=0.05cm]a4.west) -- ([yshift=0.05cm]a4.west);
        \draw[-{Triangle Cap []. Fast Triangle[] Fast Triangle[]}, draw=black, line width=1mm] ([yshift=0.05cm]a4.east) -- ([xshift=2.2cm, yshift=0.05cm]a4.east);

        % skill-tsne comparison
        \node[inner sep=-1pt, rectangle, rounded corners=4pt, clip, draw=none, fill=none, below](pick)at([yshift=-0.8cm]2.south){\includegraphics[width=3cm]{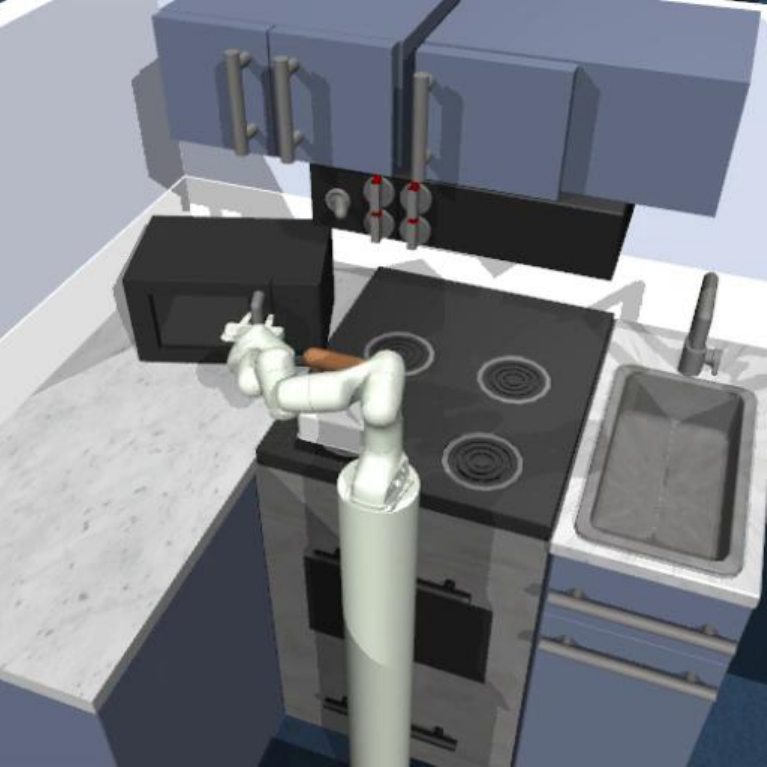}};
        \node[inner sep=-1pt, rectangle, rounded corners=4pt, clip, draw=none, fill=none, right](rr)at([xshift=0.1cm]pick.east){\includegraphics[width=3cm]{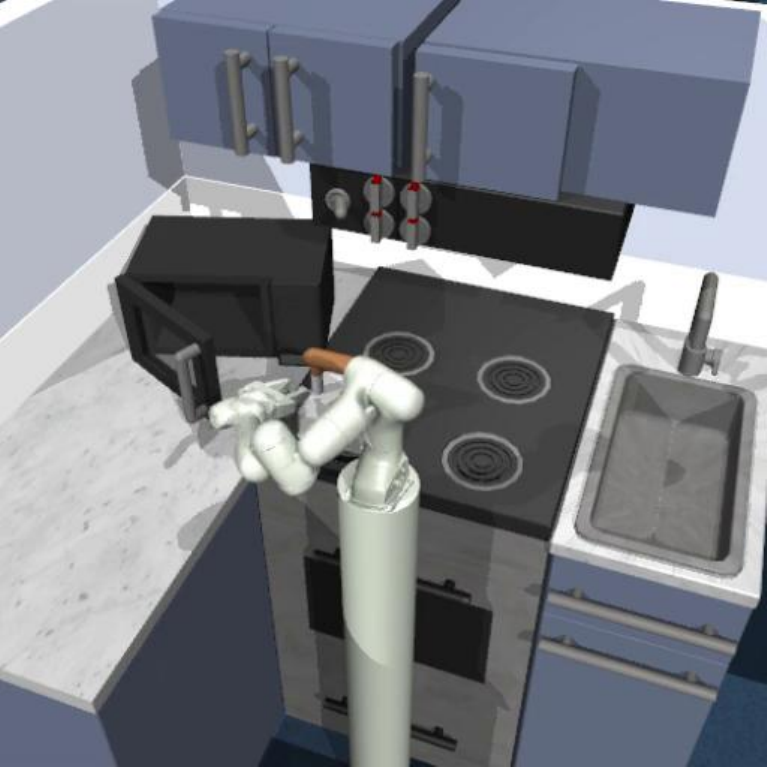}};
        \node[inner sep=-1pt, rectangle, rounded corners=4pt, clip, draw=none, fill=none, right](place)at([xshift=0.1cm]rr.east){\includegraphics[width=3cm]{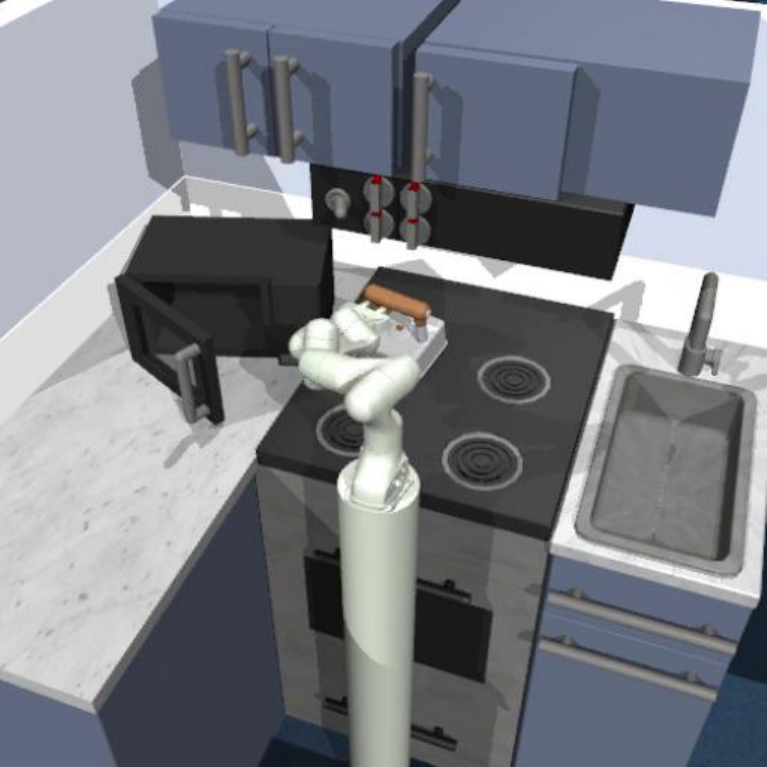}};
        \node[inner sep=-1pt, rectangle, rounded corners=4pt, clip, draw=none, fill=none, right](pull)at([xshift=0.1cm]place.east){\includegraphics[width=3cm]{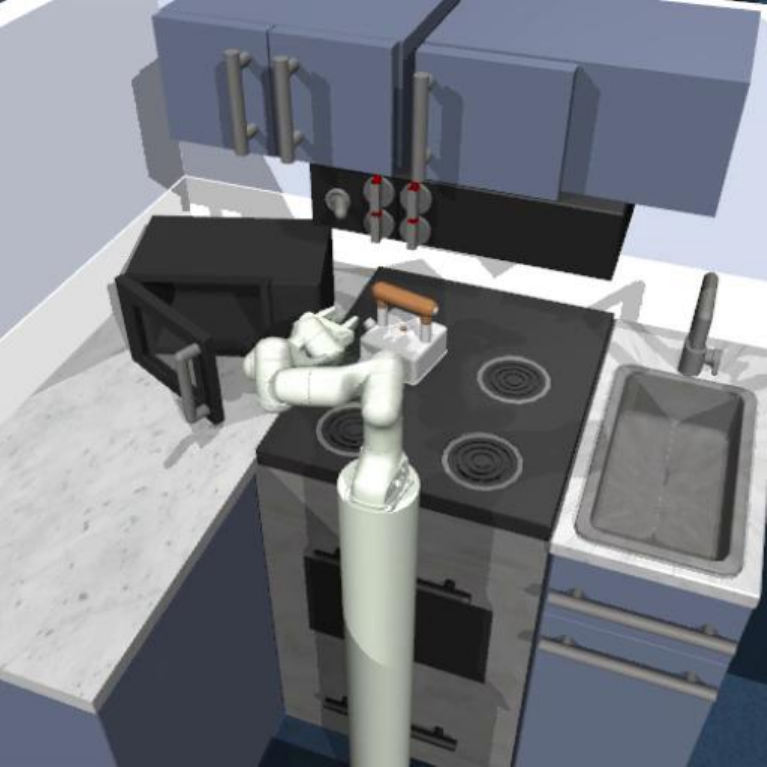}};
        \node[inner sep=-1pt, rectangle, rounded corners=4pt, clip, draw=none, fill=none, right](lr)at([xshift=0.1cm]pull.east){\includegraphics[width=3cm]{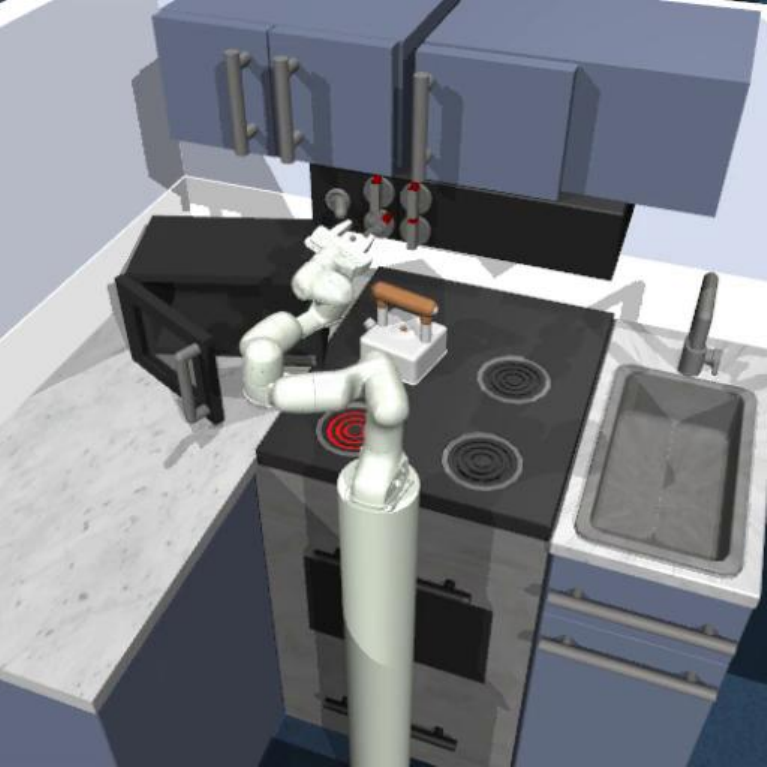}};
        \node[inner sep=-1pt, rectangle, rounded corners=4pt, clip, draw=none, fill=none, right](rot)at([xshift=0.1cm]lr.east){\includegraphics[width=3cm]{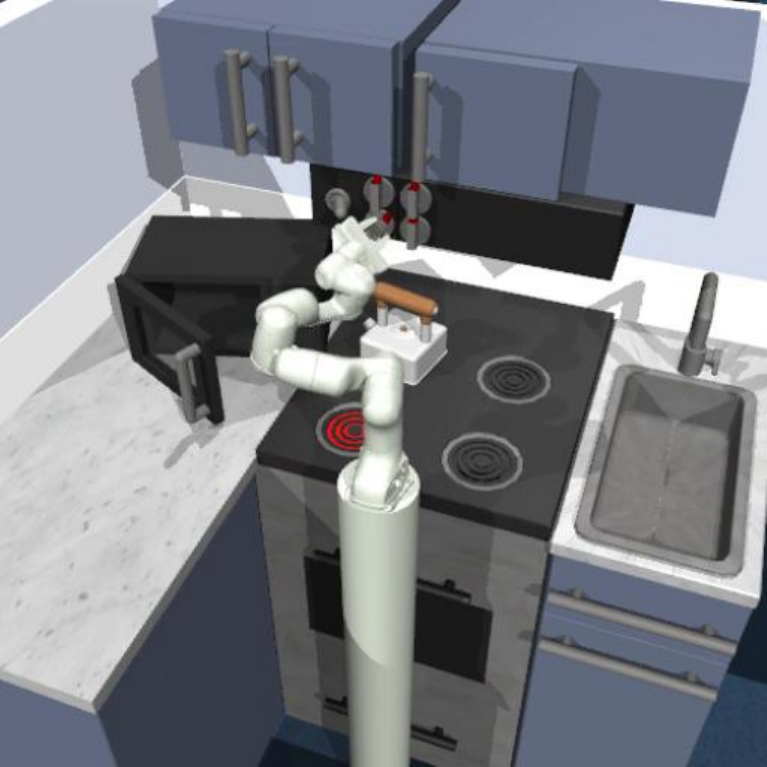}};
        \node[inner sep=-1pt, rectangle, rounded corners=4pt, clip, draw=none, fill=none, right](strum)at([xshift=0.1cm]rot.east){\includegraphics[width=3cm]{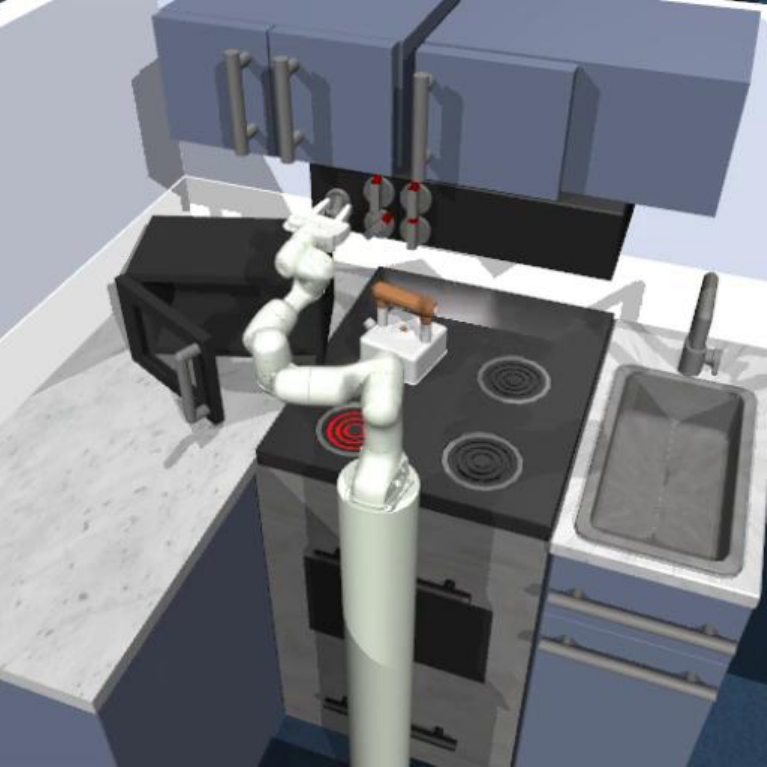}};

        % -- arrows --
        % pick
        \draw[-{Latex[width=7pt, length=4pt]}, draw=fforange_pv, line width=2pt] (2.0, -0.1) to[out=60, in=-150] (2.4, 0.3) {};
        \draw[-{Latex[width=7pt, length=4pt]}, draw=fforange_pv, line width=2pt] (3.1, -0.1) to[out=120, in=-30] (2.7, 0.3) {};
        % right reach
        \draw[-{Latex[width=7pt, length=4pt]}, draw=fforange_pv, line width=2pt] (5.0, 0) to[out=60, in=-150] (6, 0.4) {};
        % place
        \draw[-{Latex[width=7pt, length=4pt]}, draw=fforange_pv, line width=2pt] (9.5, 0.6) -- (9.5, 0) {};
        % pull
        \draw[-{Latex[width=7pt, length=4pt]}, draw=fforange_pv, line width=2pt] (12.5, 0.0) to[out=-140, in=30] (11.5, -0.6) {};
        % left reach
        \draw[-{Latex[width=7pt, length=4pt]}, draw=fforange_pv, line width=2pt] (15.5, 0) to[out=180, in=0] (14.5, 0.25) {};
        % rotate
        \draw[-{Latex[width=7pt, length=4pt]}, draw=fforange_pv, line width=2pt] (18, 0.9) to[bend left=60] (18.5, 0.5) {};
        % toggle
        \draw[-{Latex[width=7pt, length=4pt]}, draw=fforange_pv, line width=2pt] (21.25, 0.6) to[bend left=45] (20.75, 0.6) {};

        \node[rectangle, clip, draw=none, fill=none, below](tsne1)at([yshift=-0.1cm]pick.south){\includegraphics[width=3cm]{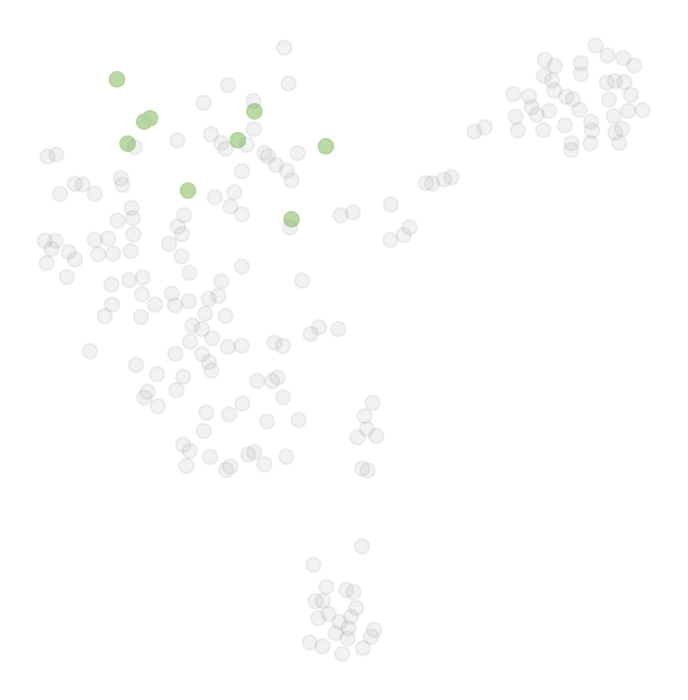}};
        \node[rectangle, clip, draw=none, fill=none, below]at([yshift=-0.1cm]rr.south){\includegraphics[width=3cm]{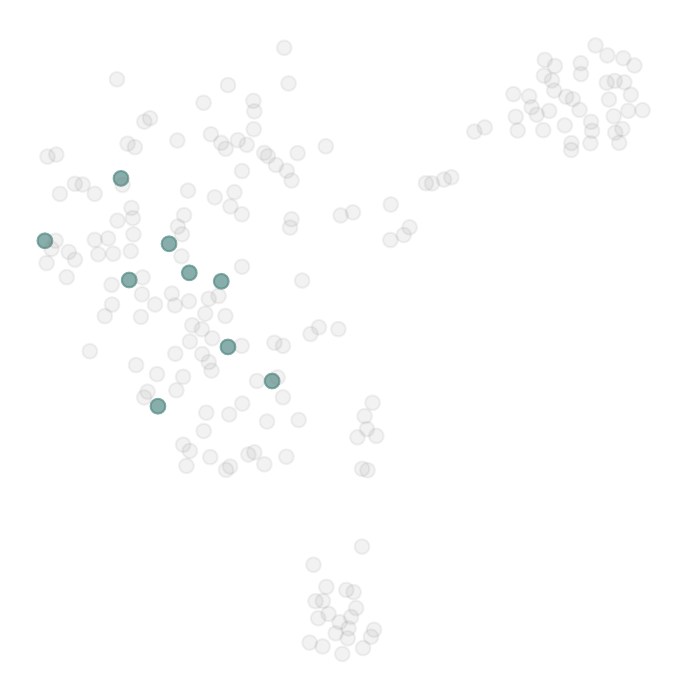}};
        \node[rectangle, clip, draw=none, fill=none, below]at([yshift=-0.1cm]place.south){\includegraphics[width=3cm]{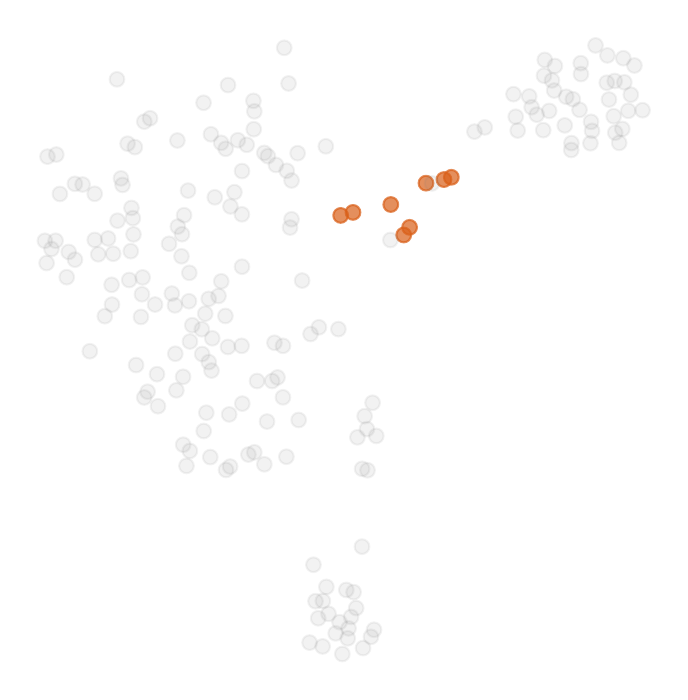}};
        \node[rectangle, clip, draw=none, fill=none, below]at([yshift=-0.1cm]pull.south){\includegraphics[width=3cm]{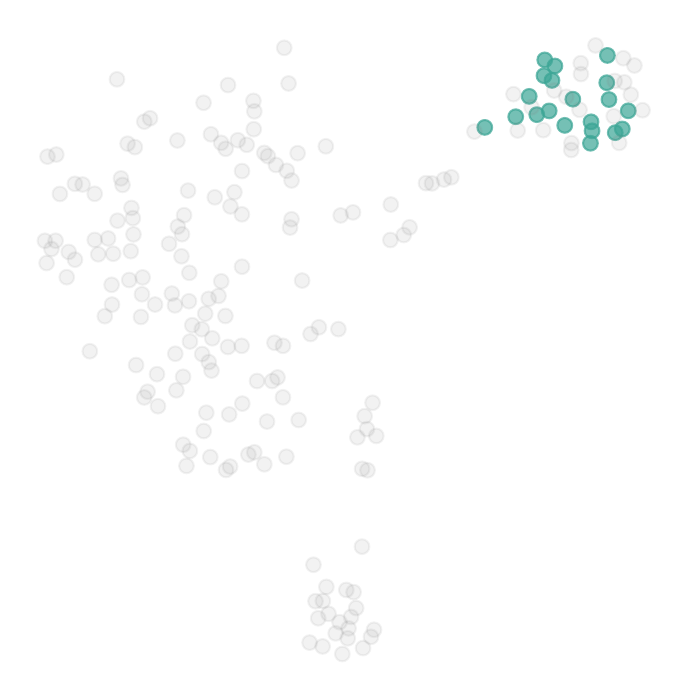}};
        \node[rectangle, clip, draw=none, fill=none, below]at([yshift=-0.1cm]lr.south){\includegraphics[width=3cm]{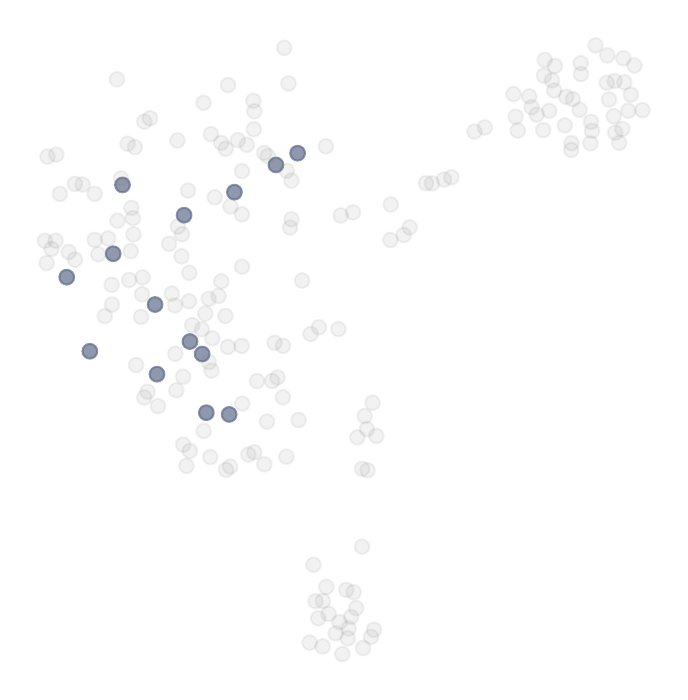}};
        \node[rectangle, clip, draw=none, fill=none, below]at([yshift=-0.1cm]rot.south){\includegraphics[width=3cm]{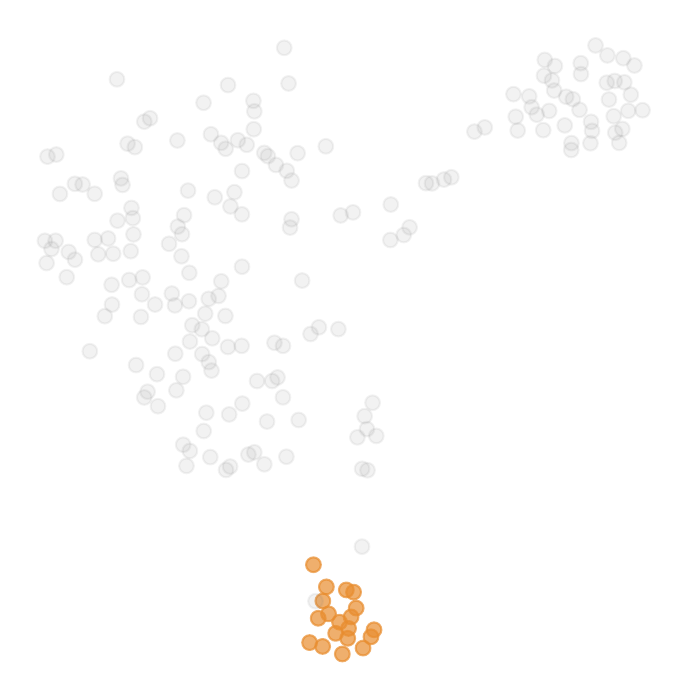}};
        \node[rectangle, clip, draw=none, fill=none, below]at([yshift=-0.1cm]strum.south){\includegraphics[width=3cm]{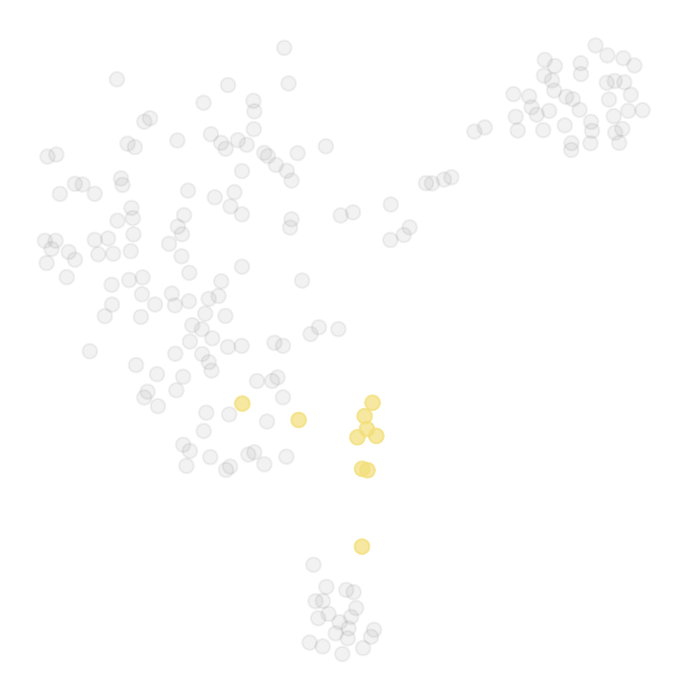}};
        \node[rectangle, clip, draw=none, fill=none, left]at([xshift=-0.1cm]tsne1.west){\includegraphics[width=3cm]{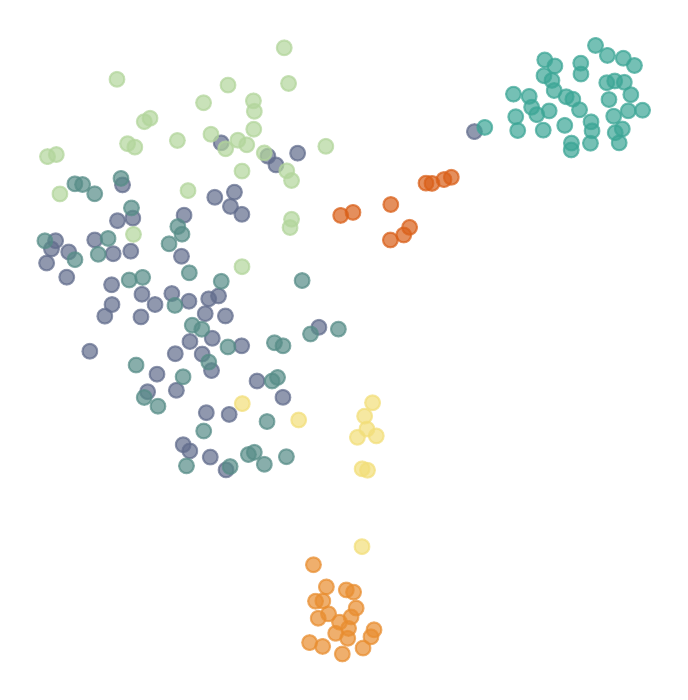}};

        \node[rectangle, rounded corners=4pt, fill=black, opacity=0.5, text opacity=1, above]at(pick.south){\textcolor{fflightgreen!80}{\textbf{\textsf{Pick}}}};
        \node[rectangle, rounded corners=4pt, fill=black, opacity=0.5, text opacity=1, above]at(rr.south){\textcolor{ffdarkgreen!80}{\textbf{\textsf{Right switch}}}};
        \node[rectangle, rounded corners=4pt, fill=black, opacity=0.5, text opacity=1, above]at(place.south){\textcolor{ffred!80}{\textbf{\textsf{Place}}}};
        \node[rectangle, rounded corners=4pt, fill=black, opacity=0.5, text opacity=1, above]at(pull.south){\textcolor{ffgreen_pv!80}{\textbf{\textsf{Pull}}}};
        \node[rectangle, rounded corners=4pt, fill=black, opacity=0.5, text opacity=1, above]at(lr.south){\textcolor{ffblue!80}{\textbf{\textsf{Left explore}}}};
        \node[rectangle, rounded corners=4pt, fill=black, opacity=0.5, text opacity=1, above]at(rot.south){\textcolor{fforange_pv!80}{\textbf{\textsf{Rotate}}}};
        \node[rectangle, rounded corners=4pt, fill=black, opacity=0.5, text opacity=1, above]at(strum.south){\textcolor{ffyellow!80}{\textbf{\textsf{Toggle}}}};

        % pie chart
        \filldraw[draw=white,fill=ffblue!60,] (0,0) -- (0:1.3cm) arc[start angle=0,end angle=93.996,radius=1.3cm] -- cycle;
        \filldraw[draw=white,fill=ffdarkgreen!60] (0,0) -- (93.996:1.3cm) arc[start angle=93.996,end angle=164.916,radius=1.3cm] -- cycle;
        \filldraw[draw=white,fill=ffgreen_pv!60] (0,0) -- (164.916:1.3cm) arc[start angle=164.916,end angle=235.836,radius=1.3cm] -- cycle;
        \filldraw[draw=white,fill=fflightgreen!60] (0,0) -- (235.836:1.3cm) arc[start angle=235.836,end angle=289.044,radius=1.3cm] -- cycle;
        \filldraw[draw=white,fill=ffyellow!60] (0,0) -- (289.044:1.3cm) arc[start angle=289.044,end angle=306.792,radius=1.3cm] -- cycle;
        \filldraw[draw=white,fill=fforange_pv!60] (0,0) -- (306.792:1.3cm) arc[start angle=306.792,end angle=342.252,radius=1.3cm] -- cycle;
        \filldraw[draw=white,fill=ffred!60] (0,0) -- (342.252:1.3cm) arc[start angle=342.252,end angle=360,radius=1.3cm] -- cycle;
        \filldraw[color=white, fill=white](0,0) circle (0.3cm);

        \node[]at(0.6cm,0.6cm){\footnotesize $\%26.1$};
        \node[]at(-0.6cm,0.6cm){\footnotesize $\%19.7$};
        \node[]at(-0.8cm,-0.3cm){\footnotesize $\%19.7$};
        \node[]at(-0.15cm,-0.9cm){\footnotesize $\%14.8$};
        \node[]at(0.6cm,-1.0cm){\footnotesize $\%4.9$};
        \node[]at(0.8cm,-0.6cm){\footnotesize $\%9.9$};
        \node[]at(1.1cm,-0.2cm){\footnotesize $\%4.9$};

        % annotations
        \node[]at(-1.7cm, 5.5cm){\large \textsf{\textbf{a}}};
        \node[]at(-1.7cm, 1.5cm){\large \textsf{\textbf{b}}};
        \node[]at(1.4cm, 1.5cm){\large \textsf{\textbf{c}}};
        \node[]at(-1.7cm, -1.5cm){\large \textsf{\textbf{d}}};        
        
    \end{tikzpicture}
    }
    \vskip -0.1in
    \caption{\textbf{Case study: long-horizon manipulation performance in Franka Kitchen benchmark.} 
    \textbf{\textsf{a}}, Snapshots of each sub-task, demonstrating that the agent efficiently completes all assigned subtasks. 
    \textbf{\textsf{b}}, The average skill ratios applied across the entire task, based on the results from at least five trials. 
    \textbf{\textsf{c}}, Snapshots of each primitive skill motion. Our Bayesian non-parametric prior captures seven base skills: ``pick'', ``place'', ``pull'', ``rotate'', ``toggle'', and ``explore'' movements (both left and right directions).
    \textbf{\textsf{d}}, t-SNE projections of each primitive skill motion, showing the re-encoded skills through the pre-trained skill encoder $q(z|\bm{a}_i)$ and their corresponding assignments in the Bayesian non-parametric knowledge space.}
    \vskip -0.2in
    \label{fig:task_snapshots}
\end{figure*}

\subsection{Case study: how Bayesian non-parametric skill prior contributes to the RL improvement}
\paragraph{Empirical analysis} 
After RL training (Sec.~\ref{sec:rl_method} Phase II), we present average rewards for all models on Franka Kitchen in a case study (Fig.~\ref{fig:success}). Our proposed framework, HELIOS, substantially outperforms all baseline models in average reward. 
HELIOS can efficiently explore the state space and finally achieve a high average reward, stabilizing above 3.8 (with maximal 4.0), while other baselines either converge at lower reward values or exhibit slower progress. Notably, SPIRL performs better than some baselines due to its use of a skill prior, but it falls short compared to HELIOS, which leverages a more flexible Bayesian non-parametric skill prior modeled with DPM and MemoVB heuristics. This adaptive skill prior allows HELIOS to dynamically capture an optimal set of skills that are more expressive and well-suited for complex, long-horizon tasks. 
Although Flat Prior incorporates a trained skill prior module with a simple fully connected backbone, this design fails to effectively capture the underlying skill structure and context, leading to poor performance in downstream tasks.
This highlights the contribution of our proposed GRU-based skill prior network.
Without the guidance of skill prior, the SAC, BC, and SSP models struggle in this complex, sparse-reward, long-horizon task environment, achieving an average reward of less than 0.5. 
By utilizing the DPM-based skill prior, HELIOS benefits from efficient exploration and the ability to recombine learned skills in new ways, essential for tackling unseen task sequences in long-horizon scenarios. This structured and adaptable skill space enables faster and more stable convergence, highlighting HELIOS’s superior performance in the challenging Franka Kitchen environment.

% ------------------------
\paragraph{Representation traceability} 
Fig. \ref{fig:task_snapshots}a demonstrates the rendering snapshots of our proposed agent performing a sequence of subtasks in the Franka Kitchen environment, including manipulating the microwave, kettle, burner, and light switch. 
The images demonstrate that the agent completes all assigned subtasks in the scene, highlighting its ability to handle complex, multi-step, long-horizon tasks.
The renderings of these subtasks reflect the agent's capability to utilize the pre-trained Bayesian non-parametric skill prior for efficient skill selection and execution. This showcases HELIOS’s strength in combining learned skills to solve sequential subtasks in the long-horizon manipulation scene.

Fig. \ref{fig:task_snapshots}b illustrates the distribution of skill ratios utilized throughout the long-horizon task sequence, averaged over at least five trials and color-coded by respective skills. In our task dataset, the Bayesian non-parametric skill prior captures an average of seven primitive skill motions, with snapshots of each skill depicted in Figure \ref{fig:task_snapshots}c. These skills include ``pick'' (14.8\%), ``place'' (4.8\%), ``pull'' (19.7\%), ``rotate'' (9.9\%), ``toggle'' (4.9\%), and ``explore'' in both left (26.1\%) and right (19.7\%) directions. Together, these skills form the fundamental building blocks of the agent’s actions, enabling efficient recombination to perform long-horizon tasks.
By leveraging the pre-trained Bayesian non-parametric skill prior, the agent effectively utilizes these skills to manage complex object manipulations, showcasing the prior’s capability to encode and generalize key motion behaviors. 
Additionally, Fig. \ref{fig:task_snapshots}d presents the t-SNE projections of each primitive skill motion in the latent space, revealing well-defined clusters of re-encoded skills generated by the skill encoder $q(z|\bm{a}_i)$ from phase I. Each cluster corresponds to a distinct skill, highlighting the ability of the Bayesian non-parametric prior to group similar actions while maintaining clear boundaries between different skill types. 
This latent space demonstrates the prior’s strength in capturing complex, multi-peak skill distributions. 
Such precise clustering is essential for guiding the agent in downstream tasks, enabling efficient exploration and robust inference of action patterns.

\begin{figure}[ht!]
% \vskip -0.05in
    \centering
    \resizebox{.49\textwidth}{!}{
    \begin{tikzpicture}[font=\sffamily]
        % background
        \pgfdeclarehorizontalshading{bgshading}{100bp}
            {color(0bp)=(ffblue); 
            color(10bp)=(ffblue!60);
            color(30bp)=(ffdarkgreen!60);
            color(50bp)=(fflightgreen!60); 
            color(70bp)=(ffyellow!60);
            color(90bp)=(ffred!60);
            color(100bp)=(ffred_pv)
            }
        \draw[path picture={\fill [shading=bgshading] (path picture bounding box.south west) rectangle (path picture bounding box.north east);}, rounded corners=4pt, draw=none] 
        (-1.6cm, -2cm) -- (-1.6cm, 1.6cm) -- (10.5cm, 1.6cm) -- (10.5cm,-2cm) -- cycle;
        % snapshots
        \node[inner sep=-1pt, rectangle, rounded corners=4pt, clip, draw=none, fill=none](1)at(0,0){\includegraphics[width=3cm]{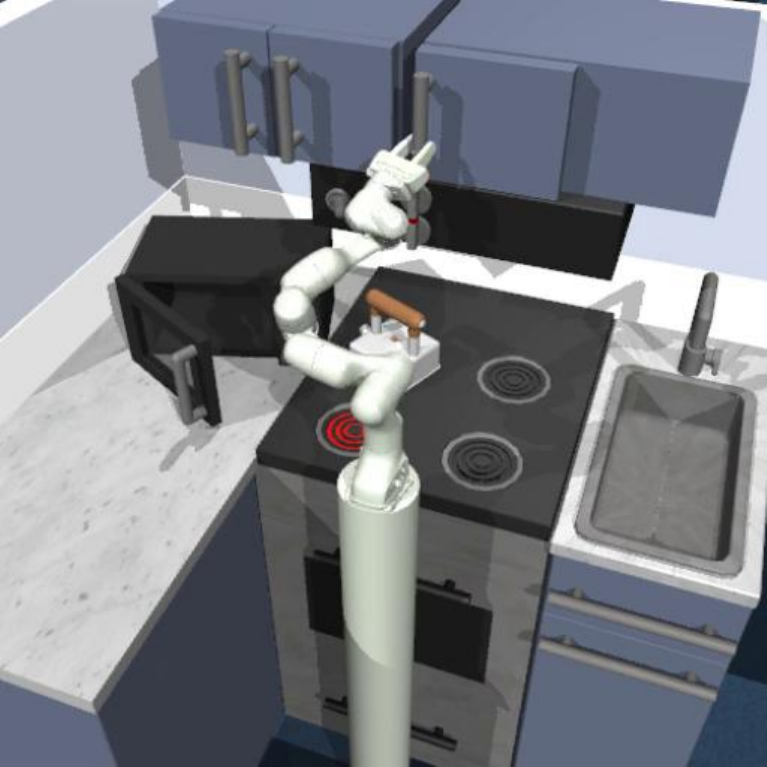}};
        \node[inner sep=-1pt, rectangle, rounded corners=4pt, clip, draw=none, fill=none, right](2)at(1.east){\includegraphics[width=3cm]{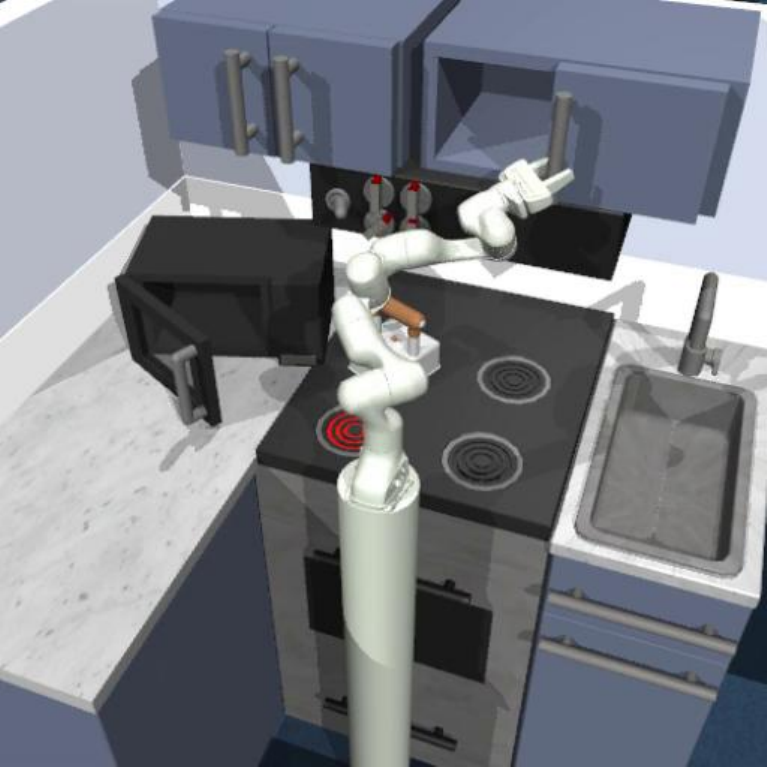}};
        \node[inner sep=-1pt, rectangle, rounded corners=4pt, clip, draw=none, fill=none, right](3)at([xshift=0.1cm]2.east){\includegraphics[width=3cm]{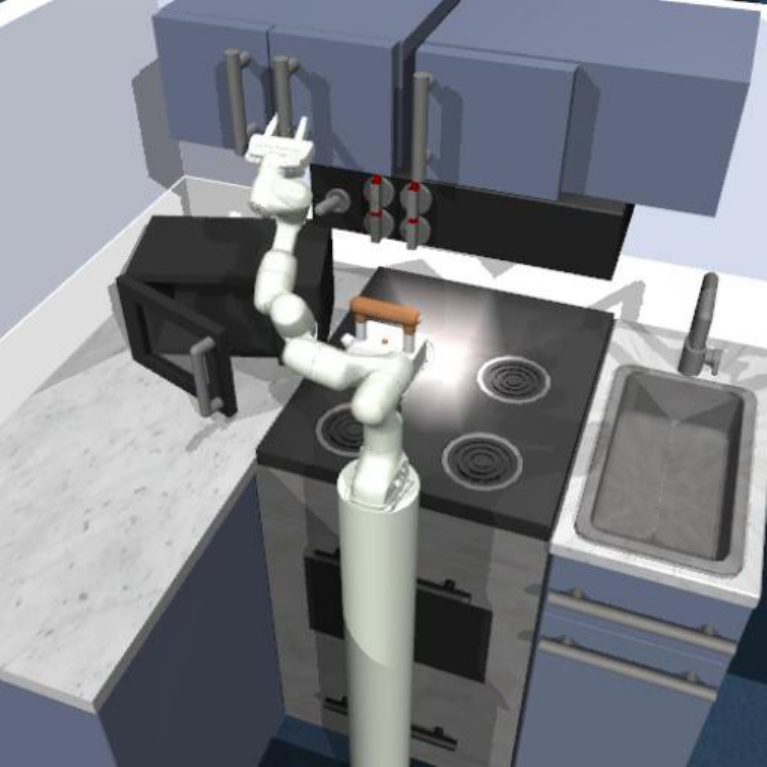}};
        \node[inner sep=-1pt, rectangle, rounded corners=4pt, clip, draw=none, fill=none, right](4)at(3.east){\includegraphics[width=3cm]{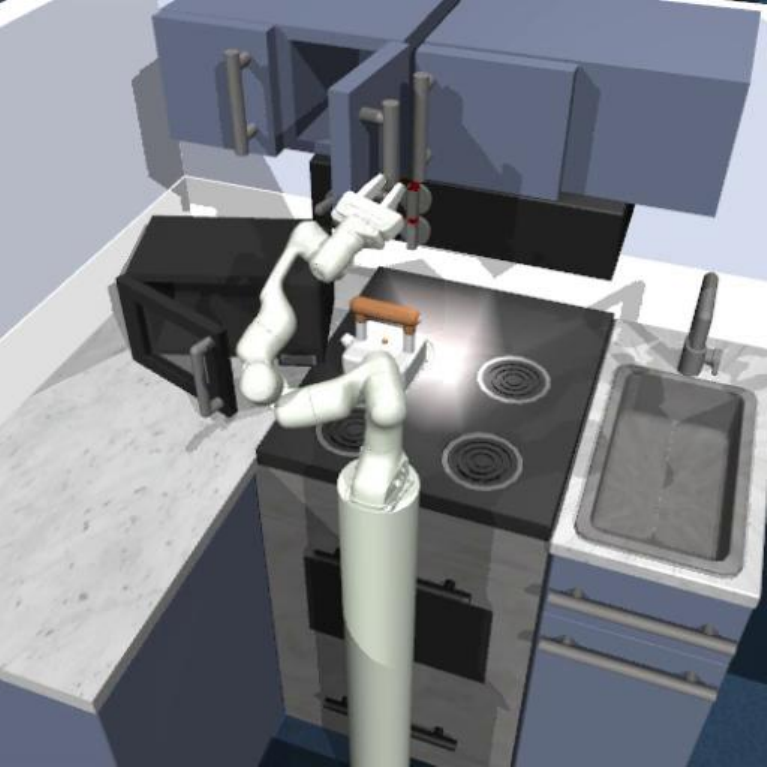}};
        \node[below]at(1.south east)(a1){\textbf{\textsf{Slider Cabinet}}};
        \node[below]at(3.south east)(a2){\textbf{\textsf{Hinge Cabinet}}};
        \draw[-, draw=black, line width=1mm] ([xshift=-1.5cm]a1.west) -- (a1.west);
        \draw[-{Triangle Cap []. Fast Triangle[] Fast Triangle[]}, draw=black, line width=1mm] (a1.east) -- ([xshift=1.5cm]a1.east);
        \draw[-, draw=black, line width=1mm] ([xshift=-1.5cm, yshift=0.05cm]a2.west) -- ([yshift=0.05cm]a2.west);
        \draw[-{Triangle Cap []. Fast Triangle[] Fast Triangle[]}, draw=black, line width=1mm] ([yshift=0.05cm]a2.east) -- ([xshift=1.5cm, yshift=0.05cm]a2.east);
    \end{tikzpicture}
    }
    \vskip -0.05in
    \caption{\textbf{Zero-shot skill adaptation for unseen sub-tasks.} 
    % Videos are available at \url{https://ghiara.github.io/HELIOS/}.
    }
    \vskip -0.1in
    \label{fig:zeroshot_task_snapshots}
\end{figure}

\paragraph{Skill adaptation}
Furthermore, we evaluate the zero-shot skill adaptation capability of our proposed framework in unseen task scenarios. To do this, we randomly incorporate the subtasks ``open slider cabinet'', ``open hinge cabinet'', and ``turn on top burner'' into the long-horizon task sequence. Importantly, the skills required to accomplish these subtasks are not present in the original data library. An example of the task rendering is shown in Fig. \ref{fig:zeroshot_task_snapshots}.
Our agent successfully adapts the knowledge and skills learned from other tasks to handle these unseen scenarios, ultimately completing the new subtasks. This highlights the generalization capability of our framework, showcasing its ability to apply prior knowledge to previously unencountered tasks.

\subsection{Scalability}
\begin{figure}[b!]
\centering
\resizebox{.49\textwidth}{!}{
\begin{tikzpicture}[font=\sffamily]
% ---------- LEFT ----------
\begin{scope}
    \begin{axis}[
        ybar,
        bar width=10pt,
        width=5cm,
        height=5cm,
        ymin=0,
        ymax=1,
        ylabel={Average reward},
        xtick=\empty,
        xticklabels={},
        enlarge x limits=0.15,
        title={Metaworld-MT10},
    ]
    % helios
    \addplot+[
        fill=ffgreen_pv!60, draw=none,
        error bars/.cd, y explicit, error bar style={color=black}, y dir=both
    ] coordinates {(1, 0.81) +- (0,0.052)};
    % spirl
    \addplot+[
        fill=fforange_pv!60, draw=none,
        error bars/.cd, y explicit, error bar style={color=black}, y dir=both
    ] coordinates {(1, 0.75) +- (0,0.023)};
    % bc
    \addplot+[
        fill=ffblue!60, draw=none,
        error bars/.cd, y explicit, error bar style={color=black}, y dir=both
    ] coordinates {(1, 0.12) +- (0,0.06)};
    % ssp
    \addplot+[
        fill=ffdarkgreen!60, draw=none,
        error bars/.cd, y explicit, error bar style={color=black}, y dir=both
    ] coordinates {(1, 0.17) +- (0,0.03)};
    % flat prior
    \addplot+[
        fill=ffred!60, draw=none,
        error bars/.cd, y explicit, error bar style={color=black}, y dir=both
    ] coordinates {(1, 0.03) +- (0,0.05)};
    % sac
    \addplot+[
        fill=ffyellow!80, draw=none,
        error bars/.cd, y explicit, error bar style={color=black}, y dir=both
    ] coordinates {(1, 0.03) +- (0,0.04)};
    \end{axis}
\end{scope}

% ---------- RIGHT ----------
\begin{scope}[xshift=5cm] % adjust spacing here
    \begin{axis}[
        ybar,
        bar width=10pt,
        width=5cm,
        height=5cm,
        ymin=0,
        ymax=2,
        ylabel={Average reward},
        xtick=\empty,
        enlarge x limits=0.15,
        title={LIBERO-long(10)},
    ]
    % helios
    \addplot+[
        fill=ffgreen_pv!60, draw=none,
        error bars/.cd, y explicit, error bar style={color=black}, y dir=both
    ] coordinates {(1, 1.52) +- (0,0.10)};
    % spirl
    \addplot+[
        fill=fforange_pv!60, draw=none,
        error bars/.cd, y explicit, error bar style={color=black}, y dir=both
    ] coordinates {(1, 1.2) +- (0,0.12)};
    % bc
    \addplot+[
        fill=ffblue!60, draw=none,
        error bars/.cd, y explicit, error bar style={color=black}, y dir=both
    ] coordinates {(1, 0.62) +- (0,0.09)};
    
    \addplot+[
        fill=ffdarkgreen!60, draw=none,
        error bars/.cd, y explicit, error bar style={color=black}, y dir=both
    ] coordinates {(1, 0.6) +- (0,0.05)};
    
    \addplot+[
        fill=ffred!60, draw=none,
        error bars/.cd, y explicit, error bar style={color=black}, y dir=both
    ] coordinates {(1, 0.02) +- (0,0.06)};
    
    \addplot+[
        fill=ffyellow!80, draw=none,
        error bars/.cd, y explicit, error bar style={color=black}, y dir=both
    ] coordinates {(1, 0.05) +- (0,0.11)};
    \end{axis}
\end{scope}

\node[fill=ffgreen_pv!60, label={[label distance=0pt]0:\footnotesize \textbf{\textsf{HELIOS (Ours)}}}](1)at(0cm, -.5cm){};
\node[fill=fforange_pv!60, label={[label distance=0pt]0:\footnotesize \textsf{SPIRL}}, right](2)at([xshift=3cm]1.east){};
\node[fill=ffblue!60, label={[label distance=0pt]0:\footnotesize \textsf{BC}}, right](3)at([xshift=3cm]2.east){};
\node[fill=ffdarkgreen!60, label={[label distance=0pt]0:\footnotesize \textsf{SSP}}, below](4)at([yshift=-.25cm]1.south){};
\node[fill=ffred!60, label={[label distance=0pt]0:\footnotesize \textsf{Flat Prior}}, right](5)at([xshift=3cm]4.east){};
\node[fill=ffyellow!60, label={[label distance=0pt]0:\footnotesize \textsf{SAC}}, right](6)at([xshift=3cm]5.east){};
\node[](a)at(-1cm, 3.75cm){\textbf{a}};
\node[](a)at(4cm, 3.75cm){\textbf{b}};
\end{tikzpicture}}
\vskip -.05in
\caption{\textbf{Scalability performance.} \textbf{a}, Metaworld-MT10 and \textbf{b}, LIBERO-long (10). For both benchmarks, we use subtask success signals as sparse rewards. All models are trained for 1.5M RL steps, except Flat-prior and SAC, which are trained for 5M steps to ensure convergence.}
\label{fig:scalability}
\end{figure}
To evaluate scalability across diverse long-horizon manipulation tasks, Fig.~\ref{fig:scalability} reports performance on Meta-World MT10 and LIBERO-Long under sparse-reward training. HELIOS achieves the highest average reward on both benchmarks, reaching 0.81 on MT10 and 1.52 on LIBERO-Long, outperforming recent SOTA framework SPIRL (0.75 and 1.20). 
This corresponds to relative performance improvements of 8.0\% on MT10 and 26.7\% on LIBERO-Long. 
HELIOS also substantially surpasses BC, SSP, Flat Prior, and SAC. 
Notably, conventional RL methods such as SAC and Flat Prior perform poorly in sparse-reward and even long-horizon settings. 
Due to inefficient exploration and the absence of structured skill abstraction, they fail to discover meaningful action patterns and cannot converge to effective behaviors even after 5M training steps, highlighting the difficulty of sparse credit assignment in long-horizon tasks. 
In contrast, our Bayesian non-parametric DPM prior adaptively adjusts skill complexity as task diversity increases, enabling more structured exploration and better modeling of heterogeneous skill modes across tasks and embodiments. 
Together with the 28.8\% gain observed on Franka Kitchen, HELIOS achieves an average relative improvement of \textbf{21.8\%} across the three benchmarks, demonstrating strong scalability and generalization in sparse-reward long-horizon manipulation.

\subsection{Real-world performance}
Fig.~\ref{fig:realworld} presents real-world validation on a KUKA iiwa platform. 
The task setups follow LIBERO-Long. 
The action space is defined in Cartesian end-effector space with an additional gripper command, which facilitates direct transfer to the heterogeneous KUKA iiwa platform without finetuning policies.
In task (a), the robot sequentially detects, grasps, and transports the white soup can and yellow protein bar before placing them into the basket. 
In task (b), it first turns on the stove by rotating the button, then grasps the moka pot and positions it onto the burner. 
In task (c), the robot picks and places the white tea cup on the plate and the pink candy bar on the side to arrange a tea set. 
In task (d), it grasps the red mug, places the mug into the microwave, and closes the door to heat the beverage. 
In task (e), the robot is required to grasp the coffee mug, place it into the drawer, and perform a closing action.
During the deployment, HELIOS demonstrates stable skill transitions and robust execution across all long-horizon scenarios. 
Quantitatively, the system achieves an overall average success rate of 0.8 across five tasks, with three tasks reaching 7/10 or above and tea set arrangement achieving perfect success (10/10), indicating reliable multi-stage skill composition under real-world uncertainties.
The results validate effective sim-to-real transfer and show that our non-parametric skill prior captures reusable long-horizon structures beyond specific embodiments and environments.
For more details, refer to our supplementary video.

\newcolumntype{C}[1]{>{\centering\arraybackslash}m{#1}}
\begin{figure*}[ht!]
    \centering
    \resizebox{.98\textwidth}{!}{
    \begin{tikzpicture}[font=\sffamily]
        % task1
        \node[inner sep=0pt, draw=none, fill=none](1)at(0, 0cm){\includegraphics[width=3cm]{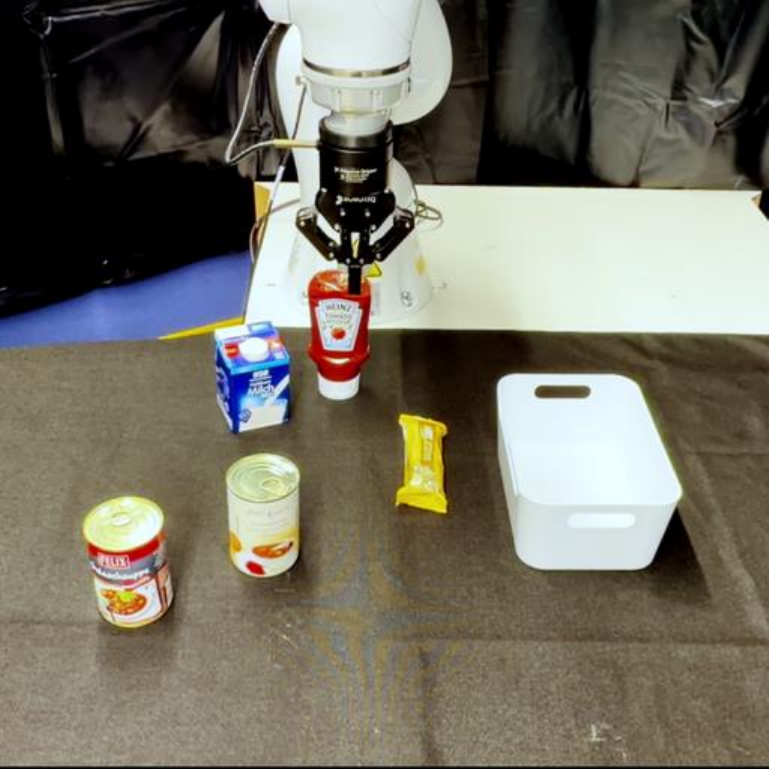}};
        \node[inner sep=0pt, draw=none, fill=none, right](2)at([xshift=0.1cm]1.east){\includegraphics[width=3cm]{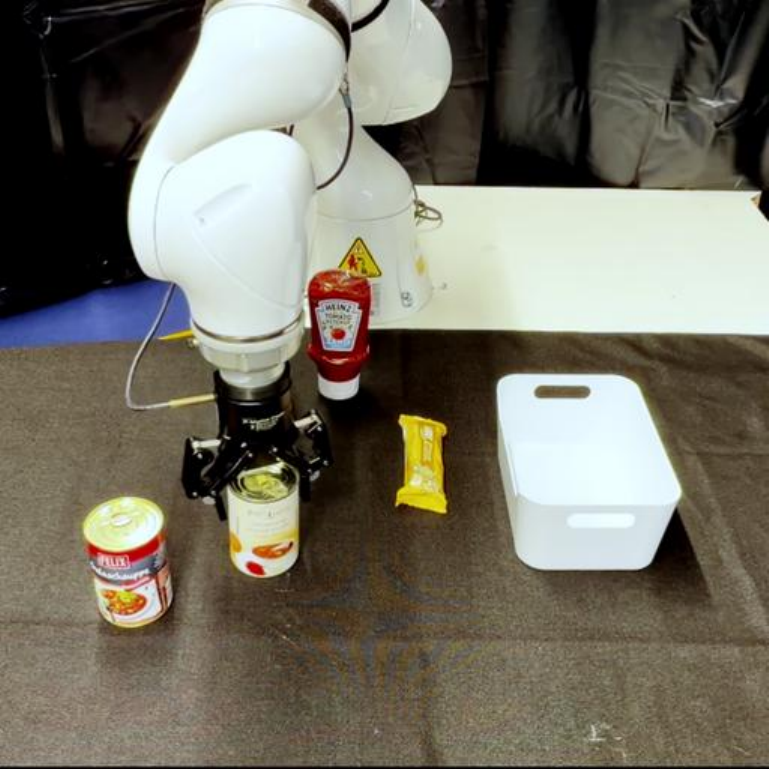}};
        \node[inner sep=0pt, draw=none, fill=none, right](3)at([xshift=0.1cm]2.east){\includegraphics[width=3cm]{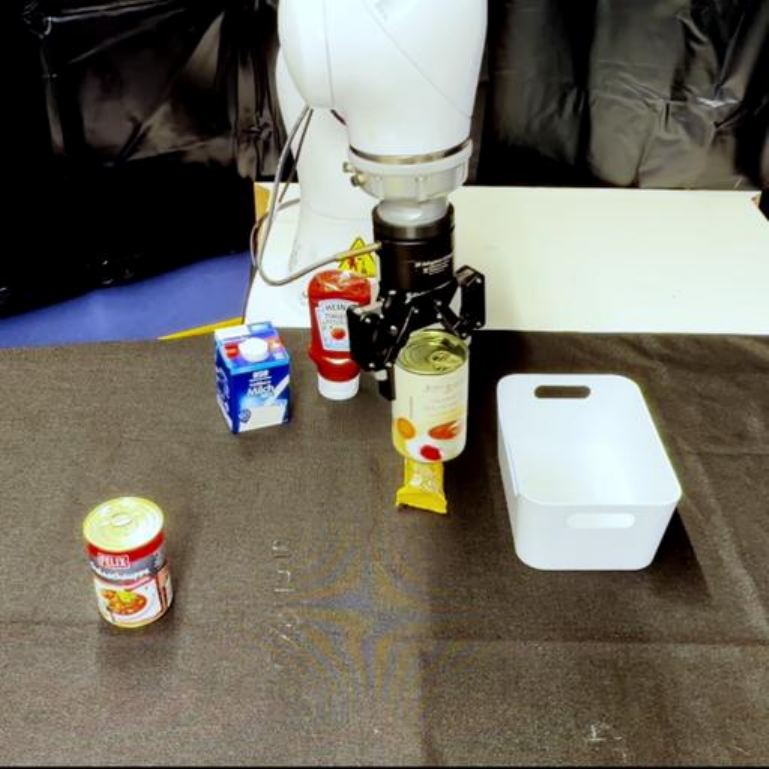}};
        \node[inner sep=0pt, draw=none, fill=none, right](4)at([xshift=0.1cm]3.east){\includegraphics[width=3cm]{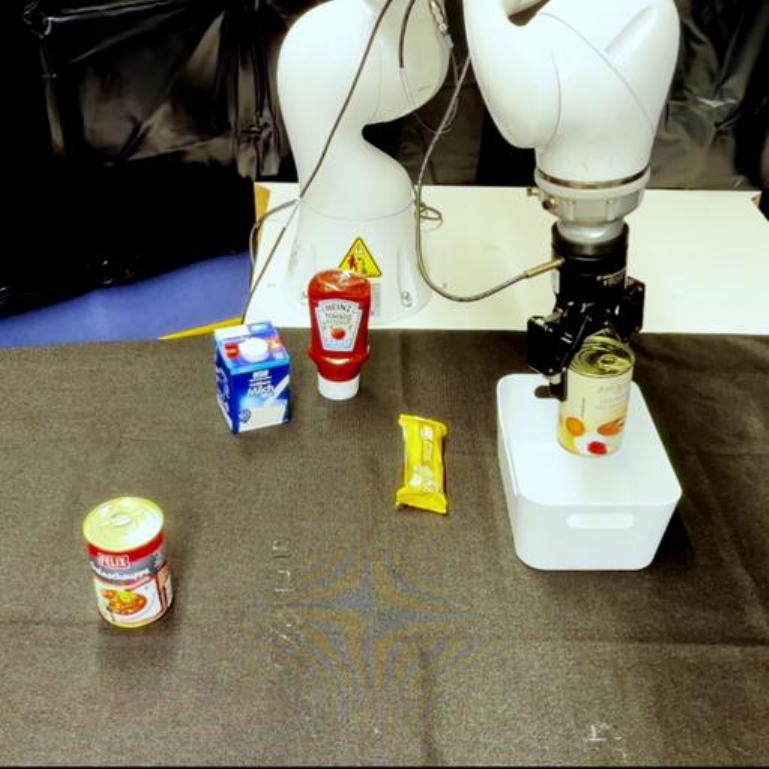}};
        \node[inner sep=0pt, draw=none, fill=none, right](5)at([xshift=0.1cm]4.east){\includegraphics[width=3cm]{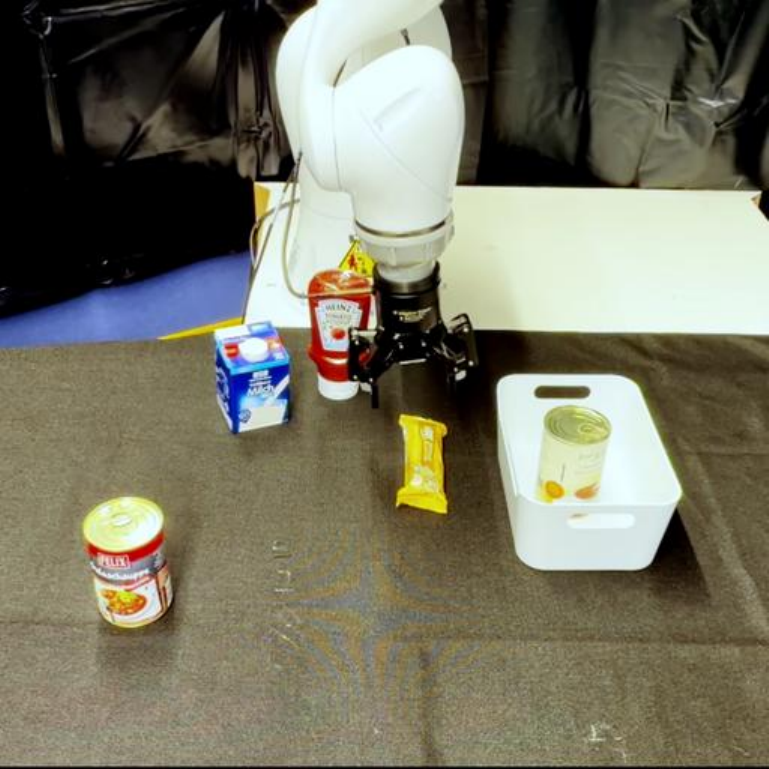}};
        \node[inner sep=0pt, draw=none, fill=none, right](6)at([xshift=0.1cm]5.east){\includegraphics[width=3cm]{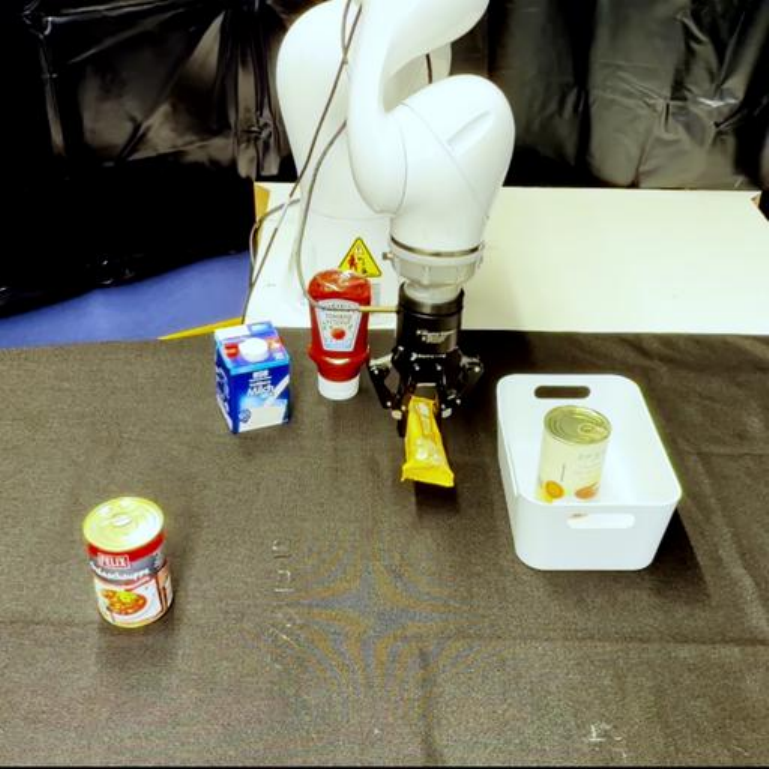}};
        \node[inner sep=0pt, draw=none, fill=none, right](7)at([xshift=0.1cm]6.east){\includegraphics[width=3cm]{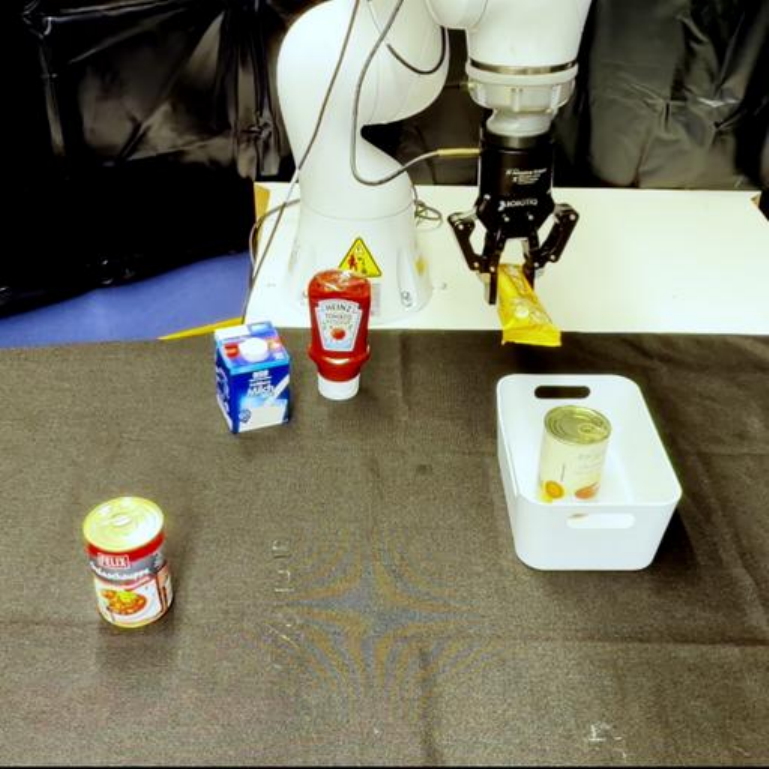}};
        \node[inner sep=0pt, draw=none, fill=none, right](8)at([xshift=0.1cm]7.east){\includegraphics[width=3cm]{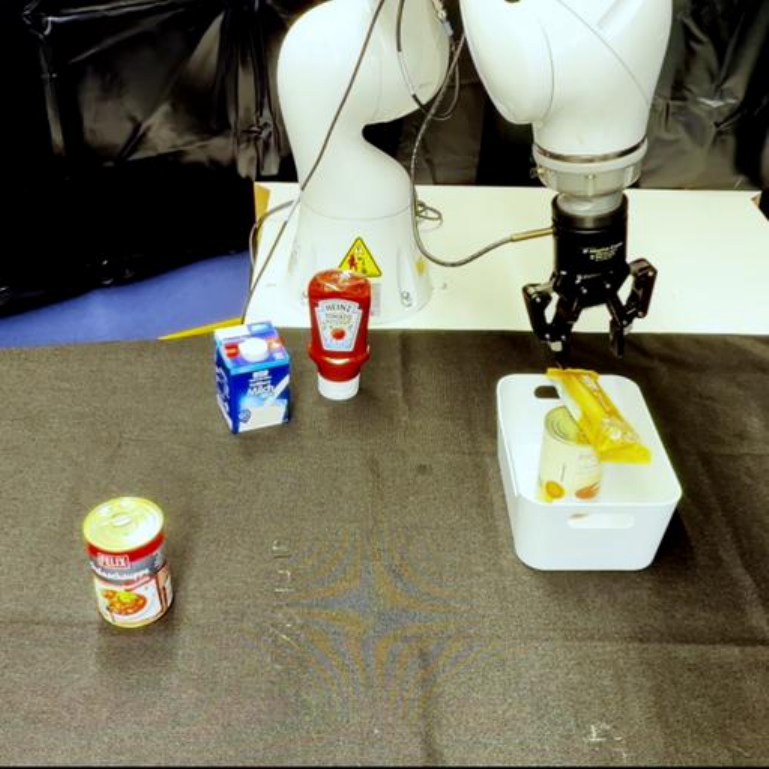}};
        % task2
        \node[inner sep=0pt, draw=none, fill=none, below](9)at([yshift=-0.1cm]1.south){\includegraphics[width=3cm]{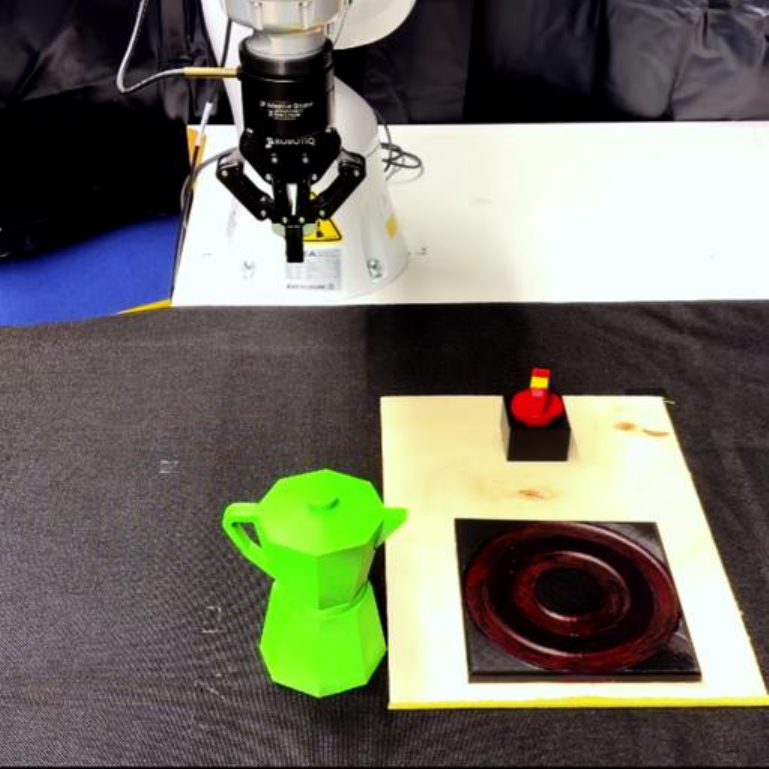}};
        \node[inner sep=0pt, draw=none, fill=none, right](10)at([xshift=0.1cm]9.east){\includegraphics[width=3cm]{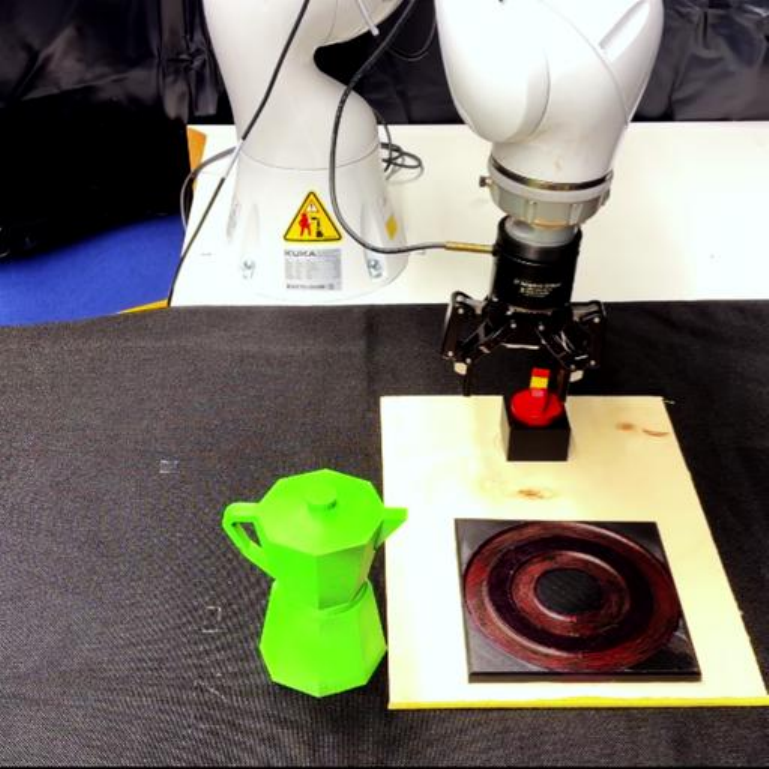}};
        \node[inner sep=0pt, draw=none, fill=none, right](11)at([xshift=0.1cm]10.east){\includegraphics[width=3cm]{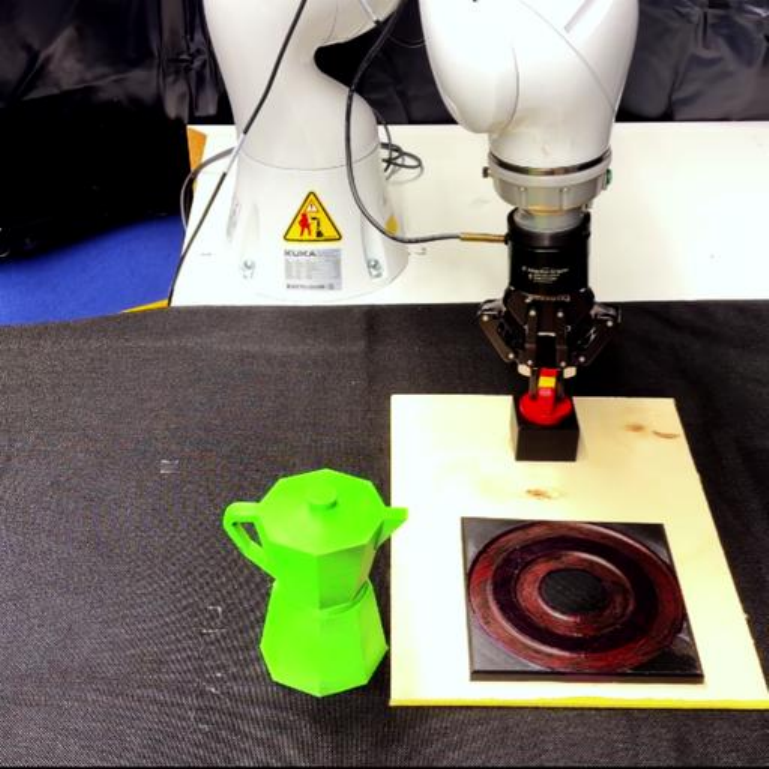}};
        \node[inner sep=0pt, draw=none, fill=none, right](12)at([xshift=0.1cm]11.east){\includegraphics[width=3cm]{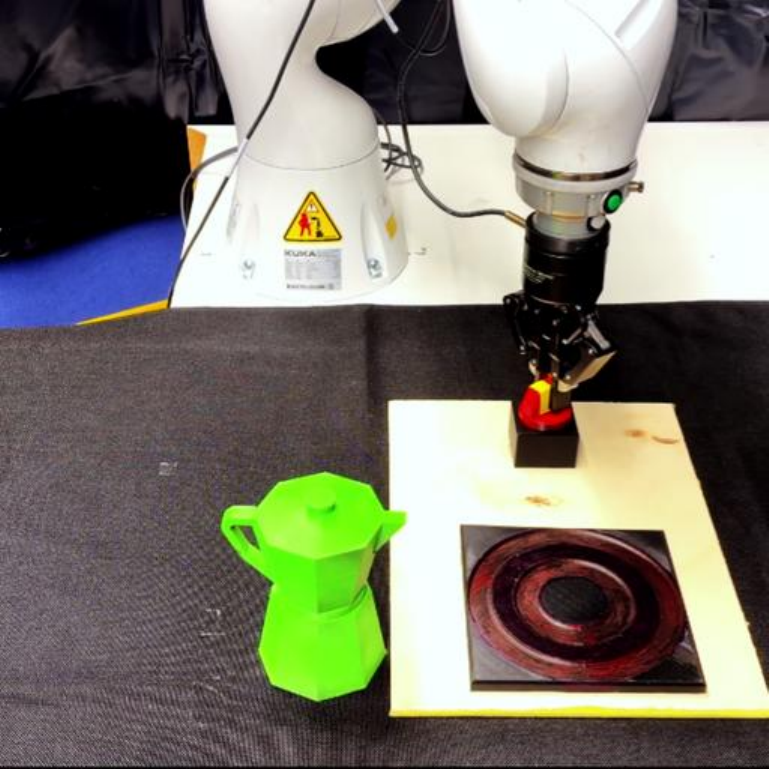}};
        \node[inner sep=0pt, draw=none, fill=none, right](13)at([xshift=0.1cm]12.east){\includegraphics[width=3cm]{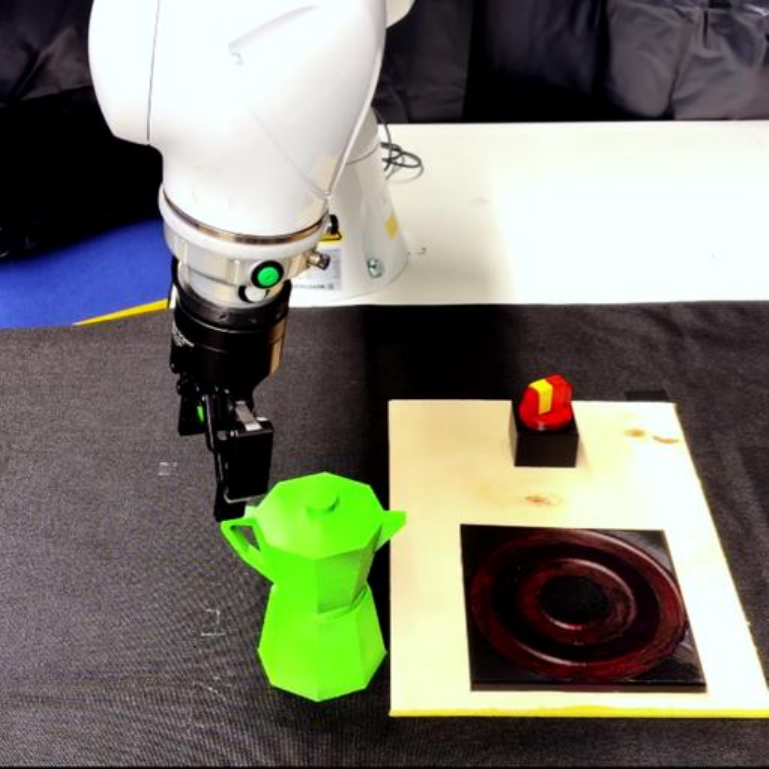}};
        \node[inner sep=0pt, draw=none, fill=none, right](14)at([xshift=0.1cm]13.east){\includegraphics[width=3cm]{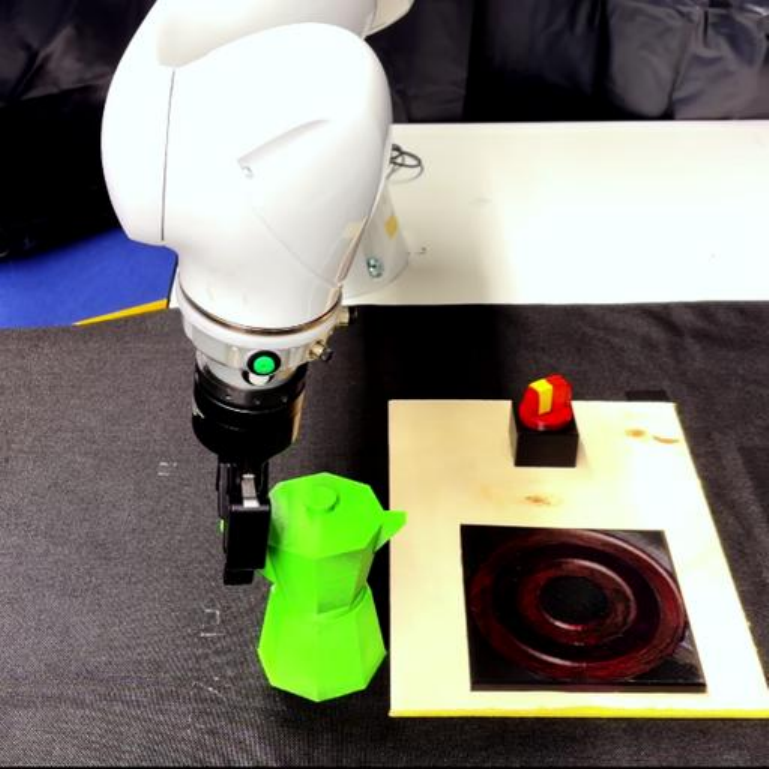}};
        \node[inner sep=0pt, draw=none, fill=none, right](15)at([xshift=0.1cm]14.east){\includegraphics[width=3cm]{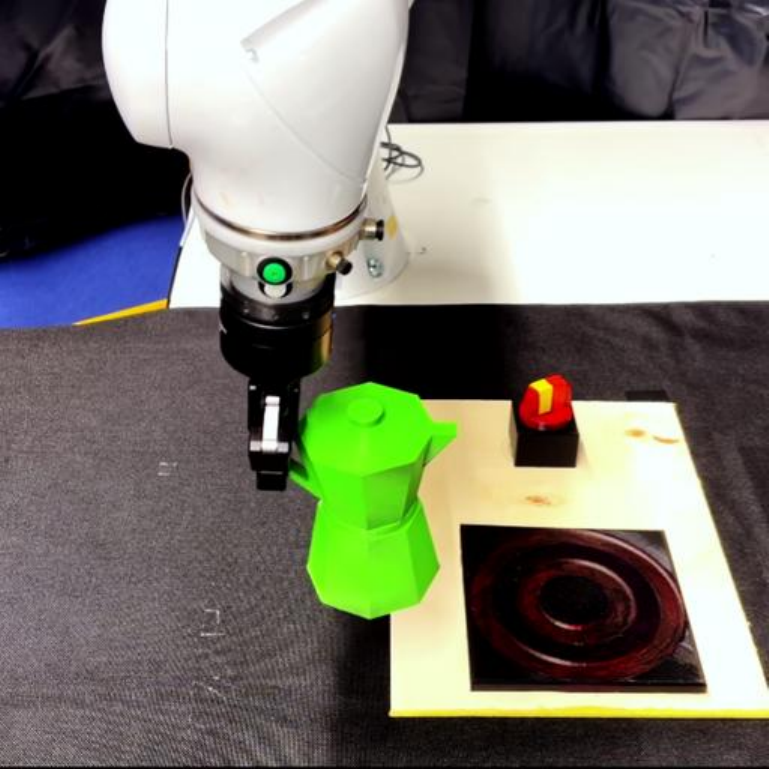}};
        \node[inner sep=0pt, draw=none, fill=none, right](16)at([xshift=0.1cm]15.east){\includegraphics[width=3cm]{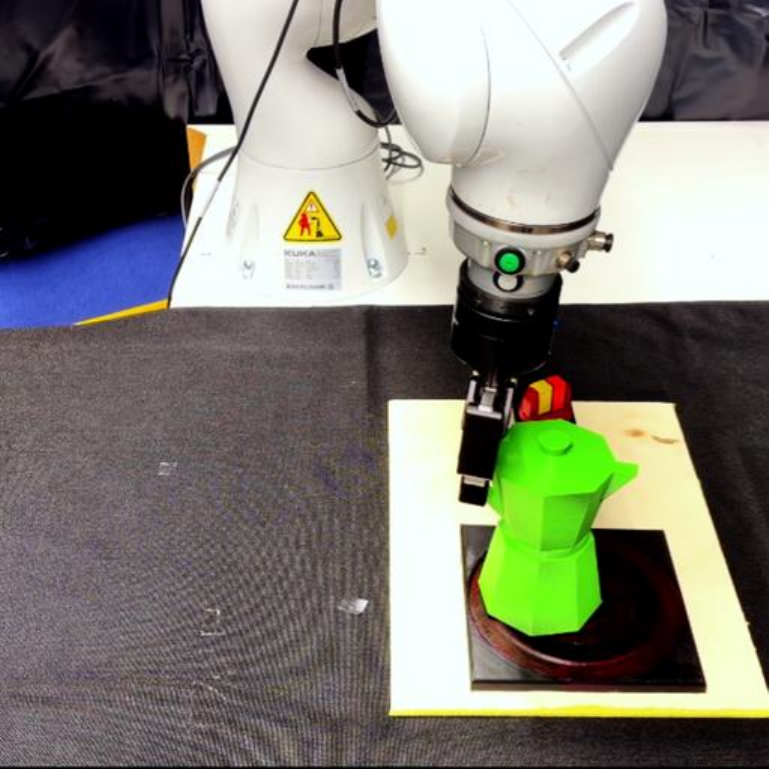}};
        % task3
        \node[inner sep=0pt, draw=none, fill=none, below](17)at([yshift=-0.1cm]9.south){\includegraphics[width=3cm]{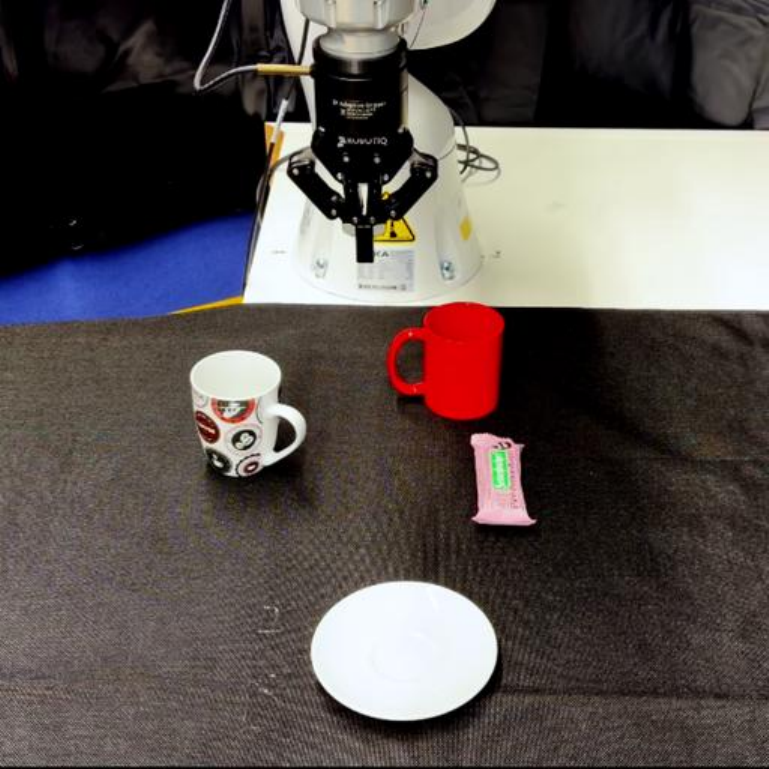}};
        \node[inner sep=0pt, draw=none, fill=none, right](18)at([xshift=0.1cm]17.east){\includegraphics[width=3cm]{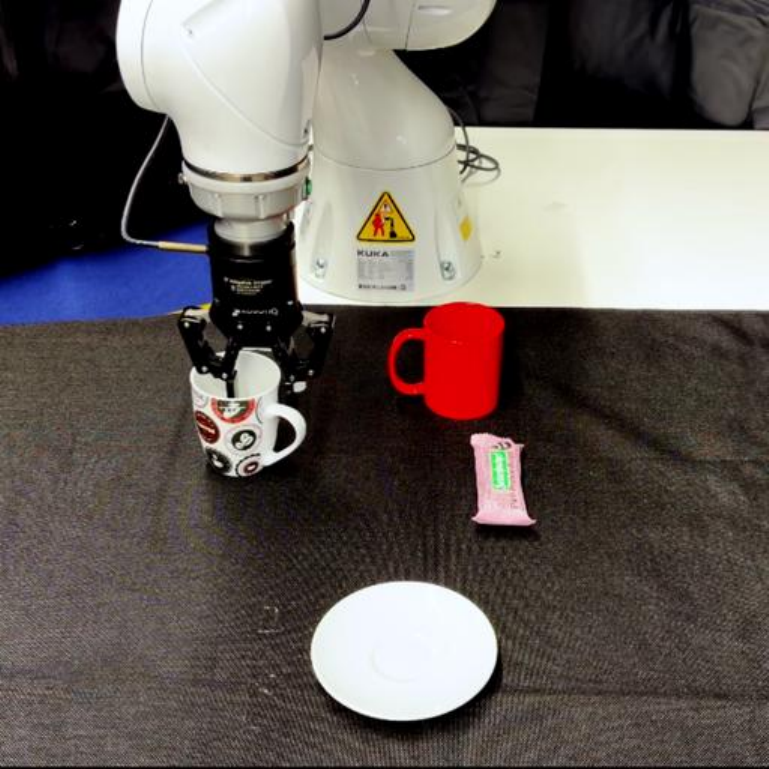}};
        \node[inner sep=0pt, draw=none, fill=none, right](19)at([xshift=0.1cm]18.east){\includegraphics[width=3cm]{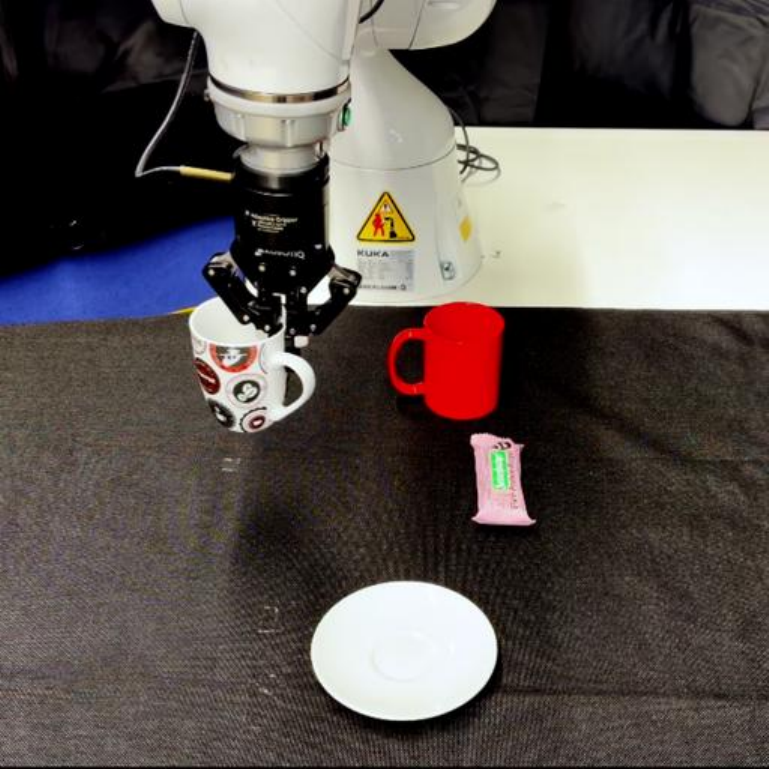}};
        \node[inner sep=0pt, draw=none, fill=none, right](20)at([xshift=0.1cm]19.east){\includegraphics[width=3cm]{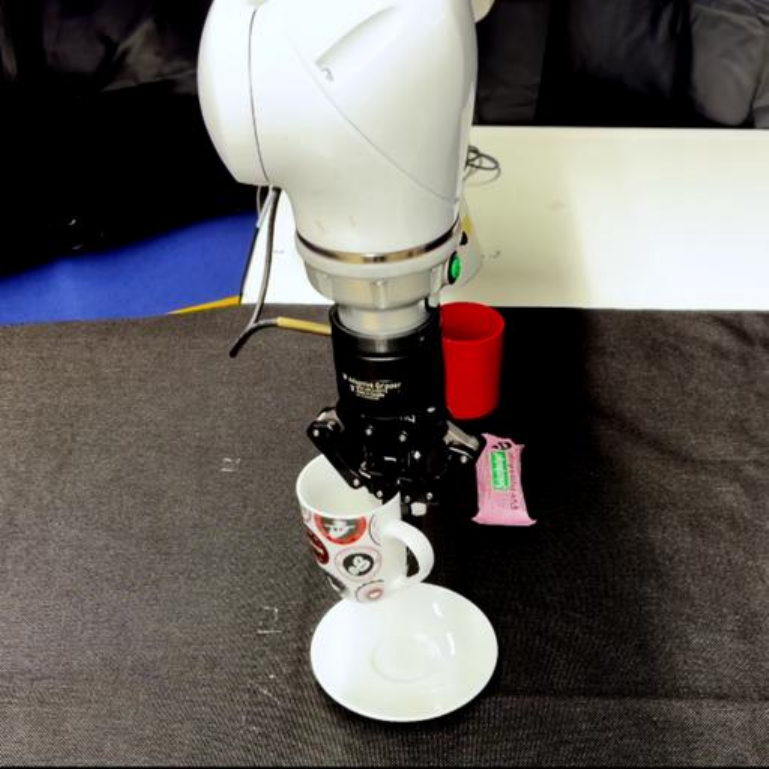}};
        \node[inner sep=0pt, draw=none, fill=none, right](21)at([xshift=0.1cm]20.east){\includegraphics[width=3cm]{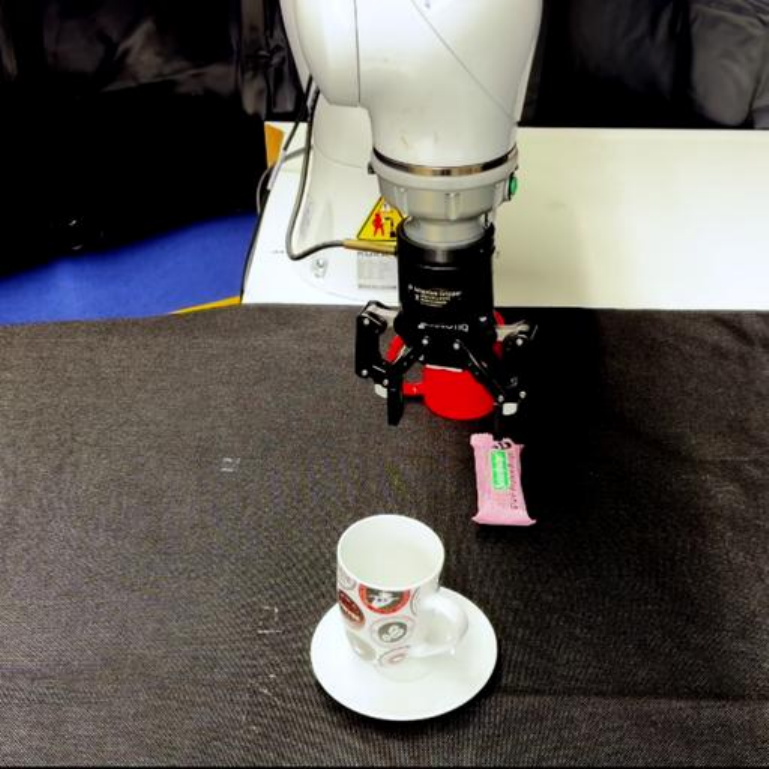}};
        \node[inner sep=0pt, draw=none, fill=none, right](22)at([xshift=0.1cm]21.east){\includegraphics[width=3cm]{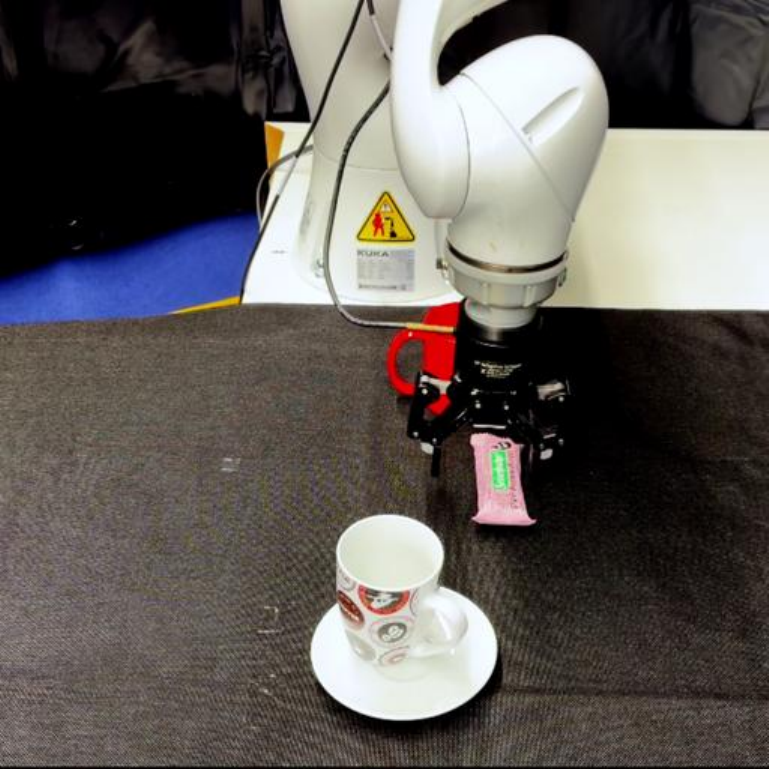}};
        \node[inner sep=0pt, draw=none, fill=none, right](23)at([xshift=0.1cm]22.east){\includegraphics[width=3cm]{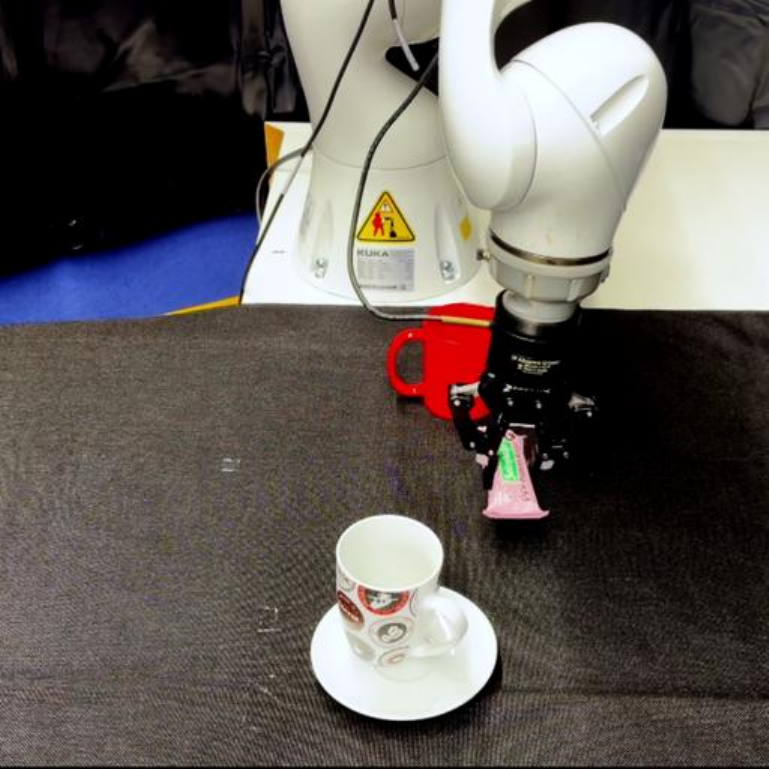}};
        \node[inner sep=0pt, draw=none, fill=none, right](24)at([xshift=0.1cm]23.east){\includegraphics[width=3cm]{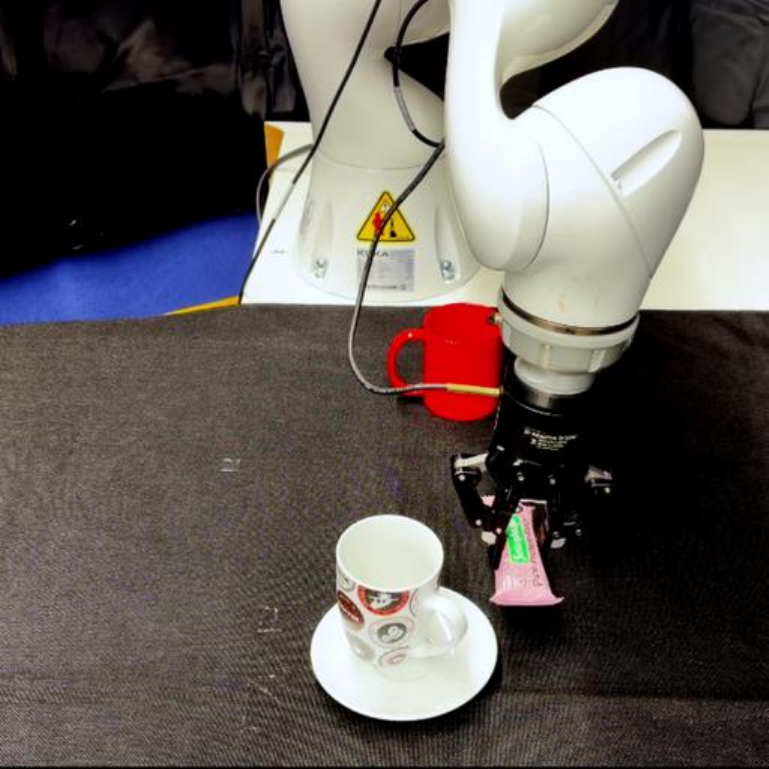}};
        % task4
        \node[inner sep=0pt, draw=none, fill=none, below](25)at([yshift=-0.1cm]17.south){\includegraphics[width=3cm]{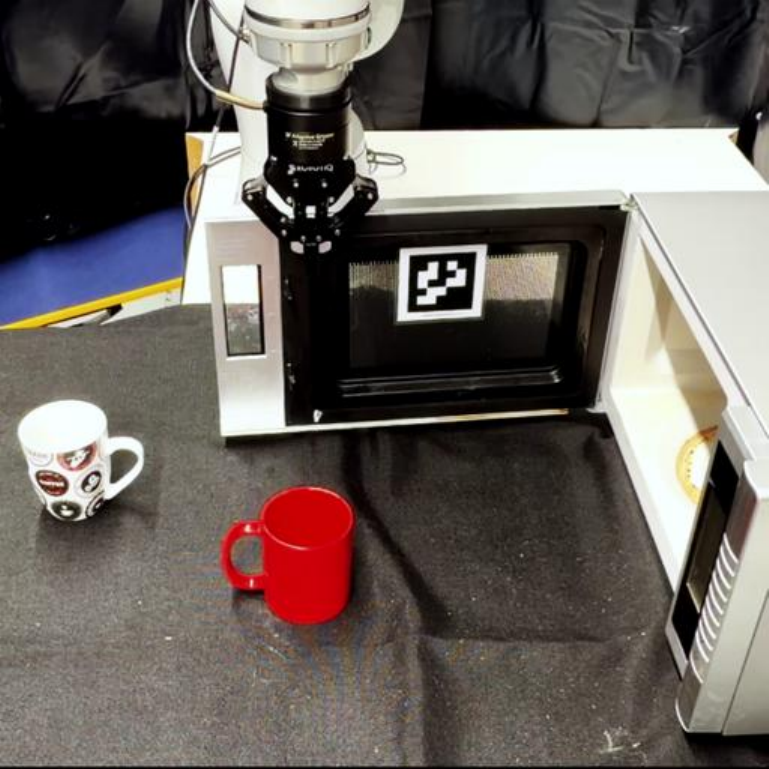}};
        \node[inner sep=0pt, draw=none, fill=none, right](26)at([xshift=0.1cm]25.east){\includegraphics[width=3cm]{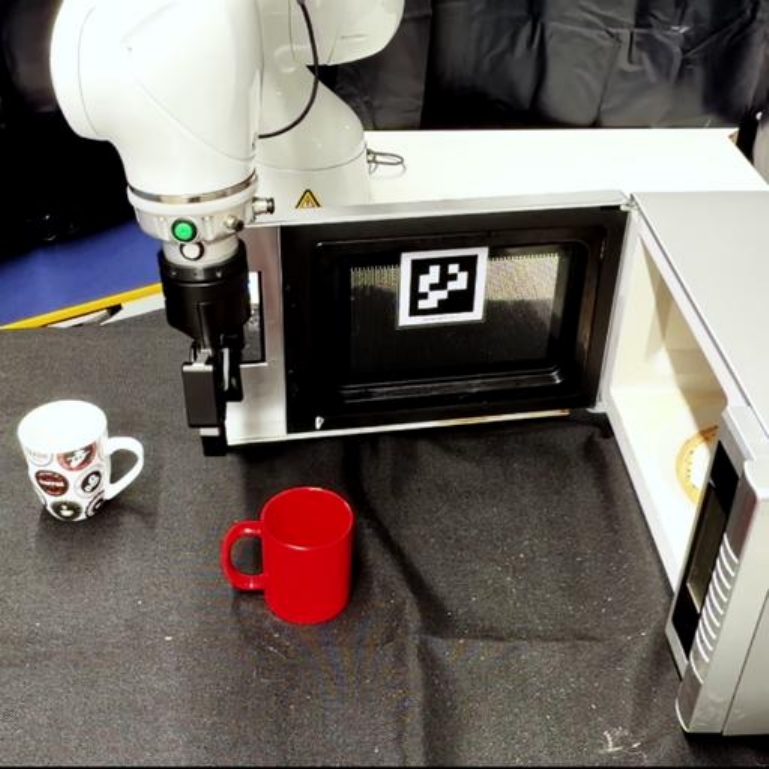}};
        \node[inner sep=0pt, draw=none, fill=none, right](27)at([xshift=0.1cm]26.east){\includegraphics[width=3cm]{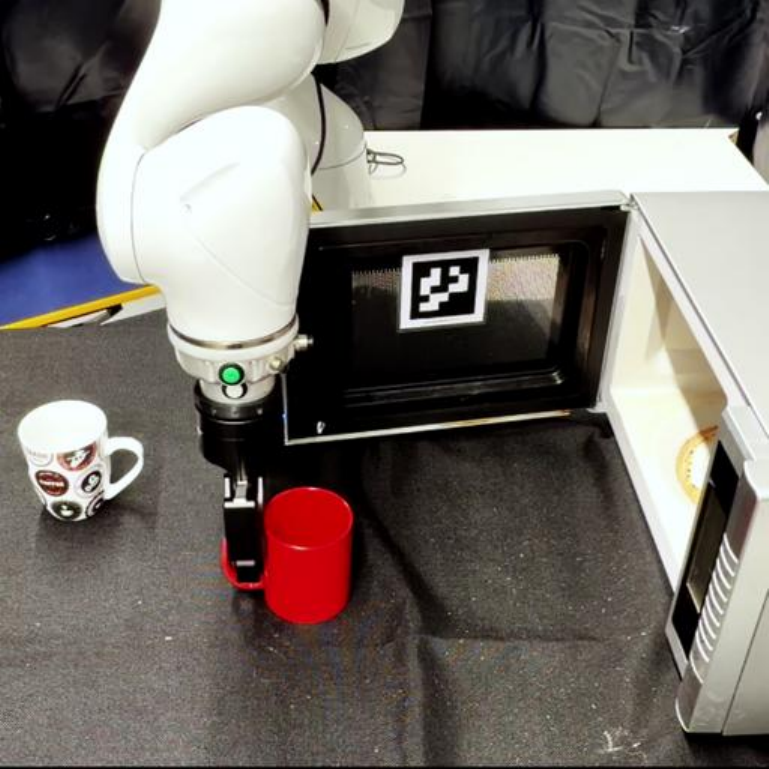}};
        \node[inner sep=0pt, draw=none, fill=none, right](28)at([xshift=0.1cm]27.east){\includegraphics[width=3cm]{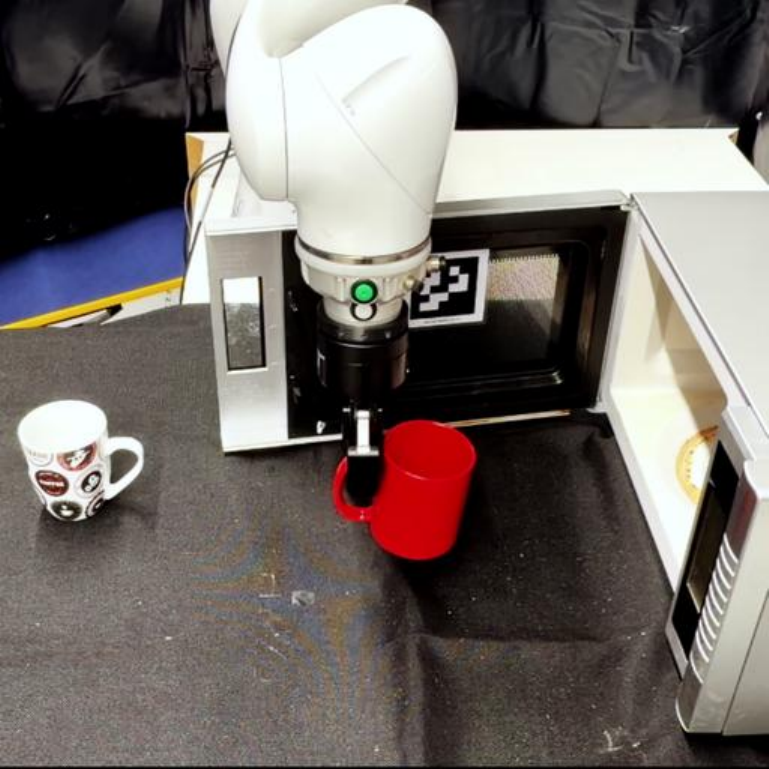}};
        \node[inner sep=0pt, draw=none, fill=none, right](29)at([xshift=0.1cm]28.east){\includegraphics[width=3cm]{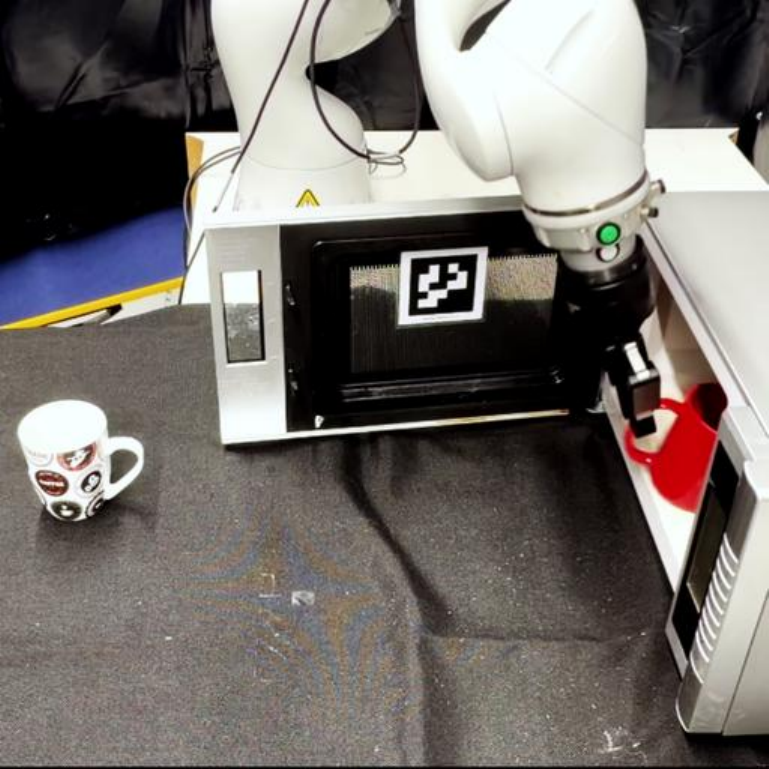}};
        \node[inner sep=0pt, draw=none, fill=none, right](30)at([xshift=0.1cm]29.east){\includegraphics[width=3cm]{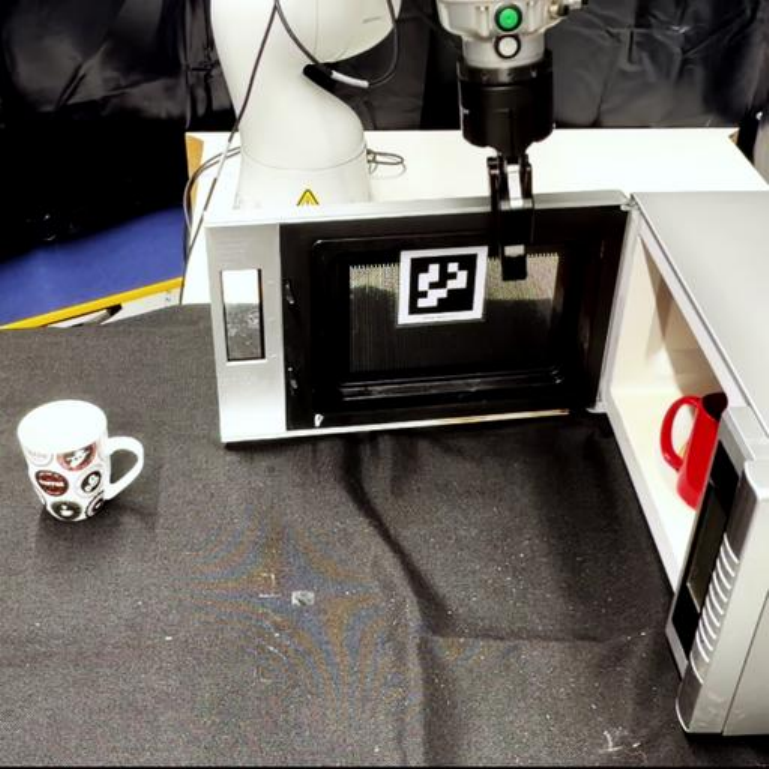}};
        \node[inner sep=0pt, draw=none, fill=none, right](31)at([xshift=0.1cm]30.east){\includegraphics[width=3cm]{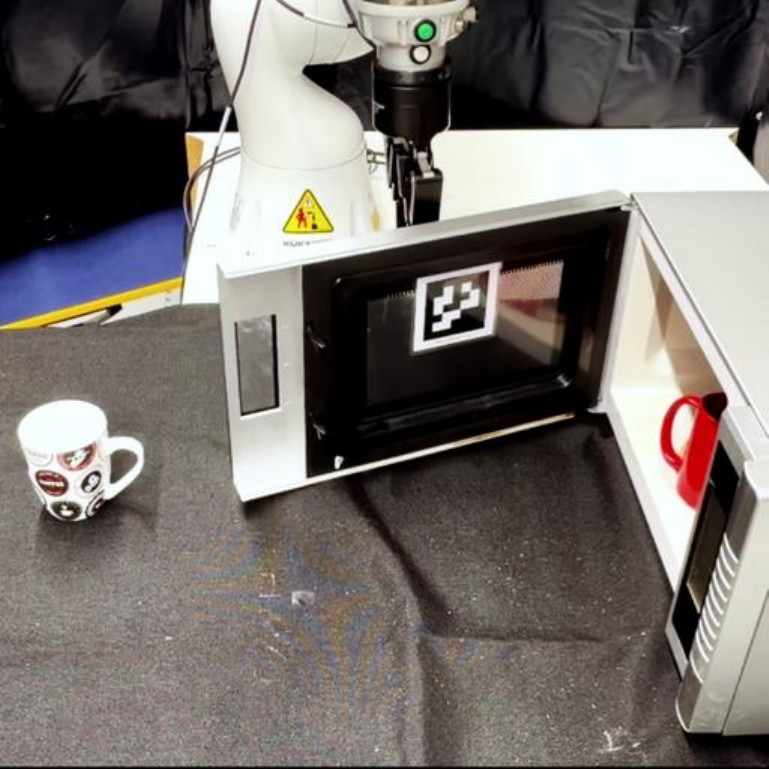}};
        \node[inner sep=0pt, draw=none, fill=none, right](32)at([xshift=0.1cm]31.east){\includegraphics[width=3cm]{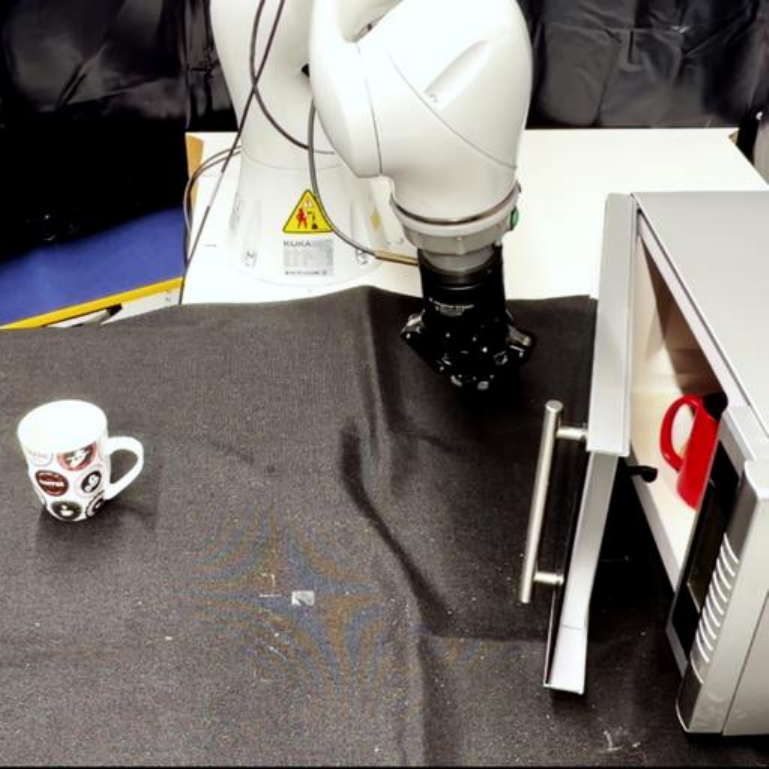}};
        % task5
        \node[inner sep=0pt, draw=none, fill=none, below](33)at([yshift=-0.1cm]25.south){\includegraphics[width=3cm]{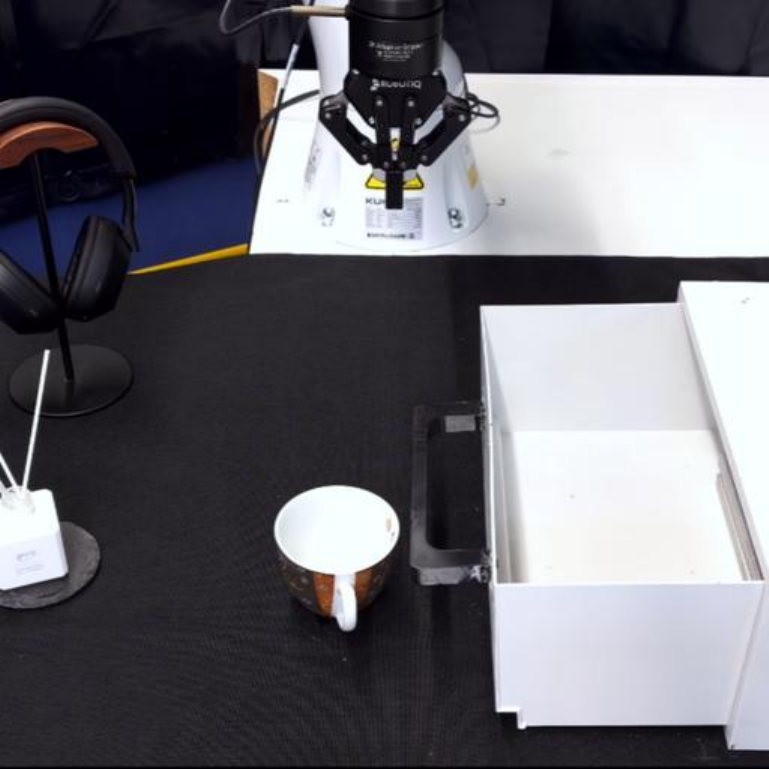}};
        \node[inner sep=0pt, draw=none, fill=none, right](34)at([xshift=0.1cm]33.east){\includegraphics[width=3cm]{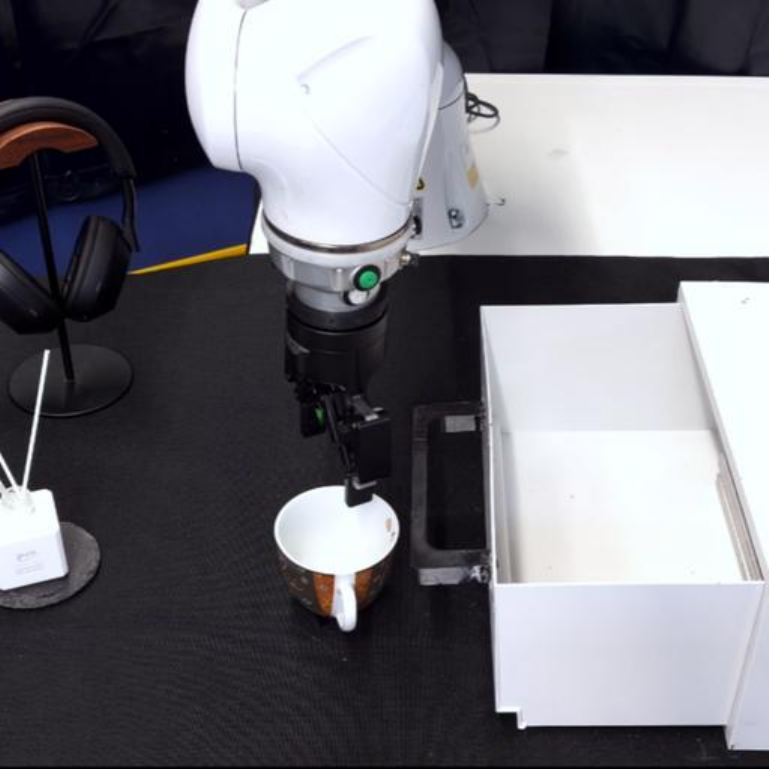}};
        \node[inner sep=0pt, draw=none, fill=none, right](35)at([xshift=0.1cm]34.east){\includegraphics[width=3cm]{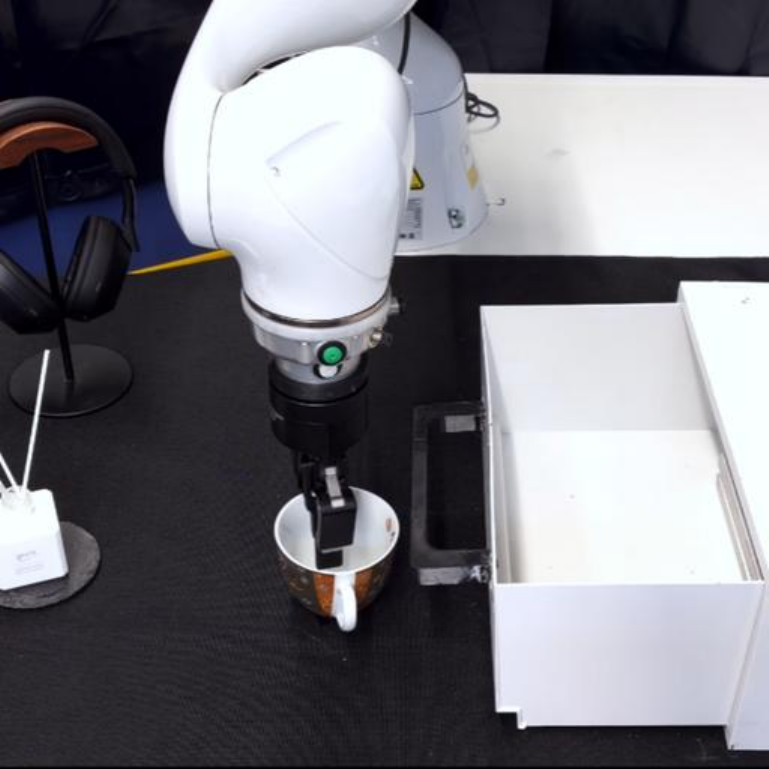}};
        \node[inner sep=0pt, draw=none, fill=none, right](36)at([xshift=0.1cm]35.east){\includegraphics[width=3cm]{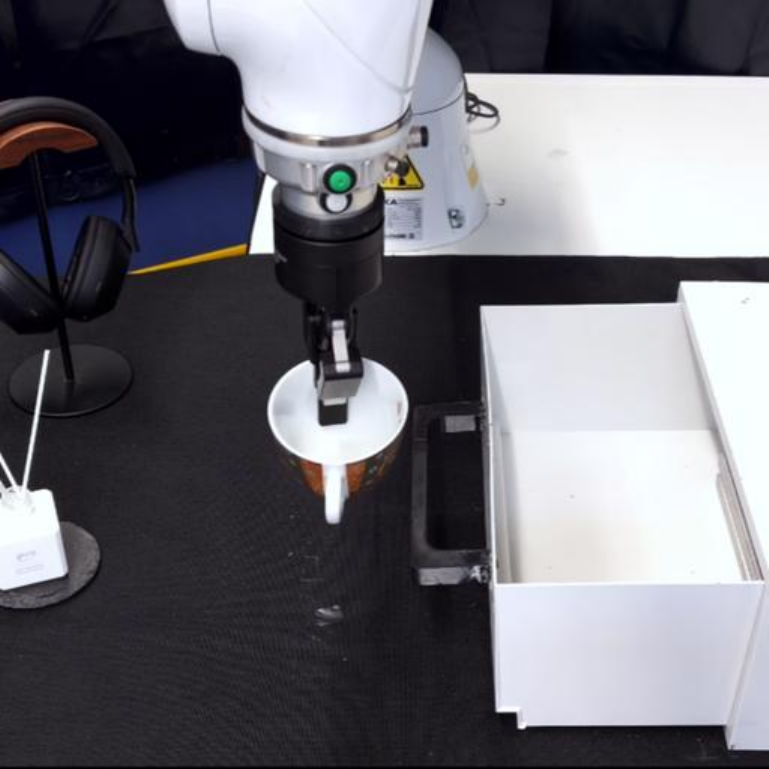}};
        \node[inner sep=0pt, draw=none, fill=none, right](37)at([xshift=0.1cm]36.east){\includegraphics[width=3cm]{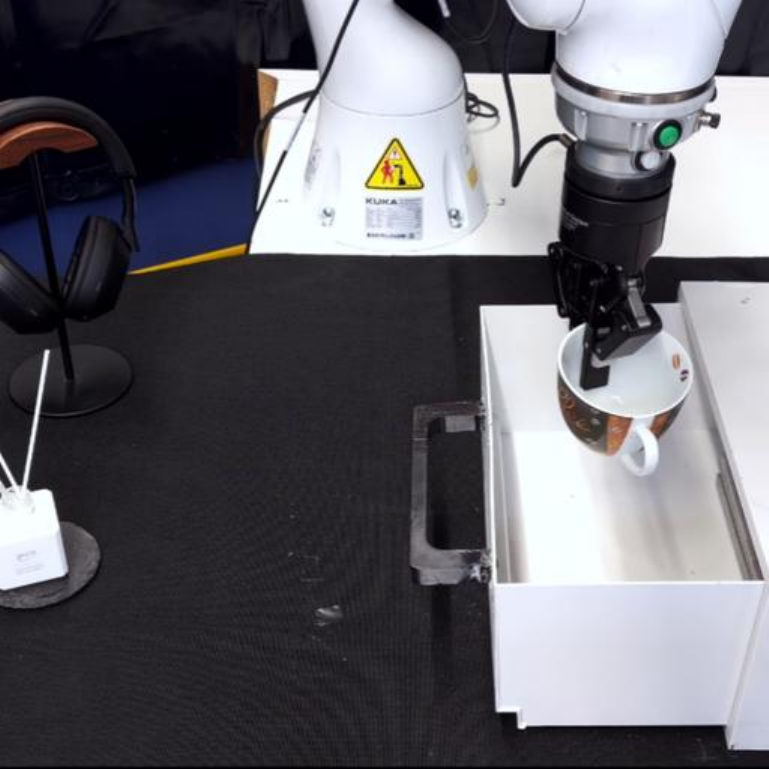}};
        \node[inner sep=0pt, draw=none, fill=none, right](38)at([xshift=0.1cm]37.east){\includegraphics[width=3cm]{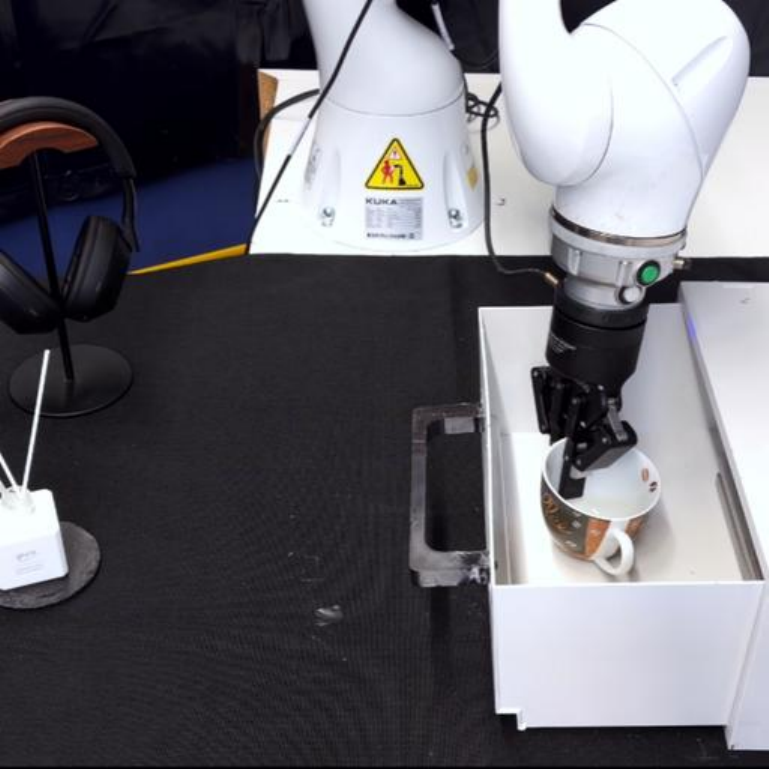}};
        \node[inner sep=0pt, draw=none, fill=none, right](39)at([xshift=0.1cm]38.east){\includegraphics[width=3cm]{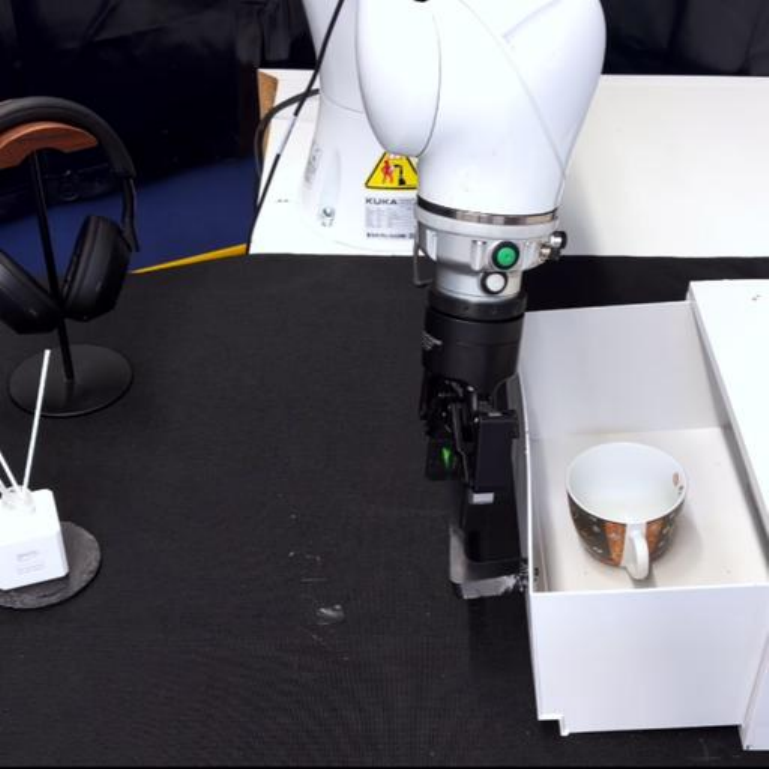}};
        \node[inner sep=0pt, draw=none, fill=none, right](40)at([xshift=0.1cm]39.east){\includegraphics[width=3cm]{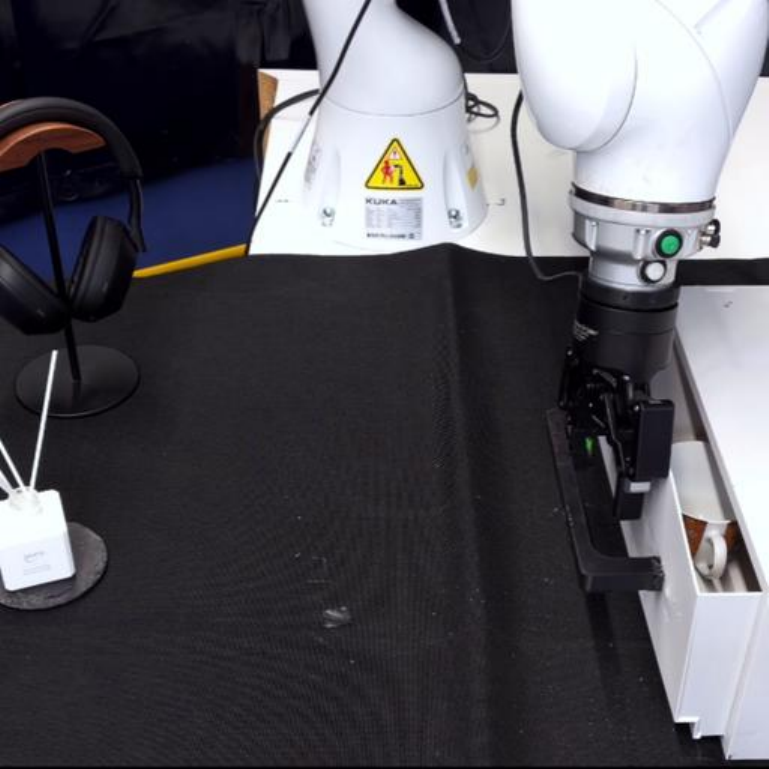}};
        % label
        \node[below right](a)at(1.north west){\large\textcolor{white}{\textbf{a}}};
        \node[below right](b)at(9.north west){\large\textcolor{white}{\textbf{b}}};
        \node[below right](c)at(17.north west){\large\textcolor{white}{\textbf{c}}};
        \node[below right](d)at(25.north west){\large\textcolor{white}{\textbf{d}}};
        \node[below right](e)at(33.north west){\large\textcolor{white}{\textbf{e}}};
        % table
        \node[below right, inner sep=0pt] at([yshift=-0.1cm]33.south west) (table) {
        % \rowcolors{0}{white}{gray!30}
            \begin{tabular}{c C{4.19cm} C{4.19cm} C{4.19cm} C{4.19cm} C{4.19cm}}
                \toprule
                \rowcolor{nature_tab_gray1}
                \textbf{average} & \textbf{a. basket placement} & \textbf{b. moka coffee brewing} & \textbf{c. tea set arrangement} & \textbf{d. microwave heating} & \textbf{e. drawer storage} \\
                \midrule
                0.8 & 8/10 & 6/10 & 10/10 & 7/10 & 9/10 \\
                \bottomrule
            \end{tabular}
        };
    \end{tikzpicture}}
    \vskip -.1in
    \caption{\textbf{Real-world demonstrations of long-horizon manipulation tasks.}
            \textbf{a}, Pick up the white soup can and protein bar and place into the basket;
            \textbf{b}, Turn on the oven and brew moka coffee;
            \textbf{c}, Arrange the tea set for a lunch break;
            \textbf{d}, Heat the beverage in the red mug using a microwave;
            \textbf{e}, Place the mug into the drawer and close it.
            Table reports the success rate over 10 trials for each scenario.
            For more details, refer to our supplementary video.
    }
    \vskip -.22in
    \label{fig:realworld}
\end{figure*}

% -- conclusions --
\section{CONCLUSIONS}
In this work, we presented HELIOS, a hierarchical RL framework with a Bayesian non-parametric skill prior for sparse-reward long-horizon manipulation. By modeling temporally extended skills with a DPM, HELIOS adaptively discovers structured latent skill components, improving exploration and credit assignment without predefining the number of skills. Experiments on Franka Kitchen, Meta-World MT10, and LIBERO-Long show consistent gains over parametric priors and RL baselines, achieving a 21.8\% average relative improvement. Real-world KUKA iiwa results further validate the transferability of the learned skill structures.

% \addtolength{\textheight}{-12cm}   % This command serves to balance the column lengths
                                  % on the last page of the document manually. It shortens
                                  % the textheight of the last page by a suitable amount.
                                  % This command does not take effect until the next page
                                  % so it should come on the page before the last. Make
                                  % sure that you do not shorten the textheight too much.

\section*{ACKNOWLEDGMENT}
This work was supported by the 2026 Innovation Project (General Program) of the National Key Laboratory for Novel Software Technology under Grant ZZKT2026A36.

% %%%%%%%%%%%%%%%%%%%%%%%%%
% % -- REFERENCES --
% \bibliographystyle{IEEEtran}
% \bibliography{references}

\begin{thebibliography}{10}
\providecommand{\url}[1]{#1}
\csname url@samestyle\endcsname
\providecommand{\newblock}{\relax}
\providecommand{\bibinfo}[2]{#2}
\providecommand{\BIBentrySTDinterwordspacing}{\spaceskip=0pt\relax}
\providecommand{\BIBentryALTinterwordstretchfactor}{4}
\providecommand{\BIBentryALTinterwordspacing}{\spaceskip=\fontdimen2\font plus
\BIBentryALTinterwordstretchfactor\fontdimen3\font minus
  \fontdimen4\font\relax}
\providecommand{\BIBforeignlanguage}[2]{{%
\expandafter\ifx\csname l@#1\endcsname\relax
\typeout{** WARNING: IEEEtran.bst: No hyphenation pattern has been}%
\typeout{** loaded for the language `#1'. Using the pattern for}%
\typeout{** the default language instead.}%
\else
\language=\csname l@#1\endcsname
\fi
#2}}
\providecommand{\BIBdecl}{\relax}
\BIBdecl

\bibitem{pertsch2021accelerating}
K.~Pertsch \emph{et~al.}, ``Accelerating reinforcement learning with learned
  skill priors,'' in \emph{Conference on robot learning}.\hskip 1em plus 0.5em
  minus 0.4em\relax PMLR, 2021, pp. 188--204.

\bibitem{zhang2024review}
L.~Zhang, K.~Cai, Z.~Sun, Z.~Bing, C.~Wang, L.~Figueredo, S.~Haddadin, and
  A.~Knoll, ``Motion planning for robotics: A review for sampling-based
  planners,'' \emph{Biomimetic Intelligence and Robotics}, vol.~5, no.~1, p.
  100207, 2025.

\bibitem{11123767}
L.~Zhang, S.~Wang, K.~Cai, Z.~Bing, F.~Wu, C.~Wang, S.~Haddadin, and A.~Knoll,
  ``Apt*: Asymptotically optimal motion planning via adaptively prolated
  elliptical r-nearest neighbors,'' \emph{IEEE Robotics and Automation
  Letters}, vol.~10, no.~10, pp. 10\,242--10\,249, 2025.

\bibitem{meng2025growing}
Y.~Meng, Z.~Sun, M.~Fest, X.~Li, Z.~Bing, and A.~Knoll, ``Growing with your
  embodied agent: A human-in-the-loop lifelong code generation framework for
  long-horizon manipulation skills,'' \emph{arXiv preprint arXiv:2509.18597},
  2025.

\bibitem{yu2020meta}
T.~Yu \emph{et~al.}, ``Meta-world: A benchmark and evaluation for multi-task
  and meta reinforcement learning,'' in \emph{Conference on robot
  learning}.\hskip 1em plus 0.5em minus 0.4em\relax PMLR, 2020, pp. 1094--1100.

\bibitem{hughes2013memoized}
M.~C. Hughes and E.~Sudderth, ``Memoized online variational inference for
  dirichlet process mixture models,'' \emph{Advances in neural information
  processing systems}, vol.~26, 2013.

\bibitem{fu2020d4rl}
J.~Fu \emph{et~al.}, ``D4rl: Datasets for deep data-driven reinforcement
  learning,'' \emph{arXiv preprint arXiv:2004.07219}, 2020.

\bibitem{liu2023libero}
B.~Liu, Y.~Zhu, C.~Gao, Y.~Feng, Q.~Liu, Y.~Zhu, and P.~Stone, ``Libero:
  Benchmarking knowledge transfer for lifelong robot learning,'' \emph{Advances
  in Neural Information Processing Systems}, vol.~36, pp. 44\,776--44\,791,
  2023.

\bibitem{rakelly2019efficient}
K.~Rakelly \emph{et~al.}, ``Efficient off-policy meta-reinforcement learning
  via probabilistic context variables,'' in \emph{International conference on
  machine learning}.\hskip 1em plus 0.5em minus 0.4em\relax PMLR, 2019, pp.
  5331--5340.

\bibitem{bing2023meta}
Z.~Bing, A.~Koch, X.~Yao, K.~Huang, and A.~Knoll, ``Meta-reinforcement learning
  via language instructions,'' in \emph{2023 IEEE International Conference on
  Robotics and Automation (ICRA)}.\hskip 1em plus 0.5em minus 0.4em\relax IEEE,
  2023, pp. 5985--5991.

\bibitem{meng2025preserving}
Y.~Meng, Z.~Bing, X.~Yao, K.~Chen, K.~Huang, Y.~Gao, F.~Sun, and A.~Knoll,
  ``Preserving and combining knowledge in robotic lifelong reinforcement
  learning,'' \emph{Nature Machine Intelligence}, vol.~7, no.~2, pp. 256--269,
  2025.

\bibitem{10.1109/TNNLS.2023.3270298}
Z.~Bing, L.~Knak, L.~Cheng, F.~O. Morin, K.~Huang, and A.~Knoll,
  ``Meta-reinforcement learning in nonstationary and nonparametric
  environments,'' \emph{IEEE Transactions on Neural Networks and Learning
  Systems}, pp. 1--15, 2023.

\bibitem{siegel2020keep}
N.~Y. Siegel \emph{et~al.}, ``Keep doing what worked: Behavioral modelling
  priors for offline reinforcement learning,'' \emph{arXiv preprint
  arXiv:2002.08396}, 2020.

\bibitem{reynolds2009gaussian}
D.~A. Reynolds \emph{et~al.}, ``Gaussian mixture models.'' \emph{Encyclopedia
  of biometrics}, vol. 741, no. 659-663, 2009.

\bibitem{nalisnick2016stick}
E.~Nalisnick and P.~Smyth, ``Stick-breaking variational autoencoders,''
  \emph{arXiv preprint arXiv:1605.06197}, 2016.

\bibitem{goyal2017nonparametric}
P.~Goyal \emph{et~al.}, ``Nonparametric variational auto-encoders for
  hierarchical representation learning,'' in \emph{Proceedings of the IEEE
  International Conference on Computer Vision}, 2017, pp. 5094--5102.

\bibitem{blei2006variational}
D.~M. Blei and M.~I. Jordan, ``Variational inference for dirichlet process
  mixtures,'' \emph{Bayesian Analysis}, vol.~1, no.~1, pp. 121--144, 2006.

\bibitem{haarnoja2018soft}
T.~Haarnoja, A.~Zhou, P.~Abbeel, and S.~Levine, ``Soft actor-critic: Off-policy
  maximum entropy deep reinforcement learning with a stochastic actor,'' in
  \emph{International conference on machine learning}.\hskip 1em plus 0.5em
  minus 0.4em\relax PMLR, 2018, pp. 1861--1870.

\bibitem{sutton1999between}
R.~S. Sutton, D.~Precup, and S.~Singh, ``Between mdps and semi-mdps: A
  framework for temporal abstraction in reinforcement learning,''
  \emph{Artificial intelligence}, pp. 181--211, 1999.

\bibitem{carion2025sam}
N.~Carion, L.~Gustafson, Y.-T. Hu, S.~Debnath, R.~Hu, D.~Suris, C.~Ryali, K.~V.
  Alwala, H.~Khedr, A.~Huang \emph{et~al.}, ``Sam 3: Segment anything with
  concepts,'' \emph{arXiv preprint arXiv:2511.16719}, 2025.

\end{thebibliography}
% %%%%%%%%%%%%%%%%%%%%%%%%%

% Generated by IEEEtran.bst, version: 1.14 (2015/08/26)

% -- End of paper --
\end{document}